%% file: main.tex
\documentclass[10pt,twocolumn,letterpaper]{article}

\usepackage{iccv}              

\definecolor{iccvblue}{rgb}{0.21,0.49,0.74}
\usepackage[pagebackref,breaklinks,colorlinks,allcolors=iccvblue]{hyperref}
\usepackage{ulem}

\def\paperID{14581} 
\def\confName{ICCV}
\def\confYear{2025}

\title{EDiT: Efficient Diffusion Transformers with Linear Compressed Attention}

\author{
Philipp Becker$^{1,2}$
\and Abhinav Mehrotra$^{1}$ 
\and Ruchika Chavhan$^{1}$ 
\and Malcolm Chadwick$^{1}$
\and Luca Morreale$^{1}$
\and Mehdi Noroozi$^{1}$ 
\and Alberto Gil C. P. Ramos$^{1}$
\and Sourav Bhattacharya$^{1}$ 
\and {$^1$} Samsung, AI Center Cambridge
\and {$^2$} Karlsruhe Institute of Technology 
}

\begin{document}
\maketitle
\input{sec/0_abstract}    
\input{sec/1_intro}

\input{sec/2_background}

\input{sec/3_method}
\input{sec/4_experiments}
\input{sec/5_related_work}
\input{sec/6_conclusion}

\clearpage
 \small \bibliographystyle{ieeenat_fullname} \bibliography{main}

\clearpage
\appendix

\input{appendix/limitations}
\input{appendix/runtime}
\input{appendix/finetuning}
\input{appendix/captions}

\end{document}

%% file: sec/0_abstract.tex
\begin{abstract}
Diffusion Transformers (DiTs) have emerged as a leading architecture for text-to-image synthesis, producing high-quality and photorealistic images.
However, the quadratic scaling properties of the attention in DiTs hinder image generation with higher resolution or on devices with limited resources. 
This work introduces an efficient diffusion transformer (EDiT) to alleviate these efficiency bottlenecks in conventional DiTs and Multimodal DiTs (MM-DiTs).
First, we present a novel linear compressed attention method that uses a multi-layer convolutional network to modulate queries with local information while keys and values are aggregated spatially.
Second, we formulate a hybrid attention scheme for multimodal inputs that combines linear attention for image-to-image interactions and standard scaled dot-product attention for interactions involving prompts.
Merging these two approaches leads to an expressive, linear-time Multimodal Efficient Diffusion Transformer (MM-EDiT).
We demonstrate the effectiveness of the EDiT and MM-EDiT architectures by integrating them into PixArt-$\Sigma$ (conventional DiT) and Stable Diffusion 3.5-Medium (MM-DiT), achieving up to $2.2\times$ speedup with comparable image quality after distillation. 
\end{abstract}

%% file: sec/1_intro.tex
\vspace{-5mm}
\section{Introduction}
\label{sec:intro}
\vspace{-2mm}

Diffusion Transformers (DiTs)~\cite{peebles2023scalable}, have quickly become the de-facto choice of architecture for state-of-the-art (SOTA) text-to-image generation models such as FLUX~\cite{flux2024}, PixArt-$\Sigma$~\cite{chen2024pixartsigma}, Stable Diffusion v3~\cite{mmdit}, and Lumina-Next~\cite{lumina}. 
The majority of these models rely on standard quadratic scaled dot-product attention \cite{vaswani2017attention} to capture long-range dependencies and global context, enabling them to generate high-quality, photorealistic, and aesthetically pleasing images. 
Earlier versions of DiTs relied mainly on the conventional transformer architecture, i.e., image tokens were processed through self-attention, whereas prompt conditioning was achieved by applying cross-attention~\cite{chen2024pixartalpha, chen2024pixartsigma}. 
More recently, Multimodal DiTs (MM-DiTs) have introduced joint attention mechanisms to handle both image and prompt tokens within a single self-attention layer \cite{mmdit}. 
While enhancing image quality significantly, joint attention scales poorly due to the increased number of involved tokens. 
This limitation is further aggravated when generating high-resolution images, mainly due to the increase in the amount of input tokens, prohibiting the efficient deployment of SOTA text-to-image models, particularly on resource-constrained devices. 

\noindent
Existing research has explored strategies to alleviate this computational bottleneck within DiTs primarily by focusing on linearizing the attention mechanism~\cite{xie2024sana, liu2024linfusion} or reducing the number of tokens~\cite{chen2024pixartsigma}. For instance, Linfusion~\cite{liu2024linfusion} relies on multi-layer non-linear transformations to map input tokens, thereby enhancing the expressiveness of the linear attention mechansim. Whereas, SANA~\cite{xie2024sana} combines standard linear attention~\cite{katharopoulos2020transformers} with convolutional operations in the feed forward part of the transformer, leveraging the inherent structure of images while capturing local relationships more effectively. 
Recent work, e.g., PixArt-$\Sigma$~\cite{chen2024pixartsigma}, reduces the number of key and value tokens to improve computational efficiency.

\noindent
In this paper, we present an \underline{E}fficient \underline{Di}ffusion \underline{T}ransformer (EDiT) architecture, using a novel linear compressed attention mechanism. 
EDiT employs: (i) a multi-layer convolutional network to effectively integrate local information in query tokens,
(ii) it spatially aggregates token information in keys and values using depthwise convolutions.
These components result in an expressive and computationally efficient attention mechanism.
We evaluate EDiT by distilling PixArt-$\Sigma$, a conventional DiT model, showing significant speedups while obtaining comparable image quality. 

\noindent
To efficiently handle multimodal inputs, we introduce a hybrid attention mechanism that combines the strengths of linear and scaled dot-product attention, enabling effective MM-DiT-based models. 
In particular, we include EDiT's linear compressed attention for the computationally intensive interactions, i.e., when computing image-to-image dependencies, on the other hand, scaled dot-product attention is used for capturing text interactions. 
These modifications introduce a novel Multimodal Efficient Diffusion Transformer (MM-EDiT) architecture that achieves linear-time complexity with respect to image size without sacrificing quality. We demonstrate the effectiveness of our hybrid attention approach by distilling Stable Diffusion 3.5-Medium, an MM-DiT-based model, and show that MM-EDiT outperforms existing linear attention alternatives. 
Our contributions can be summarized as follows: 

\begin{enumerate}
\item We introduce EDiT, using a linear compressed attention mechanism for diffusion transformers (Section \ref{sec:linear-compressed-attn}).
\item We propose MM-EDiT, integrating our linear compressed attention into MM-DiTs using a novel hybrid attention formulation (Section \ref{sec:hybrid_attn}).
\item Lastly, we demonstrate the effectiveness of these novel architectures by adapting PixArt-$\Sigma$~\cite{chen2024pixartsigma} and StableDiffusion 3.5-Medium through distillation (Section \ref{sec:experiments}). 
Our extensive experiments show that EDiT and MM-EDiT can yield a speedup of up to $2.2 \times$ without compromising on image quality.
\end{enumerate}

%% file: sec/2_background.tex
\begin{figure*}[t]
    \centering
    \includegraphics[width=0.8\textwidth]{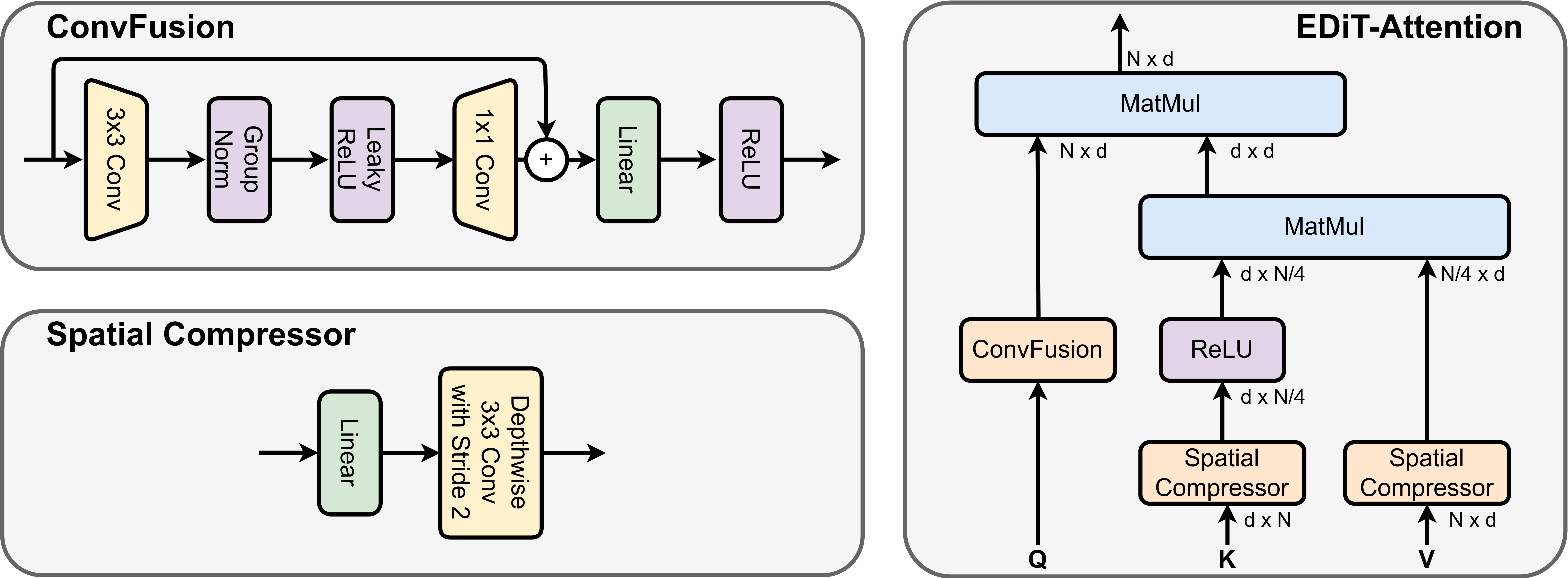}
\vspace{-2mm}
    \caption{\textbf{Left:} Our proposed ConvFusion $\phi_{\textrm{CF}}$ and Spatial Compressor functions $\phi_{\textrm{SC}}$. ConvFusion uses a multi-layer convolutional approach that integrates local information, while accounting for the $2D$-nature of the image. 
    The Spatial Compressor first projects tokens using a linear mapping before using a depthwise convolution for compression. \textbf{Right:} Combining ConvFusion for queries, the Spatial Compressor for keys and values, and linear attention results in a powerful and efficient linear compressed attention mechanism, which we use as the basis for our Efficient Diffusion Transformer (EDiT). }
\vspace{-4mm}
    \label{fig:arch}
\end{figure*}

\section{Background}
\label{sec:background}
\vspace{-2mm}
The attention mechanism, introduced by Vaswani et al. \cite{vaswani2017attention}, computes the importance of values $V$ within a sequence by comparing queries $Q$ and keys $K$. 
Conventionally, in self-attention, queries, keys, and values are computed from the same input sequence using simple linear projections of the input tokens $X \in \mathbb{R}^{N \times d}$.
Here, $N$ and $d$ denote the number of tokens and model dimension, respectively. 
Then, the output of the attention block is computed using scaled dot-product attention~\cite{vaswani2017attention}
\vspace{-2mm}
\begin{align}
\label{eq:dot-prod-attn}
Y = \mathbf{A}(Q, K, V) = \textrm{softmax}\left(\frac{QK^T}{\sqrt{d}}\right) V.
\end{align}
\vspace{-4mm}

\noindent
While this mechanism is highly successful, it requires the explicit computation of $QK^T$, which scales quadratically with the number of tokens and presents a major computational bottleneck in self-attention for approaches using scaled dot-product attention. To address this issue, several alternative attention mechanisms~\cite{katharopoulos2020transformers, gu2023mamba} have been proposed that scale \textit{linearly} with the number of tokens. 

\subsection{Linear Attention}
\vspace{-1mm} 
A closer look at Equation \eqref{eq:dot-prod-attn} suggests that the attention mechanism can be generalized with any similarity function as follows~\cite{katharopoulos2020transformers}:
\vspace{-2mm}
\begin{align}
\label{eq:general-attention}
 Y = \lbrace y_i\rbrace_{i=1:N} ~~ \text{where} ~~  y_i=\frac{\sum_{j=1}^N \operatorname{sim}\left(Q_i, K_j\right) V_j}{\sum_{j=1}^N \operatorname{sim}\left(Q_i, K_j\right)}
\end{align}
\vspace{-1mm}
If we substitute the similarity function $\operatorname{sim}(q, k)=\exp \left(\frac{q^T k}{\sqrt{d}}\right)$, Equation \eqref{eq:general-attention} is equivalent to the scaled dot-product attention in Equation \eqref{eq:dot-prod-attn}. While generalizing this even further, the self-attention can be expressed as a linear dot product of kernel feature maps: $\exp \left(\frac{q_i \cdot k_j}{\sqrt{d}}\right) v_j \approx (\phi\left(q_i\right) \cdot \phi\left(k_j\right)) \cdot v_j$, which can be further simplified to $\phi\left(q_i\right) \cdot (\phi\left(k_j\right) \cdot v_j)$ by applying the associativity property of matrix multiplication. In most linear attention works, different kernel functions are considered to formulate $K=\phi_K(X)$, $Q=\phi_Q(X)$, and $V=\phi_V(X)$. The resulting formula for linear attention is expressed as:
\vspace{-2mm}
\begin{align}
\label{eq:linear-attn}
 y_i = \mathbf{A}^{\textrm{Lin}}(Q_i, K_j, V_j) = \dfrac{Q_i \sum_{j=1}^{N} \left(K_j^T V_j \right) }{Q_i \sum_{j=1}^{N}K_j}, 
\end{align}
where $y_i$ denotes a single token in the output sequence $Y$. While the formulation of linear attention in Equation \eqref{eq:linear-attn} remains consistent across studies, the literature offers a variety of approaches to define the kernel feature maps $\phi_Q$, $\phi_K$, and $\phi_V$. Some works have demonstrated that using basic formulations for $\phi_K$ and $\phi_Q$ with functions such as $\textrm{ReLU}(x)$ or $\textrm{ELU}(x) + 1$ while keeping $\phi_V$ linear can yield performance comparable to scaled dot-product attention. 

\subsection{Efficient Attention for Diffusion Models}
\vspace{-1mm}

\textbf{SANA}~\cite{xie2024sana} adopts the conventional linear attention approach in DiTs by applying a linear projection for values and using $\phi_\textrm{SANA}(x_n) = \textrm{ReLU}(\textrm{Linear}(x_n))$ for queries and keys. However, this setup alone does not yield satisfactory results, prompting further modifications to the transformer architecture. Specifically, the standard feed-forward network at the end of each transformer block is replaced with a complex convolutional module named MixFFN, which comprises multiple convolution layers to enhance the local information of tokens.
Compared to DiTs with scaled dot-product attention, this improves latency while achieving image generation quality comparable to the SOTA.

\noindent
\textbf{LinFusion}~\cite{liu2024linfusion} explores linear attention for UNet-based diffusion approaches using Stable Diffusion v1.5~\cite{rombach2022sd15} and Stable Diffusion XL~\cite{podell2023sdxl}.
To this end, Liu et al. compute queries and keys with the following function $\phi_{\textrm{LF}}(x_n) = $
$$1 + \textrm{ELU}(x_n + \textrm{Linear}(\textrm{LN}( \textrm{LeakyReLU}(\textrm{Linear}(x_n))))),$$
where LN denotes layer normalization~\cite{ba2016layer}. The first linear layer projects each $x_n$ to a lower dimensional space, while the second linear layer projects back to the original space. For the values, they use a linear projection. This linear attention mechanism is then integrated into existing models, which are subsequently distilled.
Although, this approach works for UNets, results presented in both ~\cite{liu2024linfusion} and in this paper (see Table~\ref{tab:pixart}) show that it fails for DiTs.

\noindent
\textbf{Key--Value Token Compression.} Another approach to accelerate attention is key-value (KV) token compression, which reduces the number of keys and value tokens. This technique is implemented in the DiT-based PixArt-$\Sigma$, where Chen et al~\cite{chen2024pixartsigma} use a depthwise convolution to compress keys and values.
In practice, this operation aggregates $k \times k$ blocks of tokens into a single compressed token, which lowers the amount of keys and values by a factor of $k$. Hence, the size of the matrix $QK^T$ is also smaller by a factor of $k$,  thus improving efficiency. In PixArt-$\Sigma$, the same compression kernel is used for both keys and values. As this kernel is integrated with conventional self-attention, the computational complexity remains quadratic.

%% file: sec/3_method.tex
\section{Method}
\vspace{-2mm}

We first present EDiT, our novel linear compressed attention for DiTs that combines \textit{linear} attention with efficient token \textit{compression}. We then introduce MM-EDiT, a hybrid attention approach for MM-DiT models. 

\subsection{EDiT - Linear Attention for DiTs} 
\label{sec:linear-compressed-attn}

EDiT's linear compressed attention processes queries with multi-layered convolutions and compresses key and value tokens spatially using convolution-based feature maps. 
\noindent
\textbf{Processing Queries.} 
Inspired by the intuition behind LinFusion~\cite{liu2024linfusion} and SANA~\cite{xie2024sana}, we propose the \textit{ConvFusion} function. This operation effectively combines LinFusion and SANA's MixFFN into a unified convolutional mapping from tokens to queries. In particular, we process queries using $\phi_{\textrm{CF}}(\cdot)$ from input tokens $X$ using:
\vspace{-2mm}
\begin{align}
\label{eq:convfusion}
&Q^\textrm{EDiT} = \phi_{\textrm{CF}}(X) = \\
&\textrm{ReLU} (\textrm{Linear} (X + \textrm{Conv}(\textrm{LeakyRELU}(\textrm{GN}(\textrm{Conv}(X)))))),  \nonumber 
\end{align}
where, $\textrm{GN}$ denotes group normalization~\cite{wu2018group}. The first convolution compresses along the channel dimension, while the second convolution upsamples the channels back to their original size. Unlike Linfusion~\cite{liu2024linfusion}, our proposed ConvFusion does not treat each token individually but utilizes the local information in the tokens. In particular, we reshape the sequence to the original latent image's shape and then apply 2D convolutions. To keep convolution efficient, the compression is performed with kernel size $3\times3$, while upsampling with a $1\times1$ kernel. We illustrate the ConvFusion mechanism in Figure \ref{fig:arch}.

\noindent
\textbf{Processing Keys and Values.} Next, we introduce \textit{Spatial Compressor}, 
a convolution-based spatial compression applied to keys and values (shown in Figure~\ref{fig:arch}). This operation reduces the complexity of $K^TV$ in Equation \ref{eq:linear-attn} and it is formulated as follows:
\begin{align}
\label{eq:compressing-conv}
    K^{\textrm{EDiT}} = & \textrm{ReLU}(\phi_{\textrm{SC}}(X)) ~~ \text{and} ~~ V^{\textrm{EDiT}} = \phi_{\textrm{SC}}(X) \\
    &\text{where,} ~~ \phi_{\textrm{SC}}(X) = \textrm{Conv}(\textrm{Linear}(X)) . \nonumber
\end{align}

\noindent
In practice, $\phi_\textrm{SC}(X)$ has a $3 \times 3$ depthwise convolutional kernel with a stride of $2$, which reduces the number of keys and values by a factor of $4$. 
Although, token compression techniques have been explored in PixArt-$\Sigma$~\citep{chen2024pixartsigma}, their investigation is limited to scaled dot-product attention. In contrast, we integrate Spatial Compressor into linear attention mechanisms, improving efficiency over PixArt-$\Sigma$.

\noindent
\textbf{Formulating Linear Attention in EDiT.} We now plug queries $Q^\textrm{EDiT}$ from ConvFusion with keys $K^\textrm{EDiT}$ and values $V^\textrm{EDiT}$ from Spatial Compressor into the linear attention formulation in Equation \eqref{eq:linear-attn}. Our proposed mechanism for Efficient DiTs, termed as EDiT, is then used as a replacement for self-attention in DiTs. Figure~\ref{fig:arch} shows an overview of our approach.

\subsection{MM-EDiT - Hybrid Attention for MM-DiTs} 
\label{sec:hybrid_attn}

\begin{figure*}[t]
    \centering
    \includegraphics[width=0.8\textwidth]{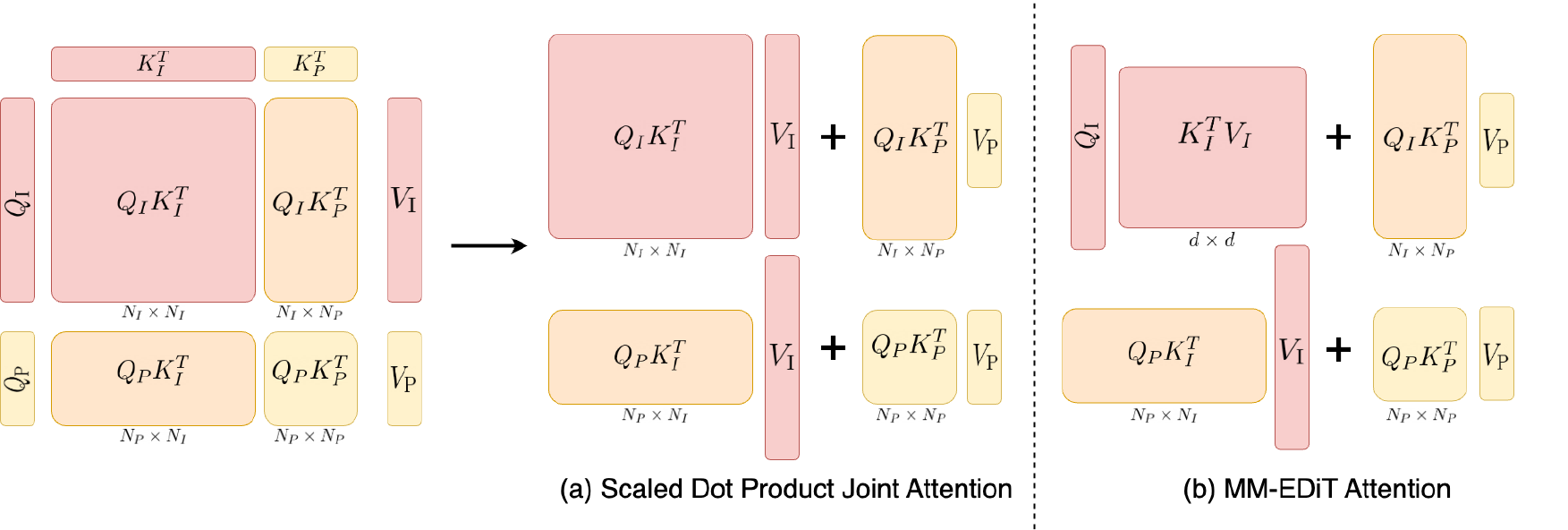}
\vspace{-4mm}
    \caption{\textbf{(a) Scaled Dot Product Joint Attention:} Image-to-Image attention matrix $(Q_IK_I^T)$ grows quadratically with number of image tokens. \textbf{(b) MM-EDiT Attention:} Our hybrid attention approach combines scaled dot product attention and linear attention, which reduces the time complexity of attention to $\mathcal{O}(N_I)$. We use ConvFusion for $Q_I$, while $K_I$ and $V_I$ are obtained through Spatial Compressor.} 
    \label{fig:hybrid_attn}
\end{figure*}

While DiTs employ separate self-attention for image tokens and cross-attention for text prompts, Multimodal Diffusion Transformers (MM-DiTs) streamline the process in a \textit{joint attention} approach, applying self-attention over concatenated tokens. 
Given that the number of image tokens typically far exceeds that of the text tokens, the quadratic complexity of the attention mechanism primarily depends on the number of image tokens. This issue is further aggravated for high-resolution images, where the number of image tokens is significantly larger.
Therefore, we design a hybrid attention mechanism to exploit linear attention for image-to-image relations, while leveraging the more powerful scaled dot-product attention for prompt tokens.
Merging this hybrid attention with EDiT's linear compressed attention leads to an expressive and linear-time Multimodal Efficient Diffusion Transformer (MM-EDiT).

\noindent
\textbf{A Closer Look at Joint Attention.}
We denote the image and prompt tokens with $X_I \in \mathbb{R}^{N_I \times d}$ and $X_P \in \mathbb{R}^{N_P \times d}$, where $N_I$ and $N_P$ are the number of image and prompt tokens. We refer to queries, keys, values for image and prompt by $Q_\textrm{I}$, $K_\textrm{I}$, $V_\textrm{I}$, and $Q_\textrm{P}$, $K_\textrm{P}$, $V_\textrm{P}$ respectively. 
Finally, we define $Q$, $K$, and $V$ as the concatenation of the corresponding image and text tokens, which are plugged into Equation \eqref{eq:dot-prod-attn} to calculate the output of the attention block. However, we can also decompose the joint attention mechanism into four components: (i) image tokens attending to image tokens, (ii) image tokens attending to prompt tokens, (iii) prompt tokens attending to image tokens, and (iv) prompt tokens attending to themselves.
This decomposition is visualized in Figure~\ref{fig:hybrid_attn}(a) and results in 
\begin{align}
\label{eq:joint-attn}
 &\mathbf{A}(Q, K, V)  =  \textrm{softmax} \left( \frac{1}{\sqrt{d}}\begin{pmatrix} Q_I \\ Q_P \end{pmatrix} \begin{pmatrix} K_\textrm{I}^T & K_P^T \end{pmatrix}  \right) \begin{pmatrix} V_I \\ V_\textrm{P} \end{pmatrix} \nonumber \\   
 &=  \textrm{softmax}  \left( \frac{1}{\sqrt{d}}\begin{pmatrix}  Q_IK_I^T &  Q_IK_P^T \\ Q_PK_I^T &  Q_PK_P^T \end{pmatrix} \right) \begin{pmatrix} V_I \\ V_P \end{pmatrix}  \\
 &=  \begin{pmatrix}
    \eta_{I} \cdot \mathbf{A}(Q_I, K_I, V_I) + (1 - \eta_{I}) \cdot \mathbf{A}(Q_I, K_P, V_P) \\
    \eta_{P} \cdot \mathbf{A}(Q_P, K_I, V_I) + 
    (1 - \eta_{P}) \cdot \mathbf{A}(Q_P, K_P, V_P)\\ 
\end{pmatrix} \nonumber
\end{align}
where we define $\eta_{I} = \eta(Q_I, K_I, K_P)$ and $\eta_{P} = \eta(Q_P, K_P, K_I)$. 
With this decomposition, the softmax in $\mathbf{A}(\cdot, \cdot, \cdot)$ is now applied over different token sets compared to joint attention, resulting in row sums that are not equal to 1. Therefore, a normalization factor, $\eta(\cdot, \cdot,  \cdot)$, as shown in Equation \eqref{eq:joint-attn}, needs to be added to adjust each element of the attention matrix so that the output remains unchanged:
\begin{align*}
\eta (\mathbf{q}, \mathbf{k}_1, \mathbf{k}_2) = \dfrac{\sum_j\exp(\mathbf{q}\mathbf{k}_1^T)_j}{\sum_j\exp(\mathbf{q}\mathbf{k}_1^T)_j + \sum_{j'}\exp(\mathbf{q}\mathbf{k}_2^T)_{j'}} 
\end{align*}

\noindent
\textbf{Combining Linear and Scaled Dot-Product Attention in MM-EDiT.}
The most computationally intensive component of joint attention is $Q_IK_I^T$, which scales quadratically with the image resolution. To optimize it, we apply EDiT to linearize the image-image attention segment, thus significantly reducing computation time. Overall, our method integrates the proposed EDiT linear attention for the image-image attention block and scaled dot-product attention for the other parts. We refer to the resulting approach as MM-EDiT and illustrate it in Figure~\ref{fig:hybrid_attn}(b).

\noindent
We begin by computing $Q_I$ using ConvFusion (Equation \eqref{eq:convfusion}) and obtaining $K_I$ and $V_I$ through Spatial Compressor (Equation \eqref{eq:compressing-conv}). For prompt tokens, $Q_P$, $K_P$, and $V_P$ are derived using simple linear projections, as in the original model. Following the MM-DiT approach, we then concatenate the image and prompt tokens. Finally, we replace $\mathbf{A}(Q_I, K_I, V_I)$ in the scaled dot-product joint attention (Equation \eqref{eq:joint-attn}) with $\mathbf{A}^{\textrm{Lin}}(Q_I, K_I, V_I)$, resulting in the hybrid joint attention matrix $\mathbf{A}^{\textrm{Hybrid}}(Q, K, V)$ formulated as follows:

\vspace{-2mm}
\begin{align}
\label{eq:hybrid-attn}
 &\mathbf{A}^{\textrm{Hybrid}}(Q, K, V)  = \\ 
 &  \begin{pmatrix}
    \eta_{I}^{Lin} \cdot \mathbf{A}^{Lin}(Q_I, K_I, V_I) + (1 - \eta_{I}^{Lin}) \cdot \mathbf{A}(Q_I, K_P, V_P) \\
    \eta_{P} \cdot \mathbf{A}(Q_P, K_I, V_I) + 
    (1 - \eta_{P}) \cdot \mathbf{A}(Q_P, K_P, V_P)\\ 
\end{pmatrix} \nonumber
\end{align}

\noindent
Here, $\eta_I^{\textrm{Lin}} = \eta^{\textrm{Lin}}(Q_I, K_I, K_P)$ where 
\[
\eta^{\textrm{Lin}}(\mathbf{q}, \mathbf{k}_1, \mathbf{k}_2) = \dfrac{\sum_j (\mathbf{q}\mathbf{k}_1^T)_j}{\sum_j(\mathbf{q}\mathbf{k}_1^T)_j + \sum_{j'}\exp( \mathbf{q} \mathbf{k}_2^T)_{j'}}
\]

\noindent
With this formulation, the corresponding normalization terms must be calculated for each block. Unfortunately, efficient implementations like FlashAttention do not provide access to them, requiring a less efficient, custom implementation of scaled dot-product attention. 
To avoid this, we can approximate $\eta^{Lin}$ via the number of tokens, i.e,
$\hat{\eta}^{\textrm{Lin}} = \frac{N_I}{N_I + N_T}$.  
We show in Section \ref{sec:eval_mmedit} that this approximation leads not only to faster computations but also to slightly better performance, and thus consider it the standard for MM-EDiT. 

%% file: sec/4_experiments.tex
\section{Experiments} \label{sec:experiments}
\vspace{-2mm}

We evaluate our EDiT and MM-EDiT models by distilling two well-known diffusion transformers using our novel EDiT and MM-EDiT architectures.
We use PixArt-$\Sigma$~\cite{chen2024pixartsigma} as a conventional DiT, and StableDiffusion 3.5-Medium (SD-v3.5M)~\cite{mmdit} as the basis for our MM-DiT experiments. 
\noindent
\textbf{Dataset.} In all experiments, we rely on the YE-POP\footnote{\url{https://huggingface.co/datasets/Ejafa/ye-pop}} dataset for training. This subset of Laion-POP comprises about 480K images with two alternative captions each. 
For evaluation, we use the PixartEval30K dataset~\cite{chen2024pixartsigma}\footnote{\url{https://huggingface.co/datasets/PixArt-alpha/PixArt-Eval-30K}}.  

\input{result_tables/pixart_main}

\noindent
\textbf{Training details for EDiT.} We use three publicly available checkpoints for PixArt-$\Sigma$, trained on $512, 1024$, and $2048$ image resolutions. We initialize our approach using the $512\times512$ PixArt-$\Sigma$ checkpoint, modify the self-attention layers using our linear compressed attention approach, and distill the model for 30 epochs. 
In particular, we follow \cite{liu2024linfusion,lee2025koala} and combine a noise prediction loss with knowledge distillation, where we minimize the difference between the student and teacher's prediction, and feature distillation with the output of each self-attention between teacher and student. Thereafter, we take the resulting model, unfreeze all weights of EDiT, and train for 10 more epochs at $1024$ resolution using the teacher checkpoint for the respective resolution.
We repeat this process with $2048$ resolution images for $8$ more epochs. 

\noindent
\textbf{Training details for MM-EDiT.} 
We use Stable Diffusion 3.5 Medium (SD-v3.5M) to evaluate the efficacy of our MM-EDiT mechanism. 
In contrast to PixArt-$\Sigma$, there is only one publicly available checkpoint for SD-v3.5M, which can be used for multiple resolutions.  
We take this pretrained SD-v3.5M checkpoint and replace the multimodal attention layers with MM-EDiT. 
We first train the hybrid attention layers for 45 epochs on $512 \times 512$ images using a combined loss similar to that used for EDiT.
We again combine a task loss, knowledge, and feature distillation, but this time, the task loss is based on the rectified flow mechanism~\cite{liu2023flow} taken from SD-v3.5M.   
Following this, we unfreeze all weights of the MM-EDiT modules and train them for 10 more epochs at a resolution of $1024 \times 1024$ only with knowledge distillation loss. 

\noindent
\textbf{Metrics:} We report CLIP similarity scores between generated images and text to evaluate prompt adherence and FID with Inception-v3 and CLIP encoders to assess the quality of generated images. 

\subsection{Evaluations for EDiT}
\vspace{-1mm}

\input{result_tables/sd35_main}

\textbf{Baselines and Ablations.}
We compare the performance of our EDiT version distilled from PixArt-$\Sigma$ with the original PixArt-$\Sigma$ model.  
Additionally, we evaluate EDiT against Linfusion-DiT, which uses Linfusion's attention mechanism~\cite{liu2024linfusion} within PixArt-$\Sigma$, reproducing the results of \cite{liu2024linfusion}. 
We further consider SANA-DiT, which uses the linear attention and MixFFN from SANA~\citep{xie2024sana} to augment the PixArt-$\Sigma$ architecture as well as the original KV Token Compression (KV Comp) approach from~\cite{chen2024pixartsigma}.
For consistency, we use our distillation approach for all architectures.  
Finally, we evaluated EDiT's design choices for computing queries, keys, and values using three settings: applying ConvFusion on queries and keys, using only Spatial Compressor on keys and values, and applying ConvFusion on queries without compressing keys and values. This helps us understand the impact of enhancing token interactions in the queries and shows that compressing does not harm performance.

\noindent
\textbf{Quantitative Results.}
Table \ref{tab:pixart} summarizes our EDiT results based on Pixart-$\Sigma$. At both $512 \times 512$ and $1024 \times 1024$ resolutions, EDiT performs on par with the teacher model (i.e., Pixart-$\Sigma$ baseline model) — exceeding its performance at the lower resolution. Additionally, our model surpasses the linear approaches SANA-DiT and Linfusion by 2.6 and 8.8 Inception-v3 FID points for $512 \times 512$ resolution, respectively, and outperforms the KV Comp token compression method by 2.5 FID points. We observe a similar performance of EDiT on $1024 \times 1024$ resolution. These results demonstrate that our EDiT model outperforms current linear attention and token compression approaches.
Comparing EDiT to its ablations shows our mixture of ConvFusion and Spatial Compressor strikes an excellent balance between image quality and runtime. 
To show the effectiveness of our approach while generating high-resolution images, in Figure \ref{fig:edit_vis_2k}, we present the results for EDiT and PixArt-$\Sigma$ at a resolution of $2048 \times 2048$, where EDiT performs comparably but requires around $2.5\times$ less time.

\input{figures/pixart_images_2k/_pixart_sigma_2k}

\subsection{Evaluations for MM-EDiT}
\label{sec:eval_mmedit}
\vspace{-1mm}

\noindent
\textbf{Baselines and Ablations.}
Besides comparing our MM-EDiT-based SD-v3.5M with the original model, we 
also compare MM-EDiT's performance with three other baselines using linear attention in MM-DiTs: Linear MM-DiT-$\alpha$, which computes queries, keys and tokens like MM-EDiT, but uses linear attention for all relations; Linear MM-DiT-$\beta$, which incorporates $\phi_{\textrm{LF}}$ from Linfusion for prompt tokens; and SANA-MM-DiT, which uses $\phi_{\textrm{SANA}}$ and MixFFN layers.
Additionally, we considered two variations of our hybrid attention mechanism: MM-EDiT with $\eta^{Lin}$ (Eq \ref{eq:hybrid-attn}), which uses theoretically correct adjustment terms, and MM-EDiT without $\phi_{CF}$ and $\phi_{SC}$.
These comparisons highlight the importance of hybrid attention and token interaction.

\noindent
\textbf{Quantitative Results}
Table \ref{tab:sd35-main} presents the performance of MM-EDiT based on SDv3.5M, comparing both hybrid and non-hybrid attention methods. Our approach outperforms non-hybrid methods such as SANA-MM-DiT, Linear MM-DiT-$\alpha$, and Linear MM-DiT-$\beta$, highlighting the importance of hybrid attention in MM-DiTs. Note that our Linear MM-DiT-$\beta$ approach achieves a comparable FID score, but the qualitative results are inferior than the MM-EDiT (see Appendix~\ref{app:sec:cap}).
Additionally, when we replace $\phi_{CF}$ and $\phi_{SC}$ with basic linear functions, the performance drops considerably, highlighting the value of our convolution-based functions.

\noindent
\subsection{Qualitative Comparisons}
\vspace{-2mm}

\input{figures/pixart_images_1k/_pixart_quali_main_1k}

\noindent
To demonstrate the visual quality of the images produced by EDiT and MM-EDiT, Figure~\ref{fig:edit_vis} shows example images generated with our models. 
Appendix~\ref{app:sec:cap} provides the used prompts and images for the remaining approaches.
These images demonstrate high quality, visually indistinguishable from that of the base model, and high prompt adherence. 

\begin{table}[t]
    \centering
    \resizebox{\linewidth}{!}{\begin{tabular}{l||ccccc}
        &  $1024 \times 1024$ & $2048 \times 2048$ & $4096 \times 4096 $ & $8192 \times 8192$\\
         \hline
         PixArt-$\Sigma$ & 0.047 & 0.387 & 4.770 &  72.96\\
         SANA-DiT & 0.043 & 0.166 & 0.687 & 21.76 \\
         \textbf{EDiT} (Ours) & 0.034 & 0.121 & 0.461 & 1.693\\
         \hline
          SD-v3.5M &  0.071 & 0.474 & 5.603 & -- \\
   SANA-MM-DiT  &  0.069 & 0.228 & 0.861 & -- \\
    \textbf{MM-EDiT} (Ours) &  0.053 & 0.169 & 0.623 & -- \\
    \end{tabular}}
\vspace{-3mm}
    \caption{Latency measurements with varying image resolution on a NVIDIA A100. Here, we present the time taken for a single forward pass through the transformer component. Our results demonstrate that EDiT and MM-EDiT are significantly faster at higher resolutions.
    SD-v3.5M cannot generate $8K$ resolution images due to the limited size of positional embedding.}
    \label{tab:latency-analysis}
\end{table}

\begin{figure}
    \centering
    \includegraphics[width=0.4\textwidth]{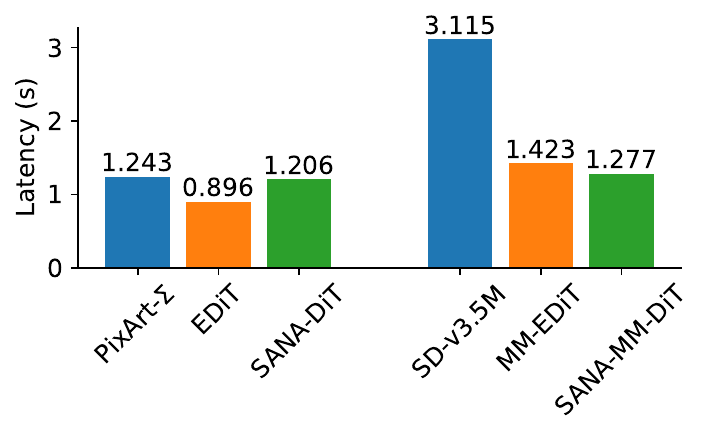}
    \vspace{-5mm}
    \caption{Latency measurements on a Samsung S25 Ultra with SM8750-AB Snapdragon 8 Elite chipset. We compute the latency with 4-bits quantized models to do a single forward pass of the DiTs to generate a $1024 \times 1024$ image.}
    \label{tab:phone_rt}
    \vspace{-2mm}
\end{figure}

\subsection{Latency Analysis}
Table \ref{tab:latency-analysis} reports latency measurements for a single forward pass through the transformer module across baseline models and our proposed EDiT and MM-EDiT architectures, evaluated on a NVIDIA A100 GPU. We test a range of image resolutions, from $1024 \times 1024$ up to $8192 \times 8192$. At the highest resolution, our EDiT model achieves a latency of approximately 1.7 seconds—making it $43\times$ faster than Pixart-$\Sigma$ and $13\times$ faster than SANA-DiT. Similarly, our MM-EDiT approach is $9\times$ faster than SD-v3.5M when generating images at $4096 \times 4096$ resolution. Notably, the latency improvement of EDiT and MM-EDiT over standard baselines scales quadratically with image resolution, underscoring their suitability for high-resolution image generation.
Overall, EDiT and MM-EDiT strike the best trade-off between runtime and quality across all considered variants. 
Appendix~\ref{sec:add_rt} shows some additional runtime analysis on a consumer-grade GPU.

\noindent
\textbf{On-device Runtime Analysis}
In this section, we demonstrate the potential of our methods for reducing runtimes on edge devices. Figure \ref{tab:phone_rt} presents a runtime analysis of MM-EDiT on a Qualcomm SM8750-AB Snapdragon 8 Elite—a processor used in state-of-the-art smartphones. We compare the runtimes of our EDiT and MM-EDiT with SANA-based baselines (SANA-DiT and SANA-MM-DiT) and their corresponding base models (Pixart-$\Sigma$ and SD-v3.5M). EDiT achieves a latency of 896 milliseconds—38\% and 26\% faster than base and SANA-DiT models, respectively. MM-EDiT reduces the runtime of SD-v3.5M by $2.2 \times $, although its runtime remains slightly higher than that of SANA-MM-DiT. Further investigation revealed that the scaled dot-product attention operations for both image-text $Q_IK_P^T$ and text-image $Q_PK_I^T$ interactions are computationally expensive on the device. This is primarily because token compression is not applied to image queries $Q_I$, making the dot-product calculation expensive. Even though text-image interactions $Q_PK_I^T$ are also resource-intensive, they incur a lower cost due to token compression applied to $K_I$. In contrast, SANA's fully linear attention mechanism likely contributes to its lower runtime compared to MM-EDiT, albeit at the expense of image quality. To address this, we encourage future research to explore alternative interaction strategies within hybrid attention for MM-DiTs.

%% file: result_tables/pixart_main.tex
\begin{table*}
\centering
\vspace{-2mm}
{\small
\begin{tabular}{l||ccc|ccc}
    & CLIP $\uparrow$ &  FID $\downarrow$ & FID  $\downarrow$ &  CLIP 
    $\uparrow$ & FID $\downarrow$ & FID  $\downarrow$  \\
    & & (Inception-v3) & (Clip) & & (Inception-v3) & (Clip)  \\ 
    \hline
    & \multicolumn{3}{c|}{512 $\times$ 512}  & \multicolumn{3}{c}{1024 $\times$ 1024}\\  
    \hline
    PixArt-$\Sigma$  & 0.285 & \phantom{0}7.57 & \phantom{0}2.50& 0.285 & \phantom{0}7.09 & \phantom{0}2.53   \\ 
    \textbf{EDiT} (ours)  & 0.283 & \phantom{0}7.06 & \phantom{0}2.57 & 0.290 & \phantom{0}7.82 & \phantom{0}2.64  \\
    \hline
    SANA-DiT & 0.283 & \phantom{0}8.43 & \phantom{0}3.31 & 0.286 &\phantom{0}9.31 & \phantom{0}3.16 \\ 
    LinFusion-DiT  & 0.289 & 15.87 & \phantom{0}5.98 & 0.283 & 44.66  & 11.01 \\ 
    KV Comp. ($k=2$) &0.275 & 10.69 & \phantom{0}3.77 & 0.283 & 10.32 & \phantom{0}3.50  \\ 
    \hline
    \multirow{4}{*}{\rotatebox[origin=c
    ]{90}{Ablations} \begin{tabular}{ccc}
        Q & K & V  \\
        \hline
        CF & CF & - \\
        - & SC & SC \\
        CF & - & -
    \end{tabular}}
     & & & & & & \\ 
    & 0.284 & \phantom{0}7.59  & \phantom{0}2.59 & 0.286 & \phantom{0}7.76 & \phantom{0}2.72  \\ 
    & 0.286 & 14.53 & \phantom{0}4.82 & 0.290 & 26.59 & \phantom{0}6.20 \\ 
    & 0.284 & \phantom{0}7.06 & \phantom{0}2.53  & 0.287 & \phantom{0}7.70 & \phantom{0}2.69  \\ 
\end{tabular}   
}
\vspace{-3mm}
\caption{Quantitative comparison of EDiT to it's teacher (PixArt-$\Sigma$), prior efficient DiT architectures and alternative design choices at resolutions $512\times512$ and $1024 \times 
1024$. We ablate different choices to compute queries, keys and values, where we denote ConvFusion and Spatial Compressor by $CF$ and $SC$. 
The results show that EDiT performs similar to the teacher, outperforms previous approaches.}
\label{tab:pixart}
\end{table*}

%% file: result_tables/sd35_main.tex
\begin{table*}
\centering
{\small
\begin{tabular}{lc||ccc|ccc}
    & Hybrid & CLIP $\uparrow$ &  FID $\downarrow$ & FID  $\downarrow$ &  CLIP 
    $\uparrow$ & FID $\downarrow$ & FID  $\downarrow$ \\
    & & & (Inception-v3) & (Clip) & & (Inception-v3) & (Clip)  \\ 
    \hline
    \multicolumn{2}{c||}{} & \multicolumn{3}{c|}{512 $\times$ 512}  & \multicolumn{3}{c}{1024 $\times$ 1024}\\  
    \hline
   
    \hline
    SD-v3.5M & -- &  0.283 & 10.49 & 3.86 & 0.286 & \phantom{0}8.83 &  2.89  \\ 
    \hline
    \textbf{MM-EDiT} (Ours) &  \textbf{\textcolor{ForestGreen}{\checkmark}} 
    & 0.285 & 11.60 & 3.91  & 0.287  &  \phantom{0}9.73  & 2.95  \\ 
    MM-EDiT with $\eta^{Lin}$  & \textbf{\textcolor{ForestGreen}{\checkmark}} & 0.283 & 13.05 & 4.44 &  0.285  & 10.77    & 3.18 \\
    MM-EDiT no $\phi_{CF}$ and $\phi_{SC}$  & \textbf{\textcolor{ForestGreen}{\checkmark}} & 0.287 & 20.98 & 5.18    & 0.279  & 17.67 &  5.96   \\
    \hline
    SANA-MM-DiT & \textbf{\textcolor{BrickRed}{$\times$}} 
    & 0.279 & 14.94 & 5.02 & 0.283  & 10.88 & 3.26  \\
    Linear MM-DiT-$\alpha$ & \textbf{\textcolor{BrickRed}{$\times$}} & 0.281 & 13.59 & 4.28 & 0.284 & 10.29 & 2.94 \\
    Linear MM-DiT-$\beta$ & \textbf{\textcolor{BrickRed}{$\times$}} & 0.281 & 13.53 & 4.37 & 0.281 &  \phantom{0}9.75 & 2.94 \\ 
\end{tabular}   
}
\vspace{-3mm}
\caption{Quantitative comparisons of MM-EDiT to it's teacher (SD-v3.5M) and alternative design choices. 
Our results demonstrate that MM-EDiT performs at par with the teacher SD-v3.5M.}
\label{tab:sd35-main}
\end{table*}

%% file: figures/pixart_images_2k/_pixart_sigma_2k.tex
\begin{figure}[t]
    \centering
    
    \begin{minipage}{0.02\textwidth}\raggedright
    \rotatebox[origin=c]{90}{PixArt-$\Sigma$}
    \end{minipage}
    \hfill
    \begin{minipage}{0.19\textwidth}
    \includegraphics[width=\textwidth]{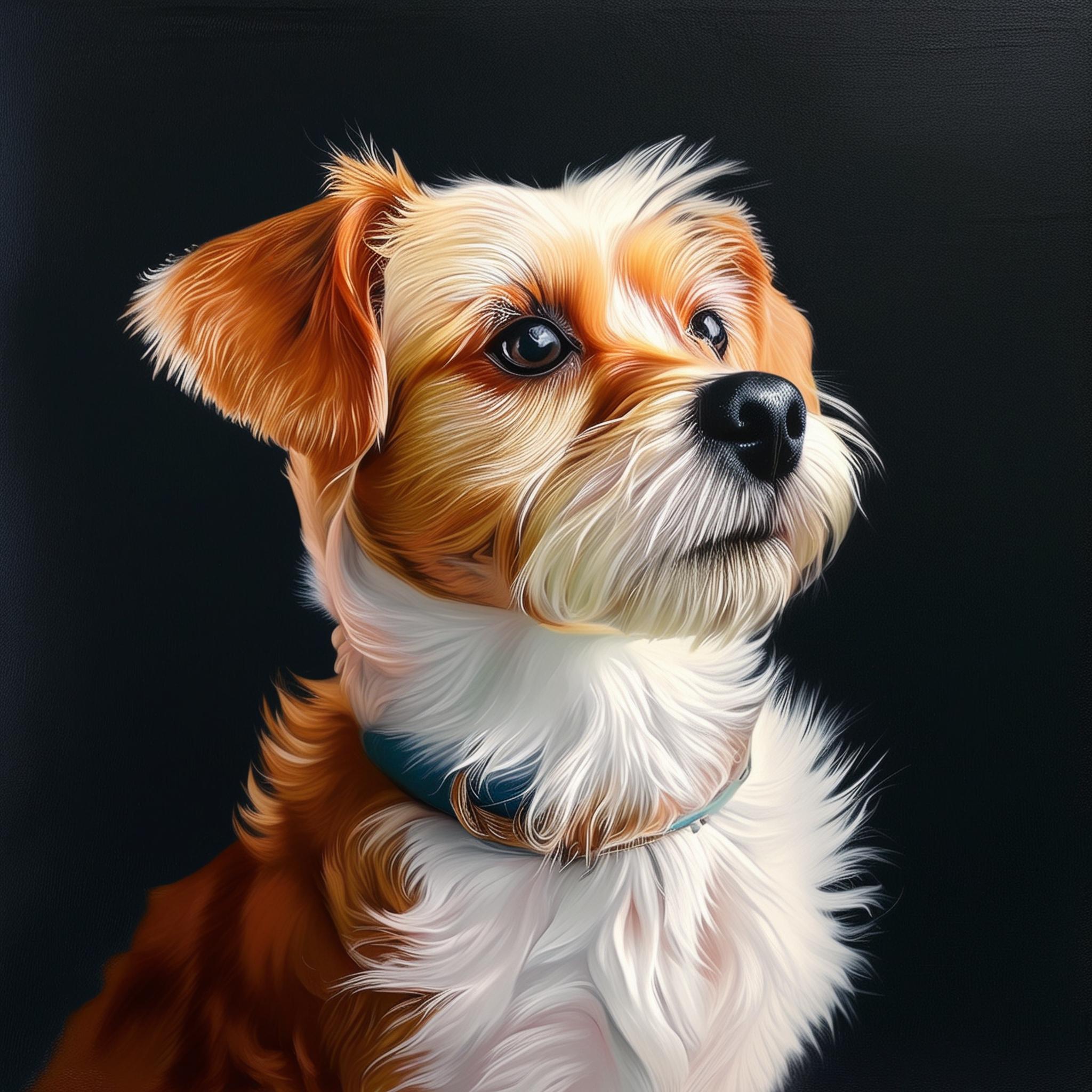}
    \end{minipage}
    \hfill
    \begin{minipage}{0.19\textwidth}
    \includegraphics[width=\textwidth]{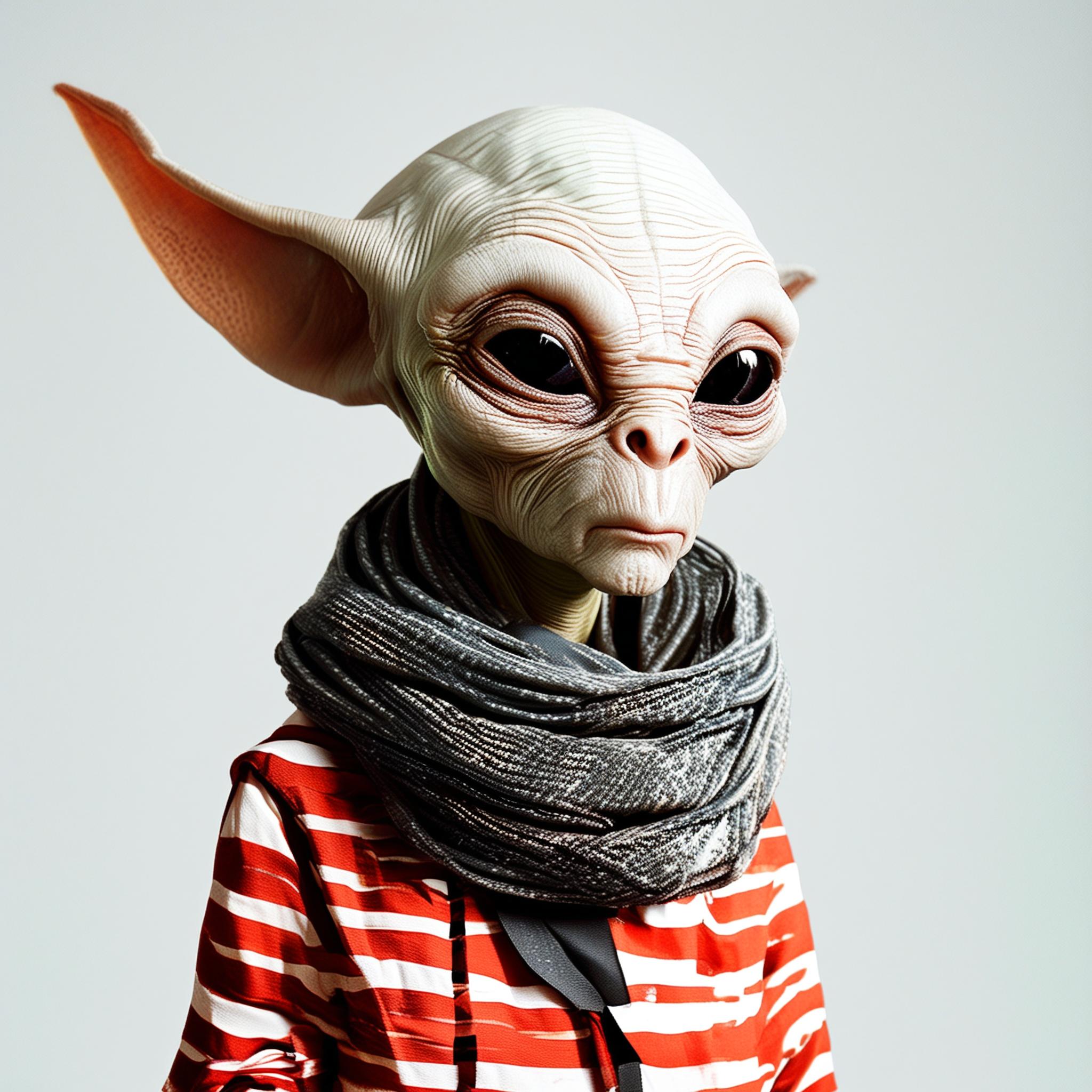}
    \end{minipage}
     \hfill

    \begin{minipage}{0.02\textwidth}\raggedright
    \rotatebox[origin=c]{90}{EDiT}
    \end{minipage}
    \hfill
    \begin{minipage}{0.19\textwidth}
    \includegraphics[width=\textwidth]{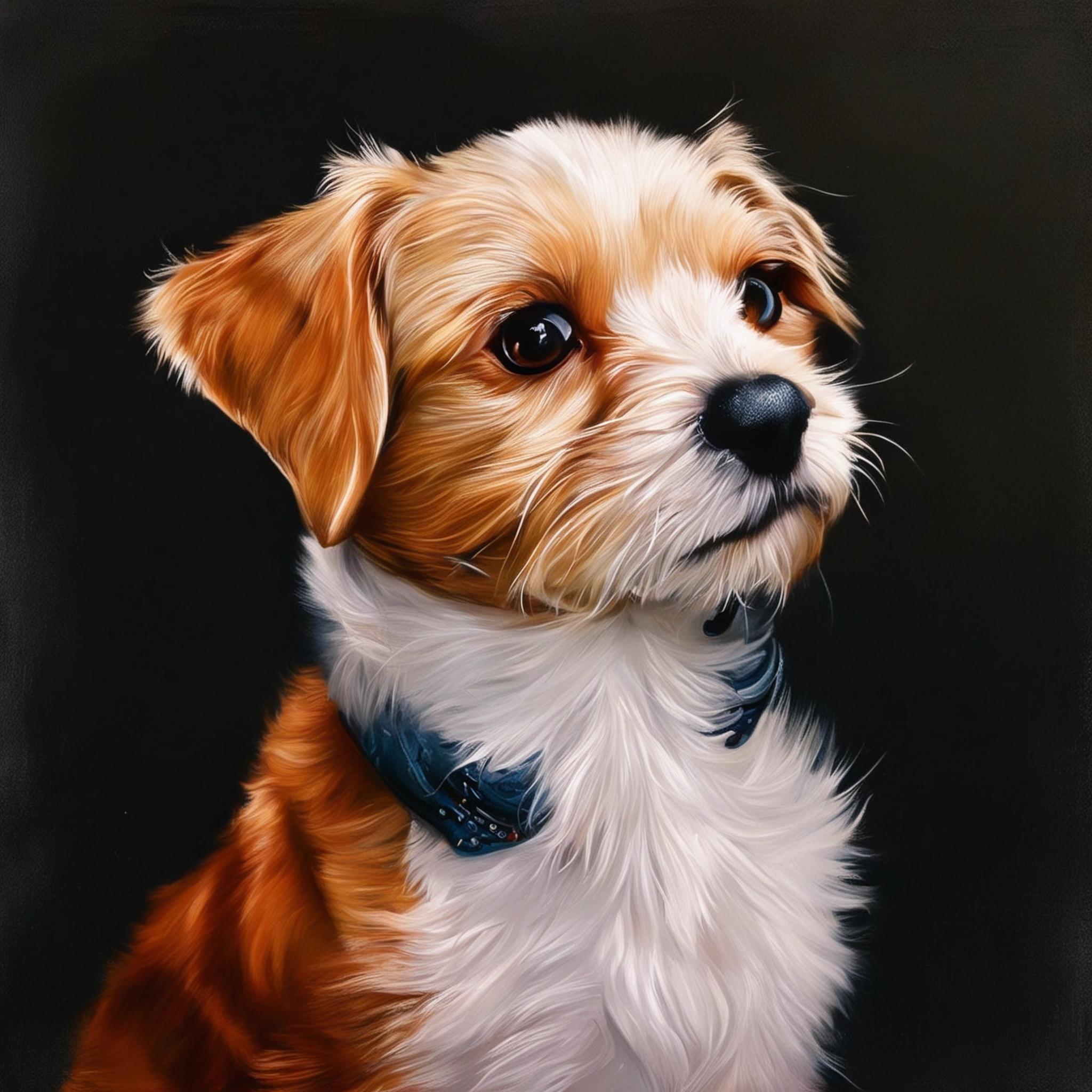}
    \end{minipage}
    \hfill
    \begin{minipage}{0.19\textwidth}
    \includegraphics[width=\textwidth]{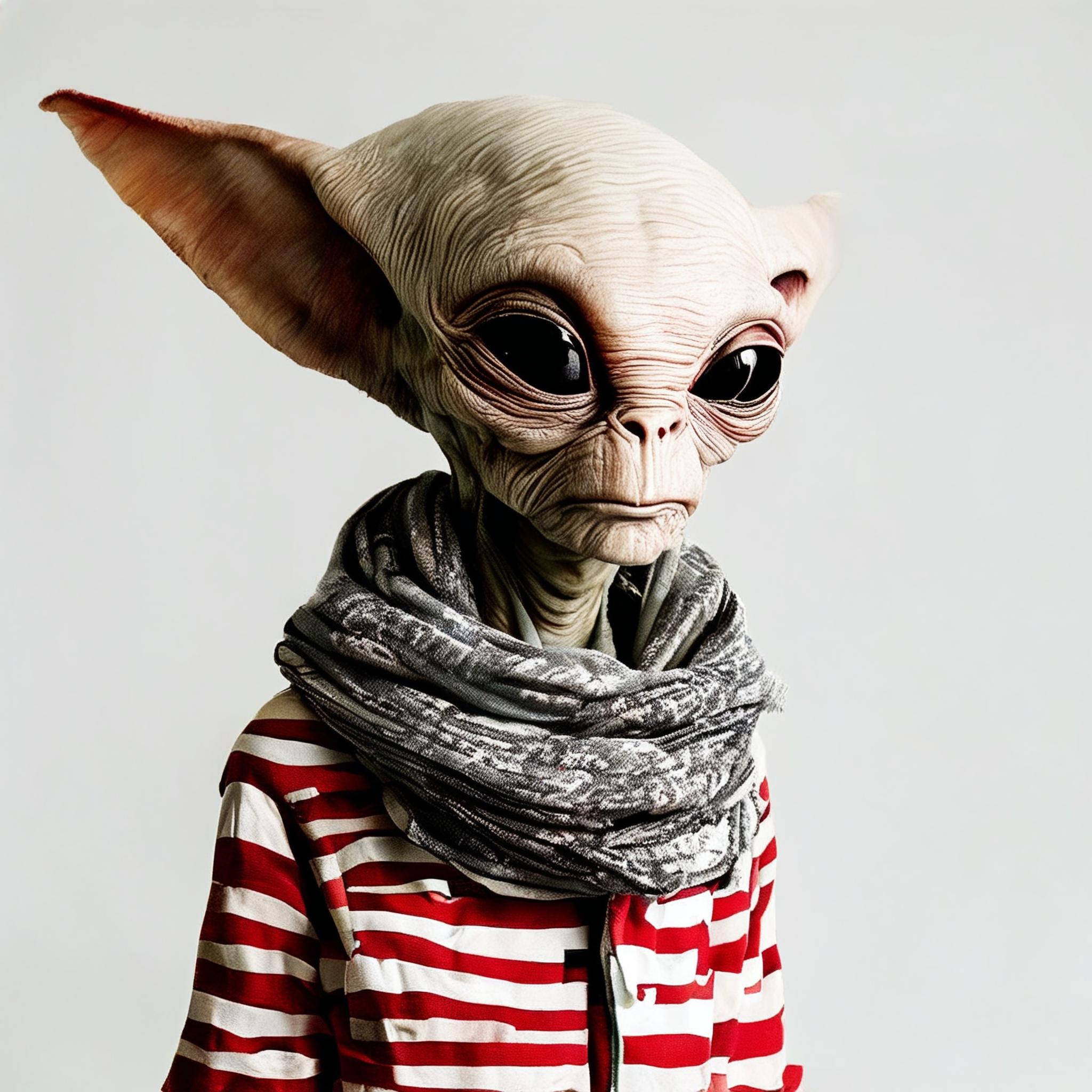}
    \end{minipage}
    \hfill

    \centering
{\small
\begin{tabular}{l||ccc}
   &  CLIP $\uparrow$ &  FID $\downarrow$ & FID $\downarrow$ \\ 
      &  &   (Inception-v3)  &(Clip)  \\ 
    \hline
    PixArt-$\Sigma$ & 0.274 & 7.48 &  2.59 \\   
    \textbf{EDiT} (ours) & 0.288 & 8.29 & 2.71 \\ 
\end{tabular}   
}
\vspace{-2mm}
\caption{Qualitative and quantitative comparision of EDiT and PixArt-$\Sigma$ at a resolution of $2048 \times 2048$ pixels. EDiT generates comparable results whie achiving a speedup of almost $2.5 \times$.}
\vspace{-2mm}
    \label{fig:edit_vis_2k}
\end{figure}

%% file: figures/pixart_images_1k/_pixart_quali_main_1k.tex
\begin{figure*}[t]
    \centering
    
    \begin{minipage}{0.02\textwidth}\raggedright
    \rotatebox[origin=c]{90}{PixArt-$\Sigma$}
    \end{minipage}
    \begin{minipage}{0.19\textwidth}
    \includegraphics[width=\textwidth]{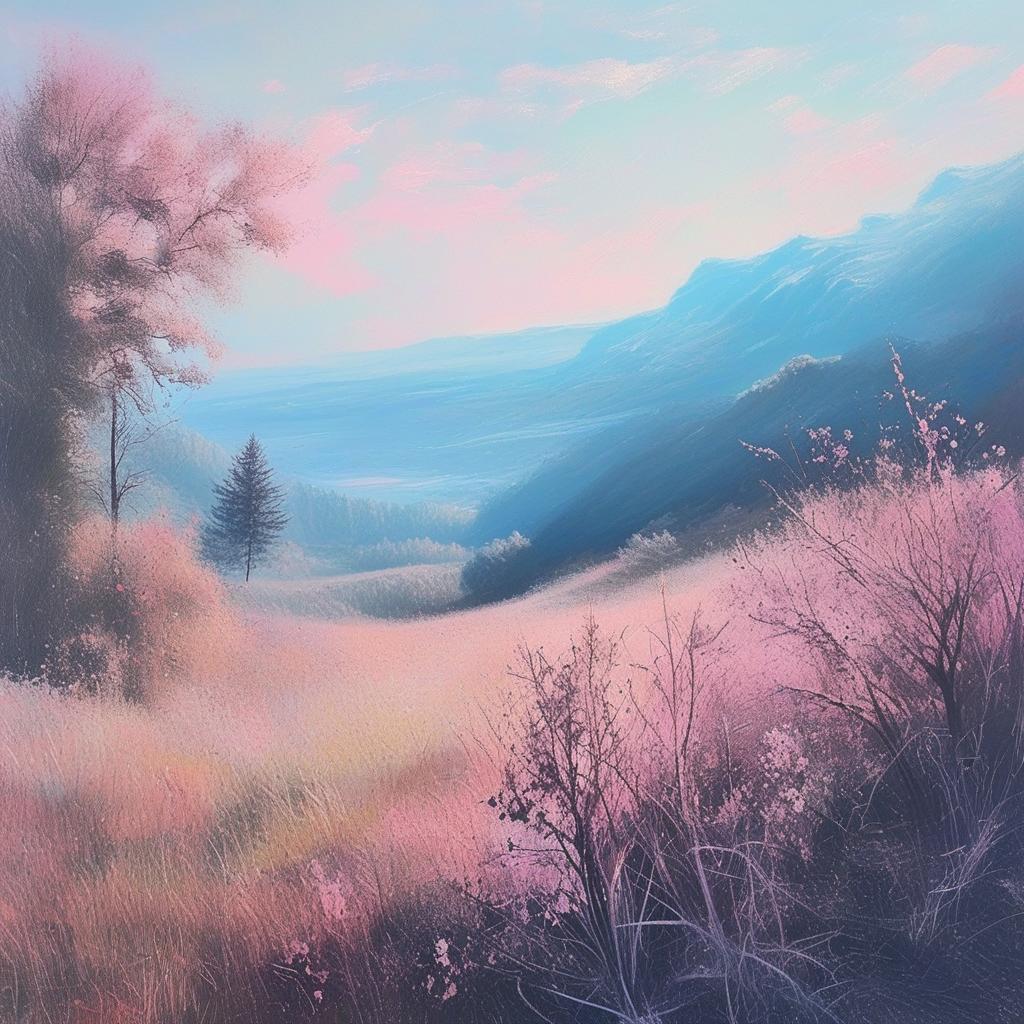}
    \end{minipage}
    \hfill
    \begin{minipage}{0.19\textwidth}
    \includegraphics[width=\textwidth]{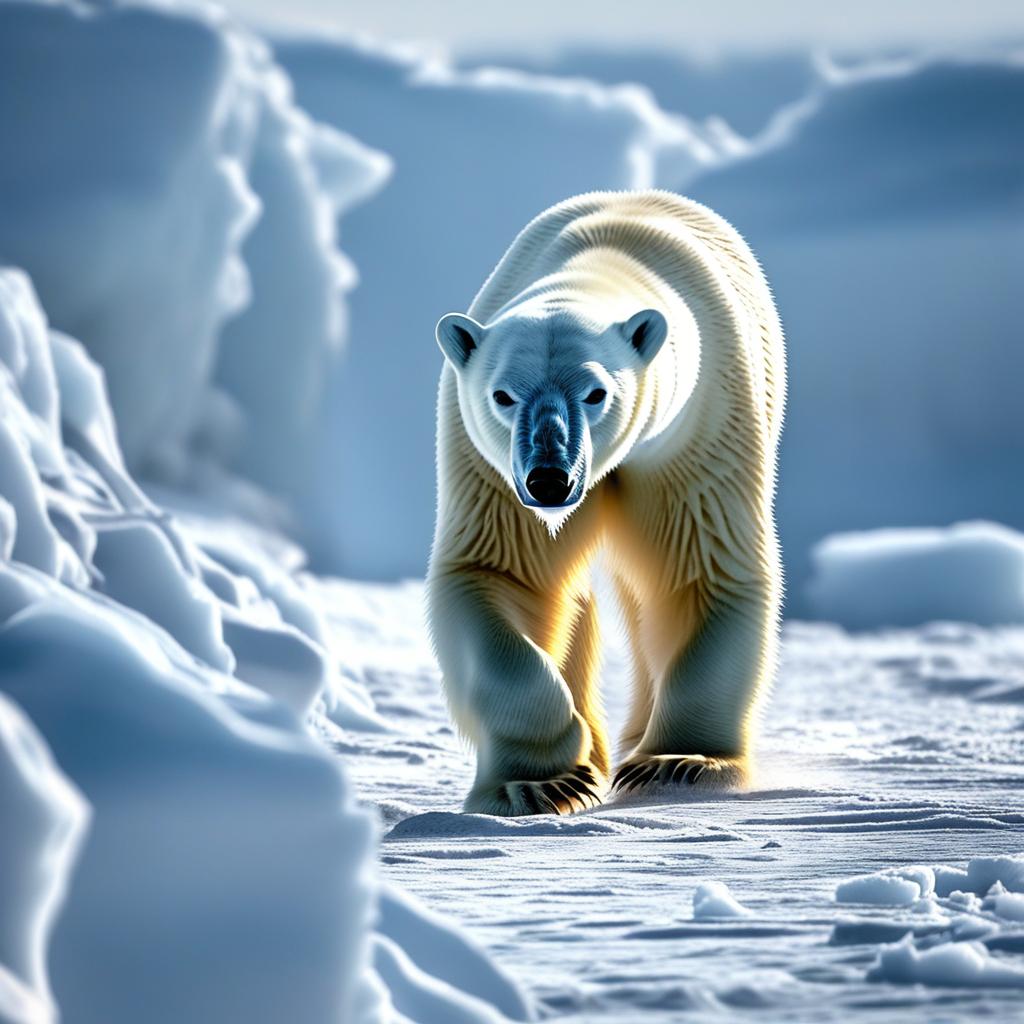}
    \end{minipage}
    \hfill
    \begin{minipage}{0.19\textwidth}
    \includegraphics[width=\textwidth]{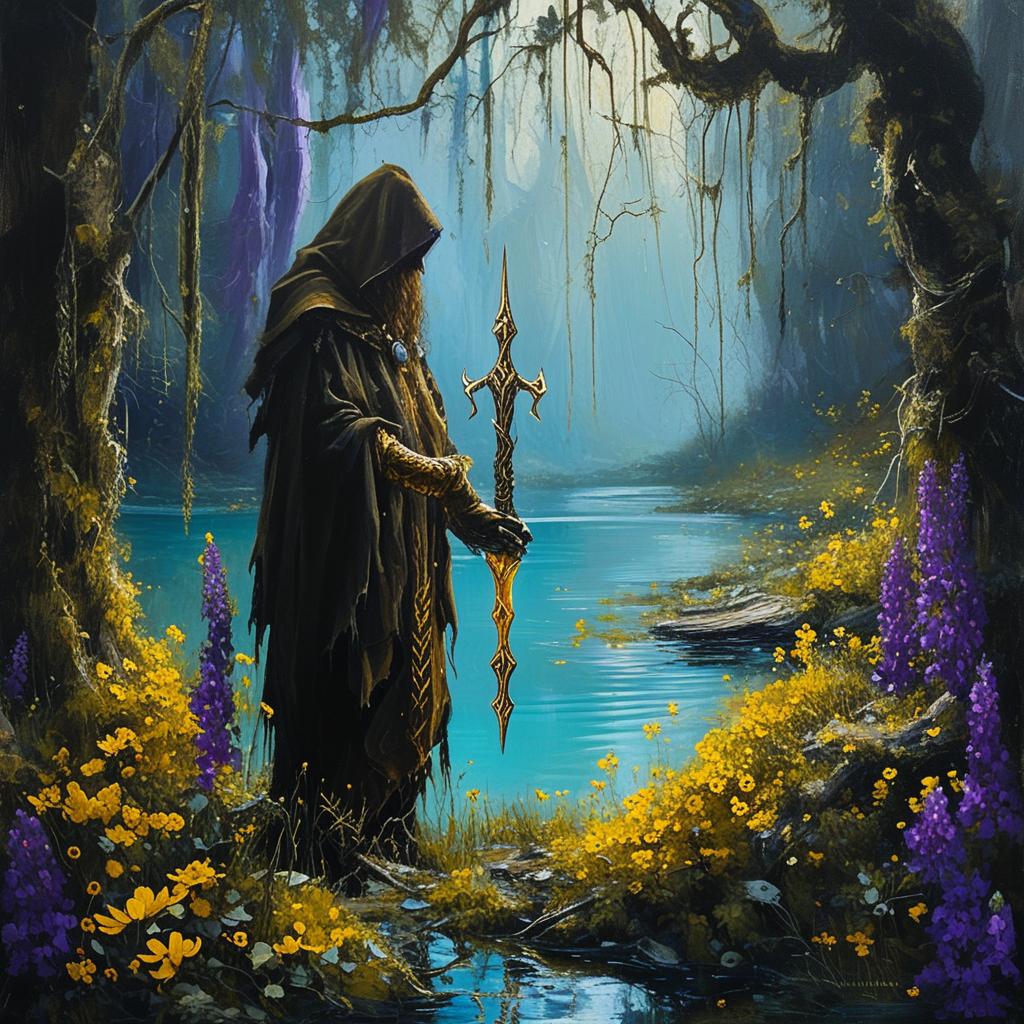}
    \end{minipage}
    \hfill
    \begin{minipage}{0.19\textwidth}
    \includegraphics[width=\textwidth]{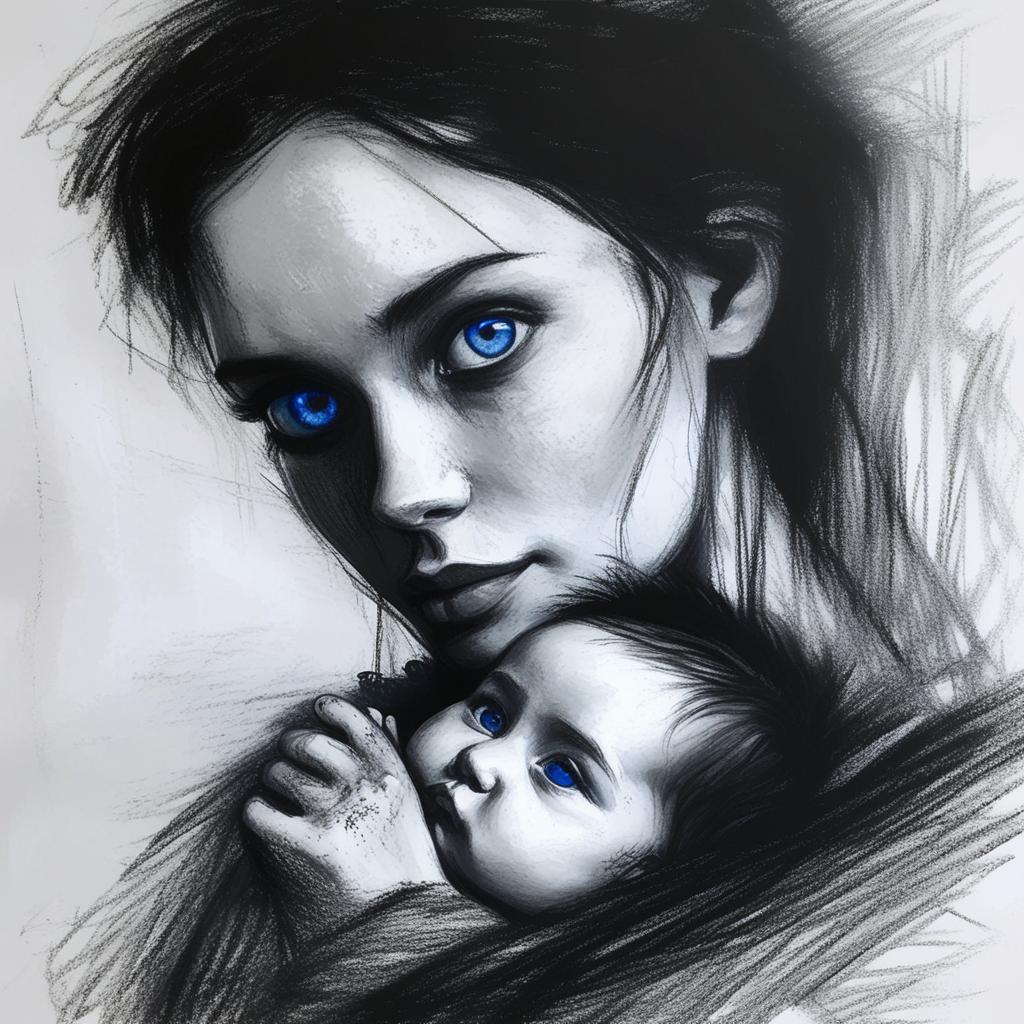}
    \end{minipage}
    \hfill
    \begin{minipage}{0.19\textwidth}
    \includegraphics[width=\textwidth]{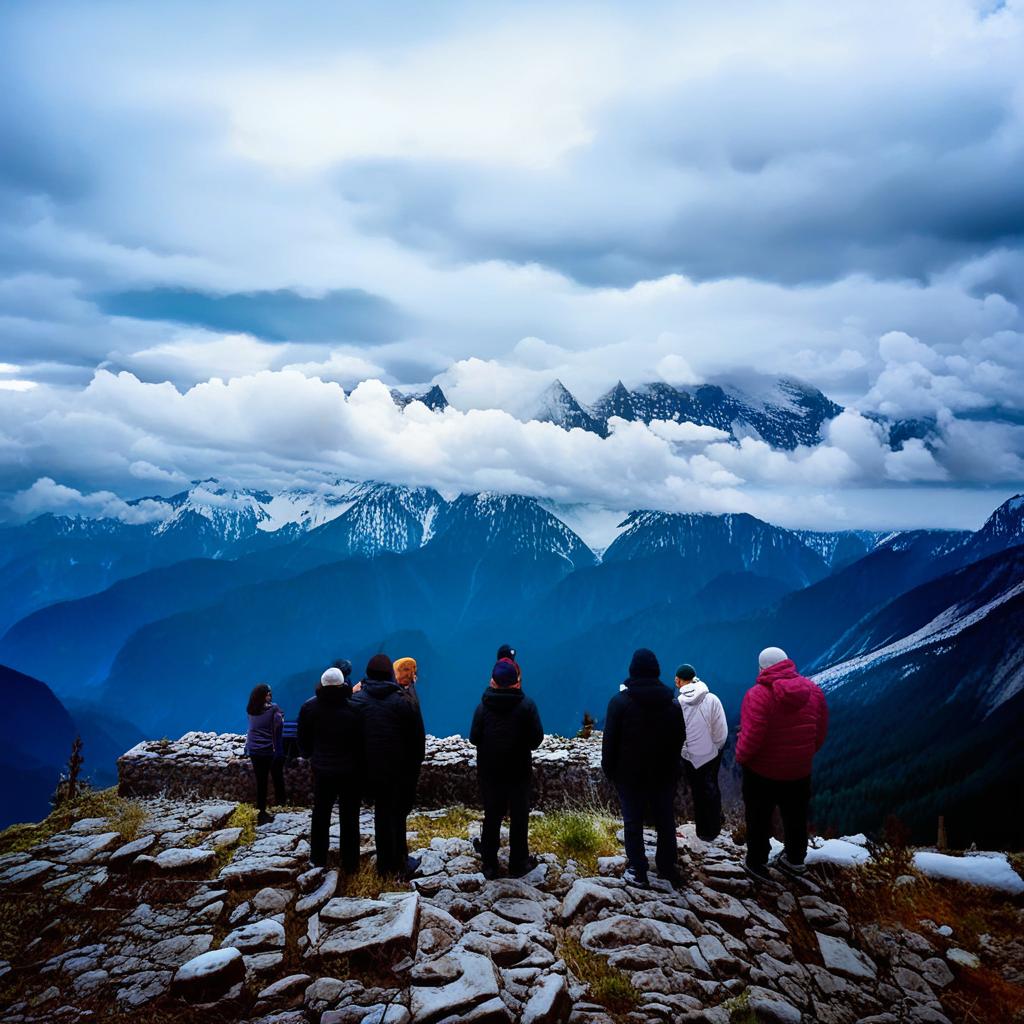}
    \end{minipage}

    \begin{minipage}{0.02\textwidth}\raggedright
    \rotatebox[origin=c]{90}{EDiT}
    \end{minipage}
    \begin{minipage}{0.19\textwidth}
    \includegraphics[width=\textwidth]{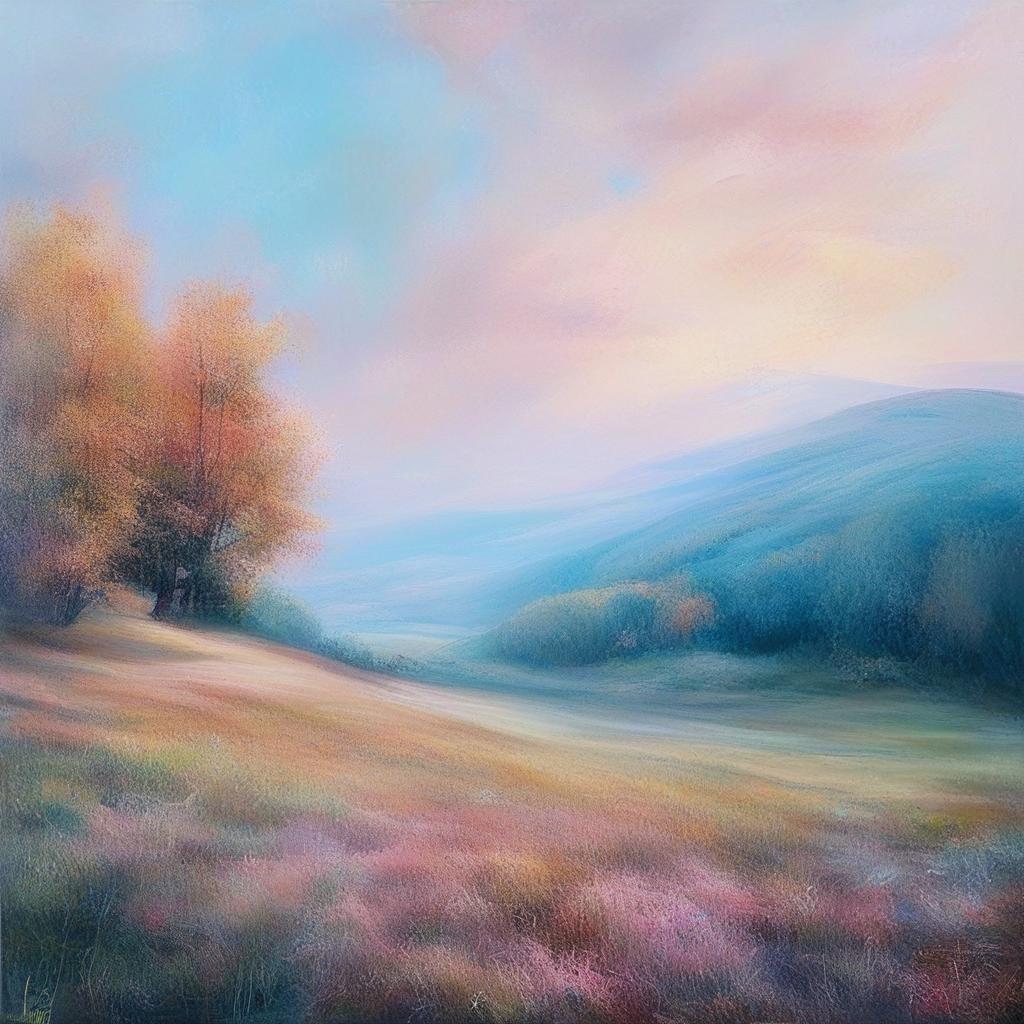}
    \end{minipage}
    \hfill
    \begin{minipage}{0.19\textwidth}
    \includegraphics[width=\textwidth]{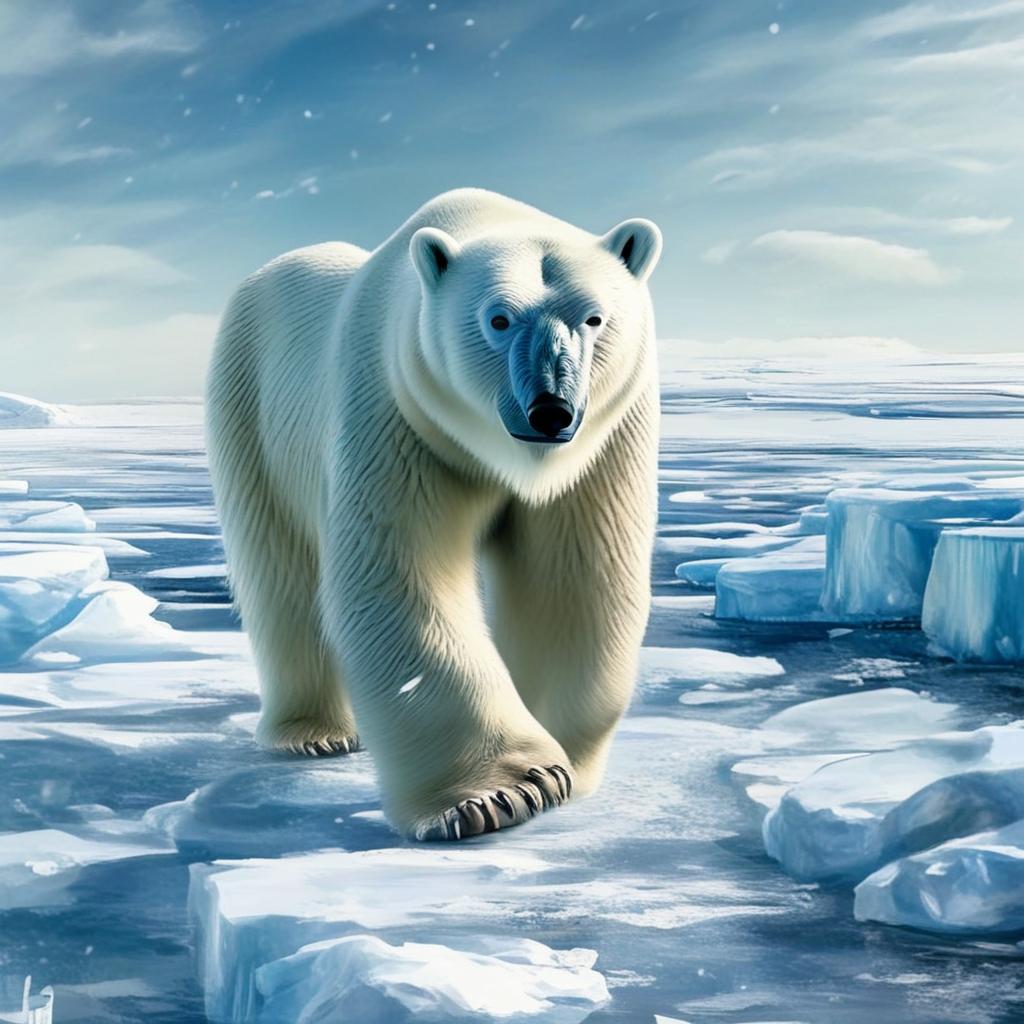}
    \end{minipage}
    \hfill
    \begin{minipage}{0.19\textwidth}
    \includegraphics[width=\textwidth]{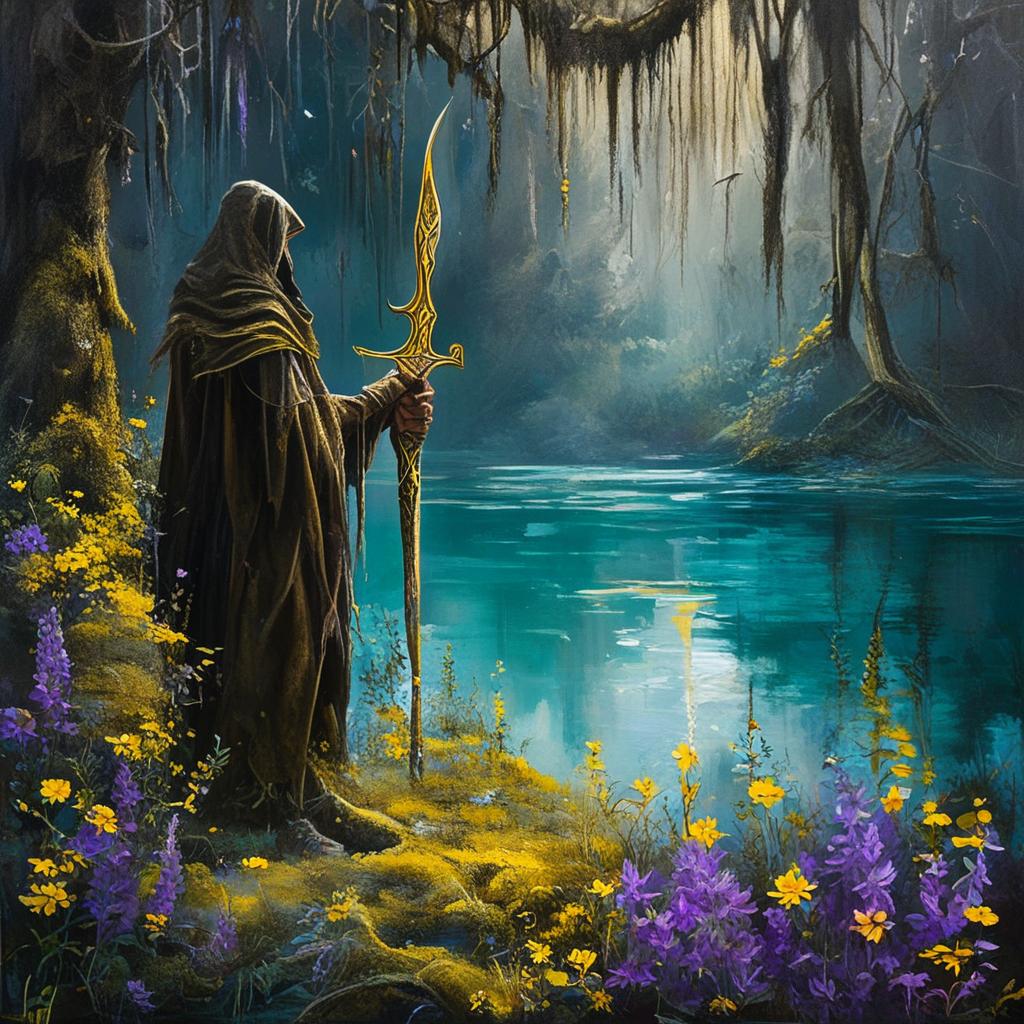}
    \end{minipage}
    \hfill
    \begin{minipage}{0.19\textwidth}
    \includegraphics[width=\textwidth]{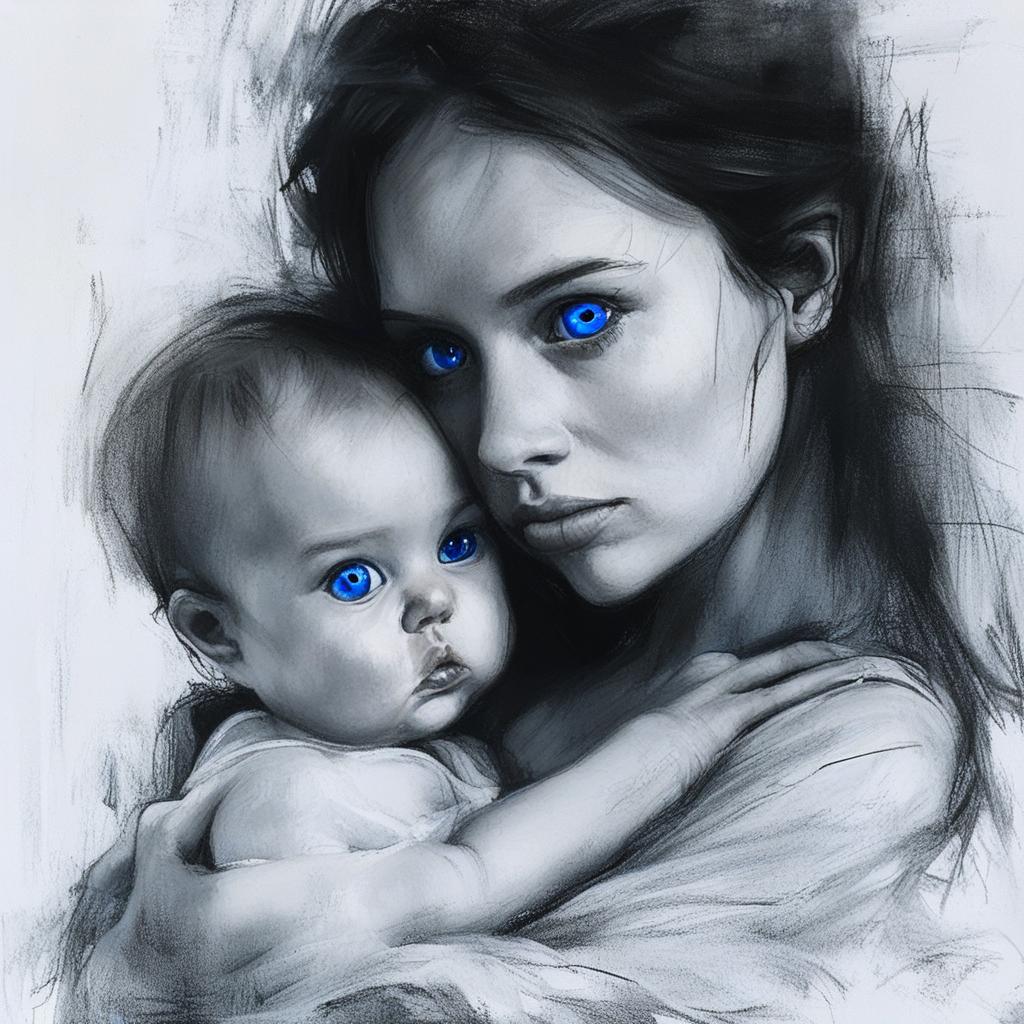}
    \end{minipage}
    \hfill
    \begin{minipage}{0.19\textwidth}
    \includegraphics[width=\textwidth]{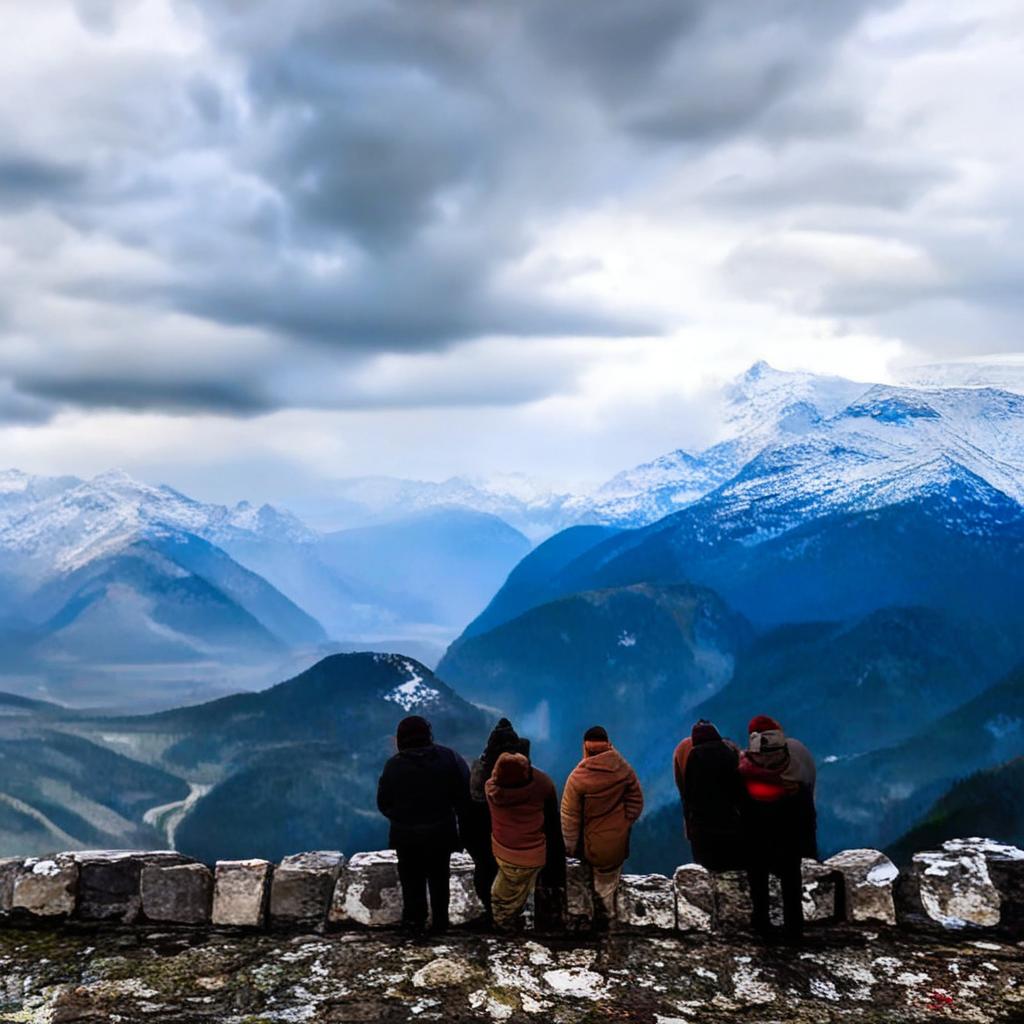}
    \end{minipage}

    \begin{minipage}{0.02\textwidth}\raggedright
    \rotatebox[origin=c]{90}{SANA-DiT}
    \end{minipage}
    \begin{minipage}{0.19\textwidth}
    \includegraphics[width=\textwidth]{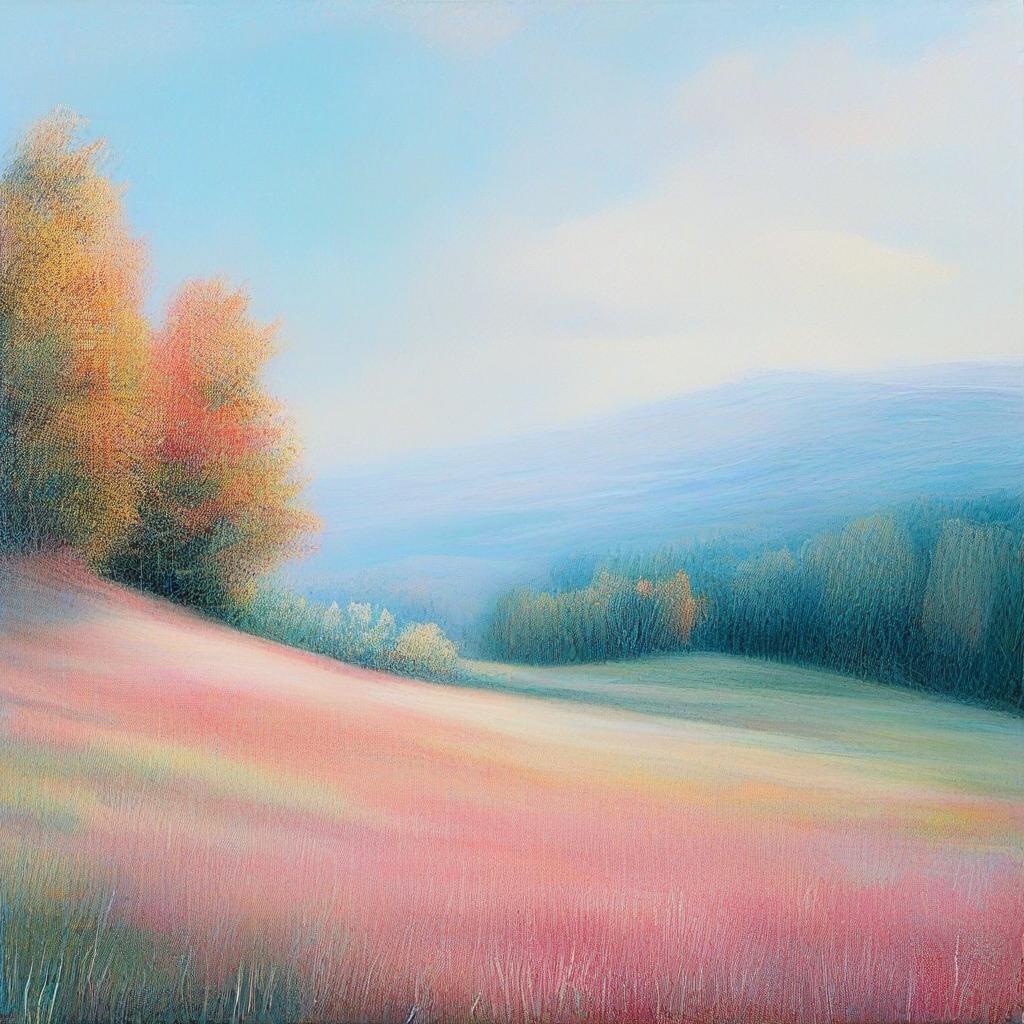}
    \end{minipage}
    \hfill
    \begin{minipage}{0.19\textwidth}
    \includegraphics[width=\textwidth]{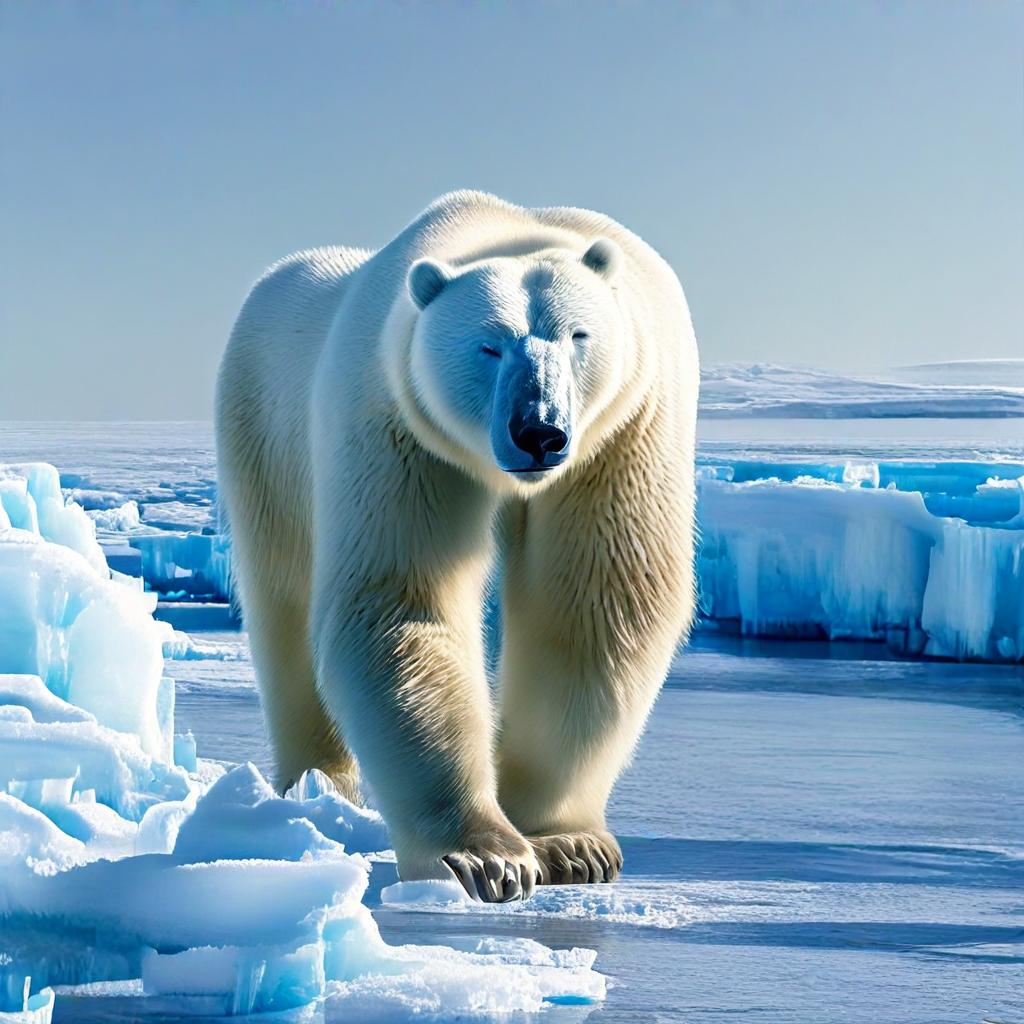}
    \end{minipage}
    \hfill
    \begin{minipage}{0.19\textwidth}
    \includegraphics[width=\textwidth]{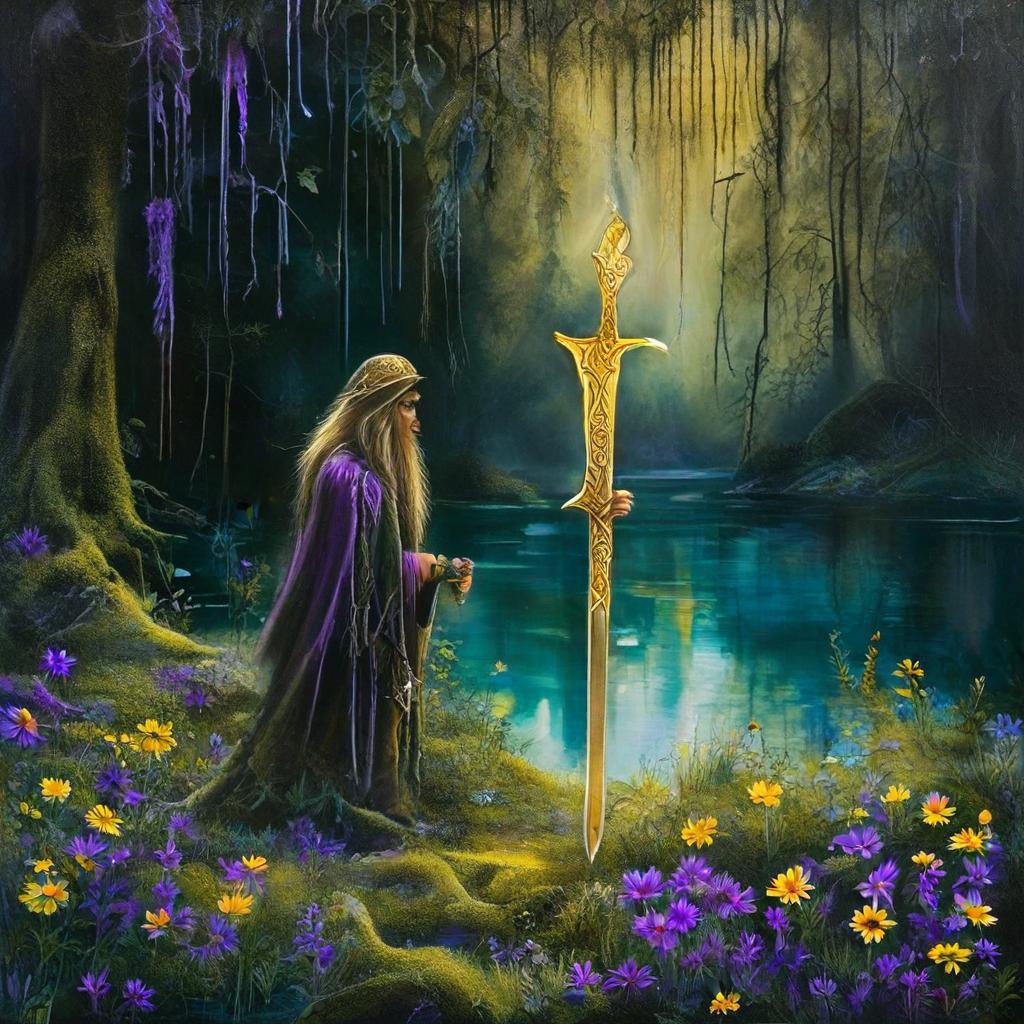}
    \end{minipage}
    \hfill
    \begin{minipage}{0.19\textwidth}
    \includegraphics[width=\textwidth]{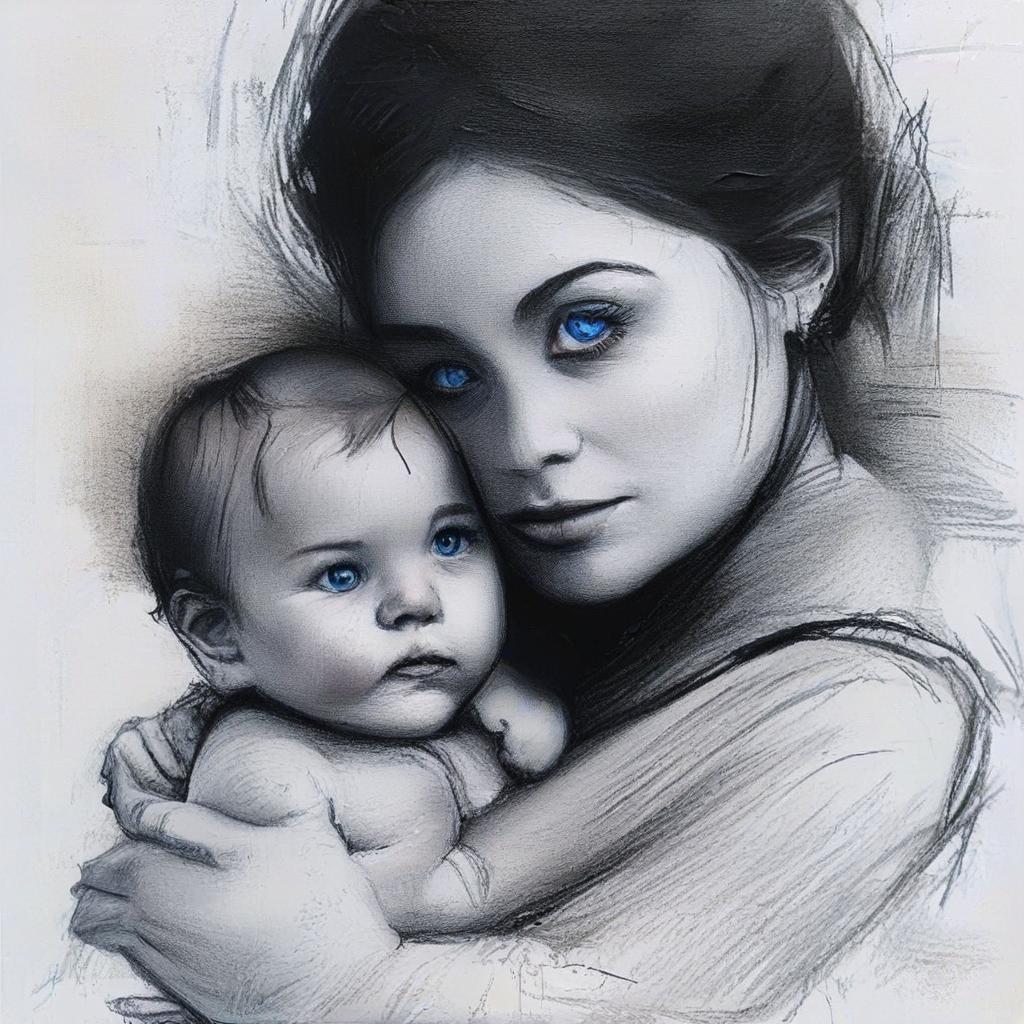}
    \end{minipage}
    \hfill
    \vspace{10pt}
    \begin{minipage}{0.19\textwidth}
    \includegraphics[width=\textwidth]{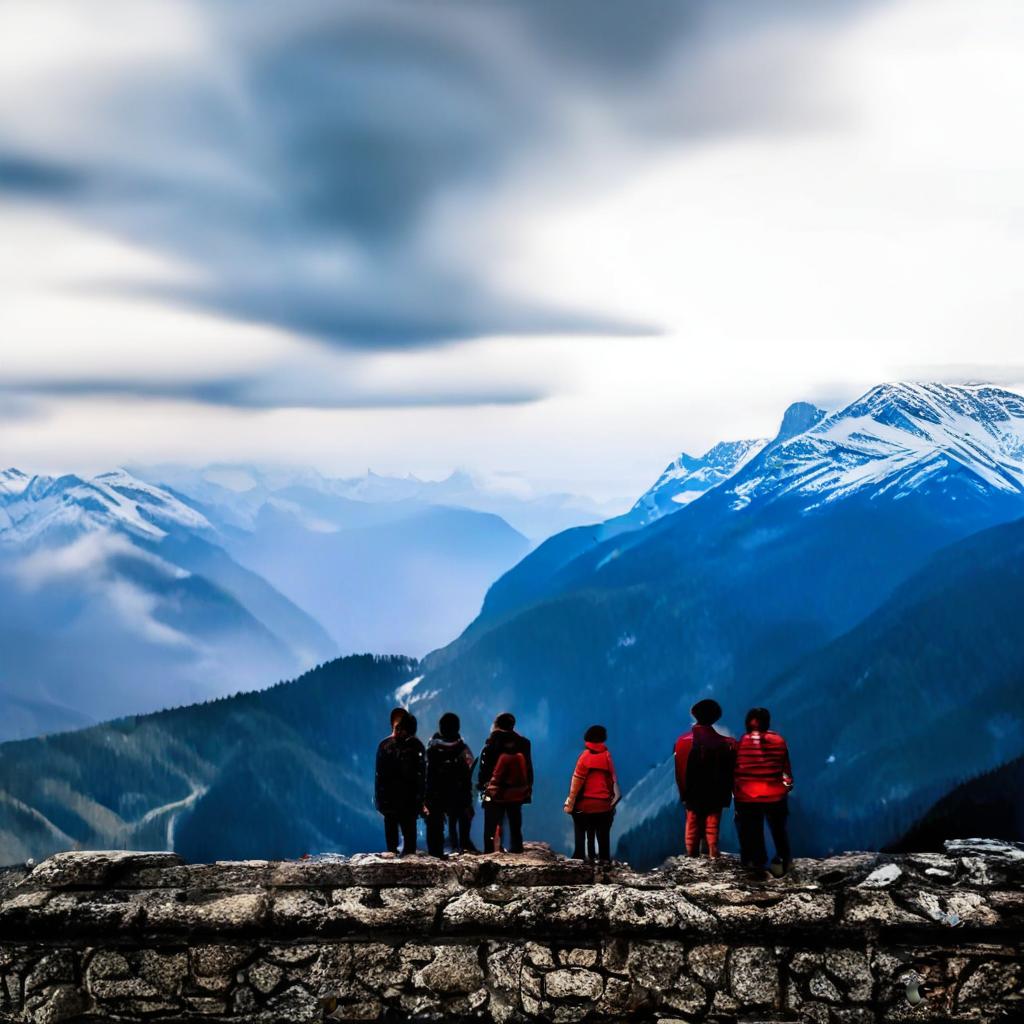}
    \end{minipage}
     \begin{minipage}{0.02\textwidth}\raggedright
    \rotatebox[origin=c]{90}{SD-v3.5M}
    \end{minipage}
    \begin{minipage}{0.19\textwidth}
    \includegraphics[width=\textwidth]{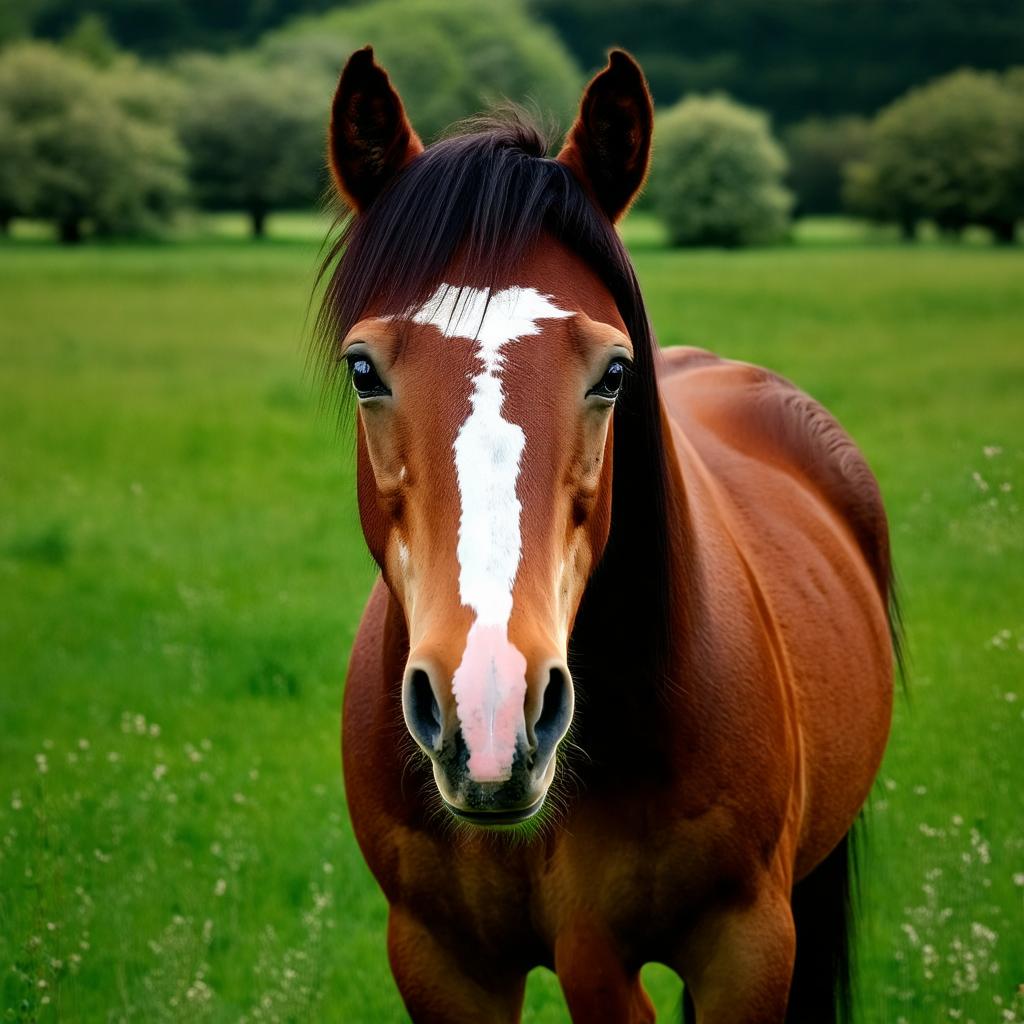}
    \end{minipage}
    \hfill
    \begin{minipage}{0.19\textwidth}
    \includegraphics[width=\textwidth]{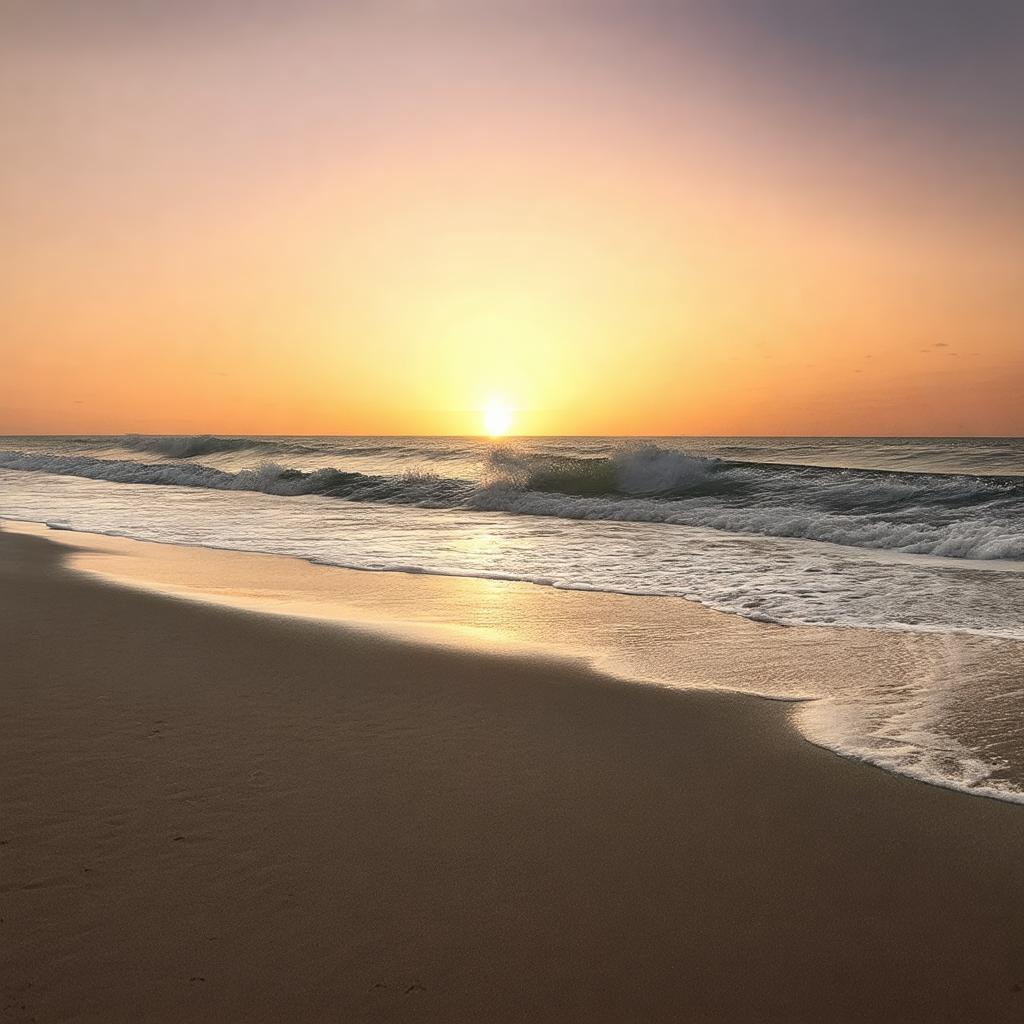}
    \end{minipage}
    \hfill
    \begin{minipage}{0.19\textwidth}
    \includegraphics[width=\textwidth]{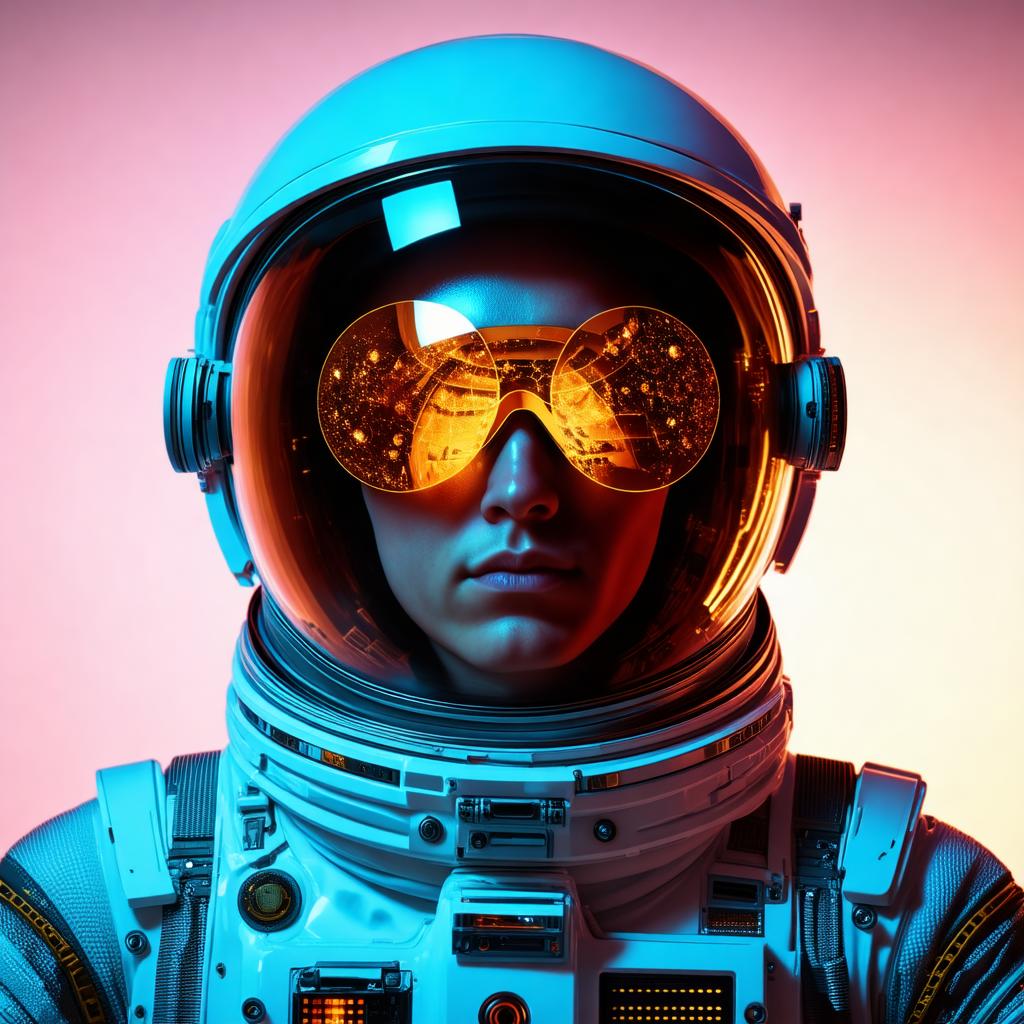}
    \end{minipage}
    \hfill
    \begin{minipage}{0.19\textwidth}
    \includegraphics[width=\textwidth]{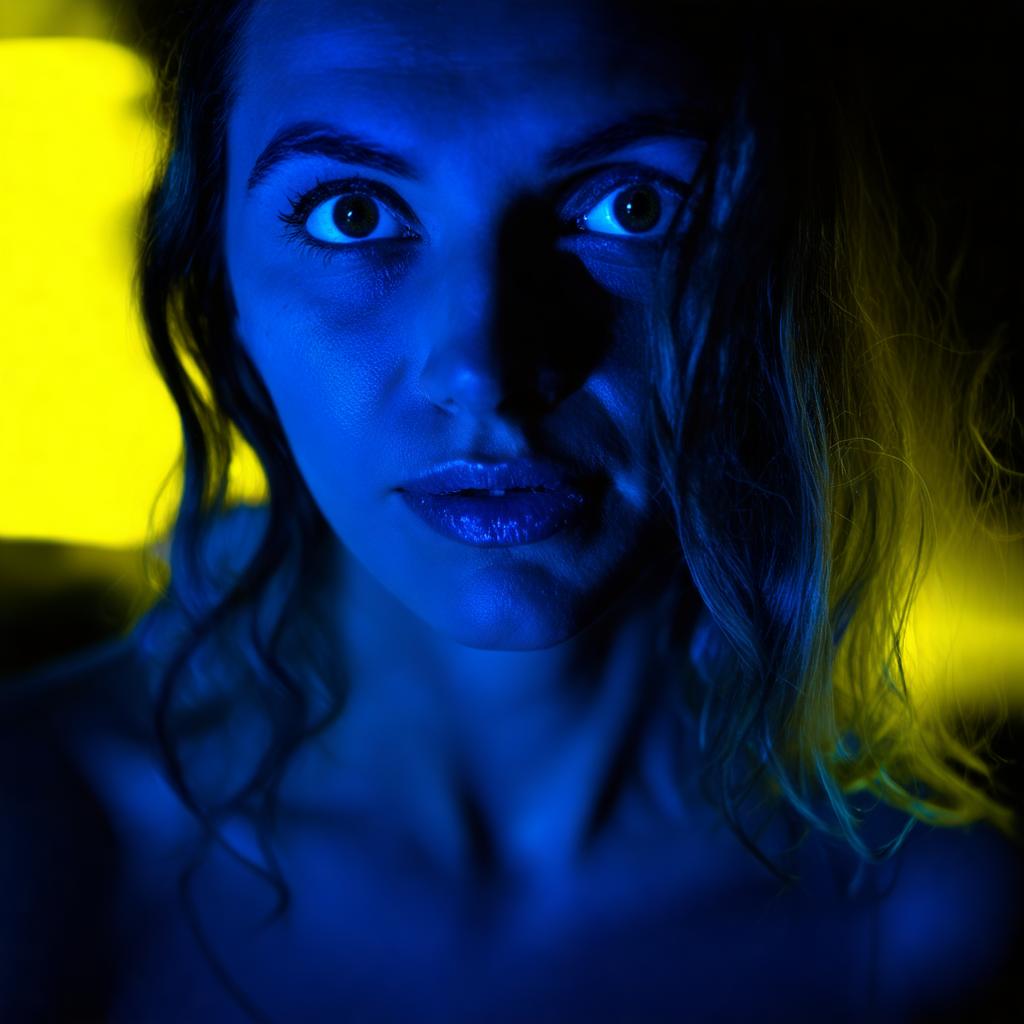}
    \end{minipage}
    \hfill
    \begin{minipage}{0.19\textwidth}
    \includegraphics[width=\textwidth]{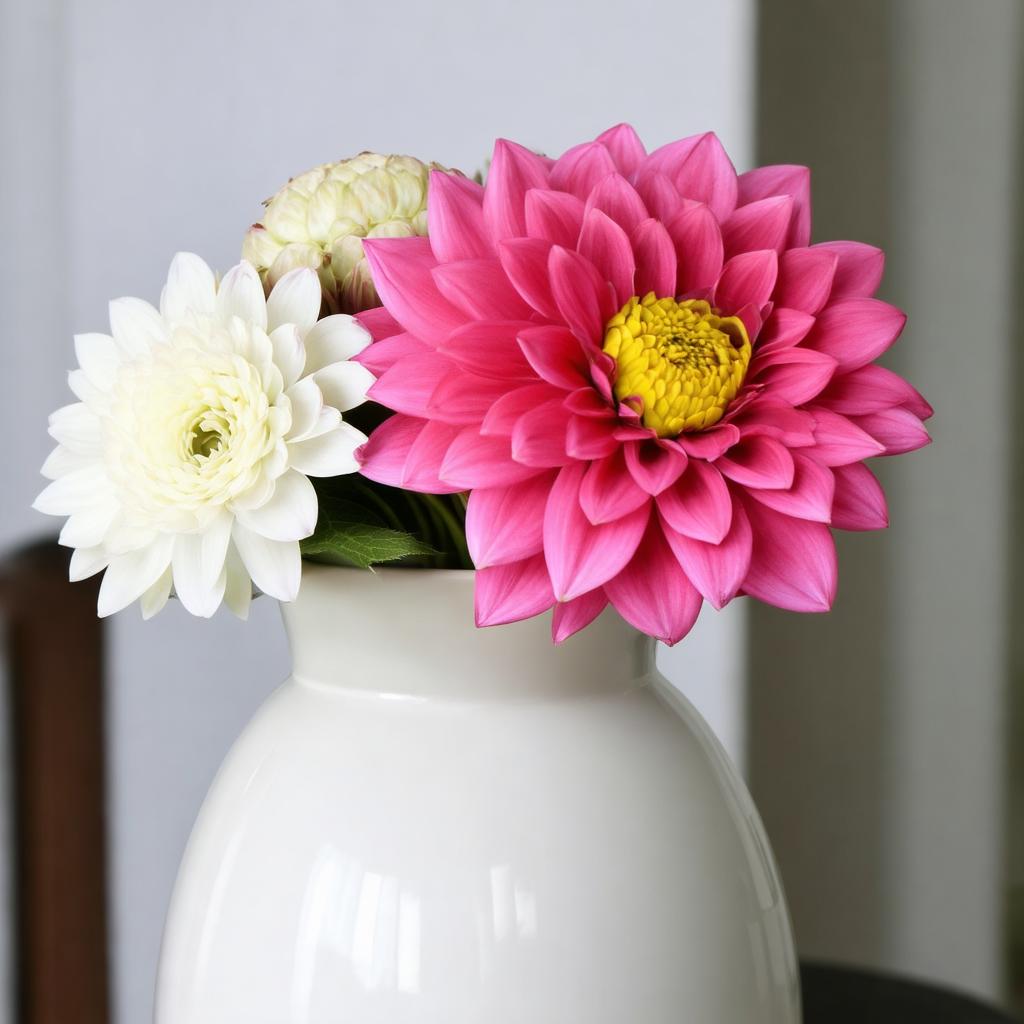}
    \end{minipage}

    \begin{minipage}{0.02\textwidth}\raggedright
     \rotatebox[origin=c]{90}{MM-EDiT}
    \end{minipage}
    \begin{minipage}{0.19\textwidth}
    \includegraphics[width=\textwidth]{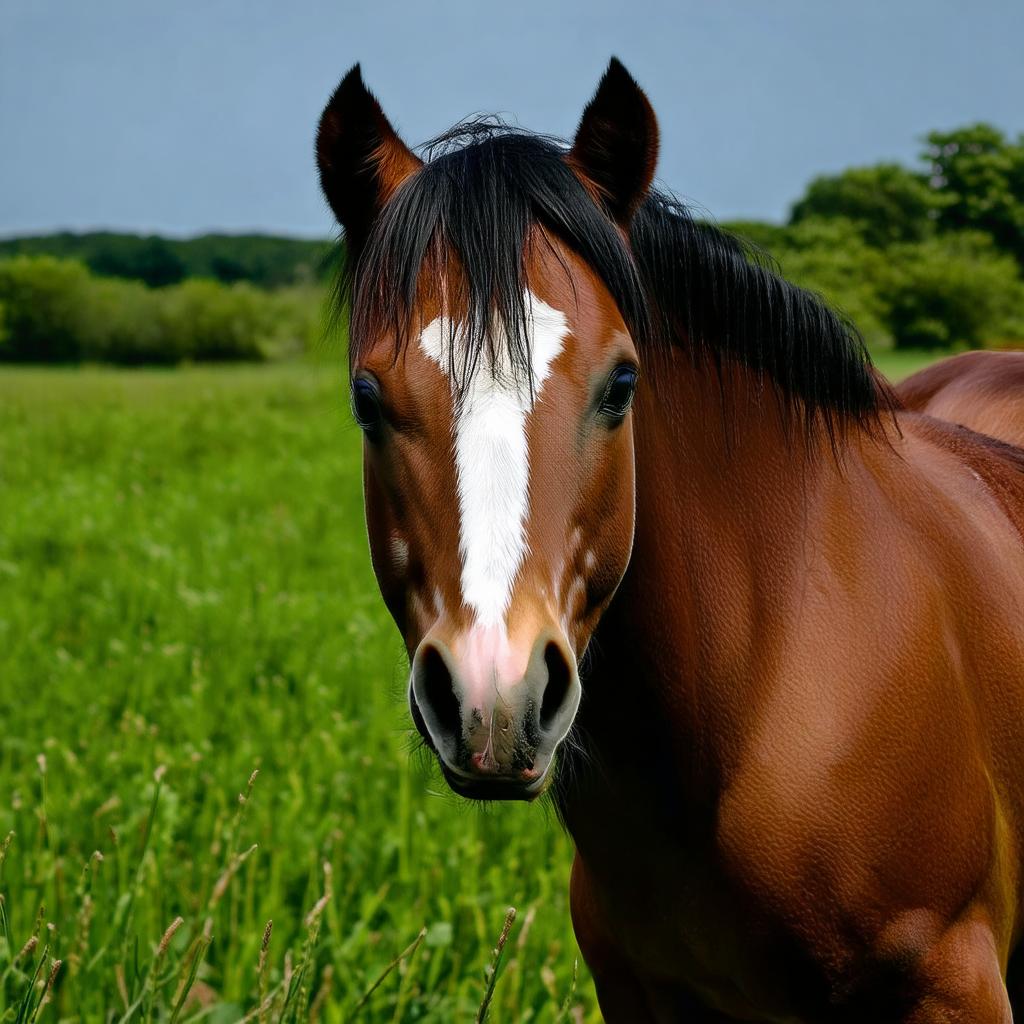}
    \end{minipage}
    \hfill
    \begin{minipage}{0.19\textwidth}
    \includegraphics[width=\textwidth]{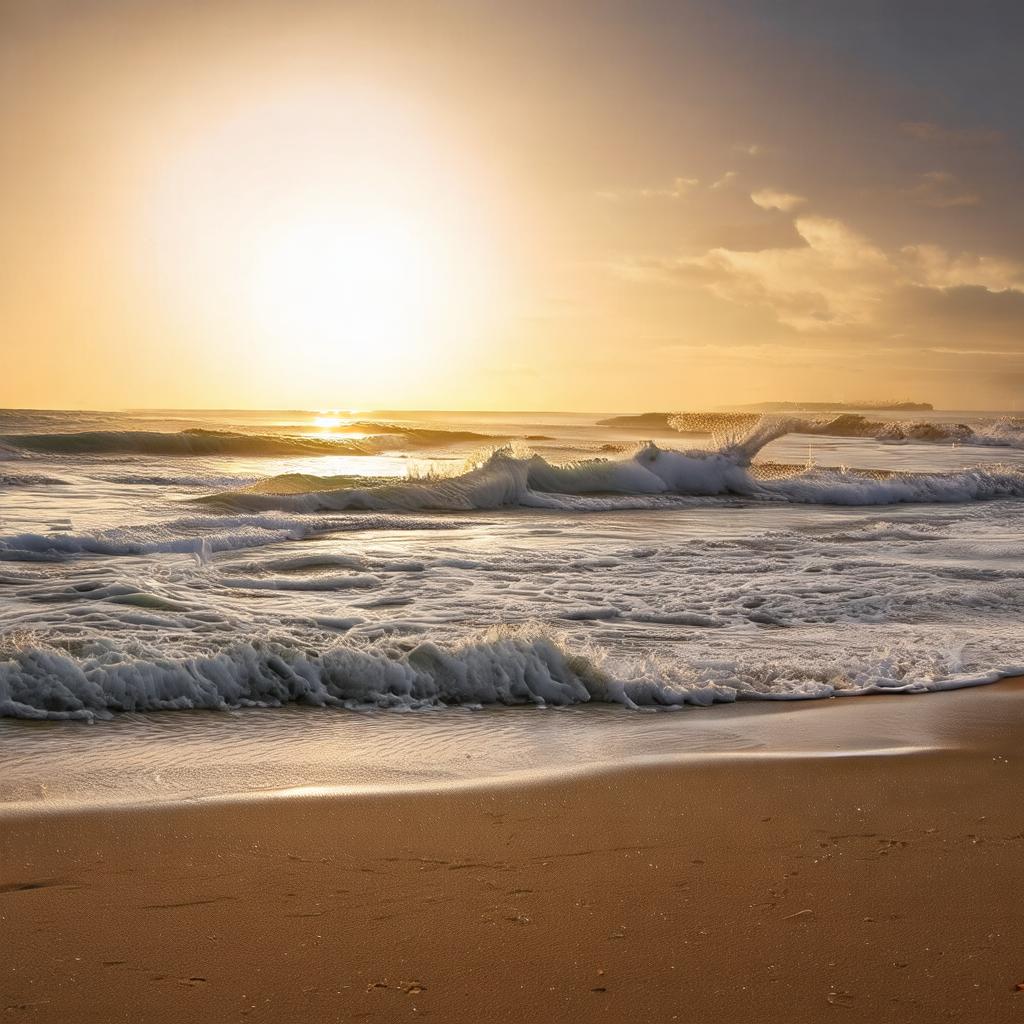}
    \end{minipage}
    \hfill
    \begin{minipage}{0.19\textwidth}
    \includegraphics[width=\textwidth]{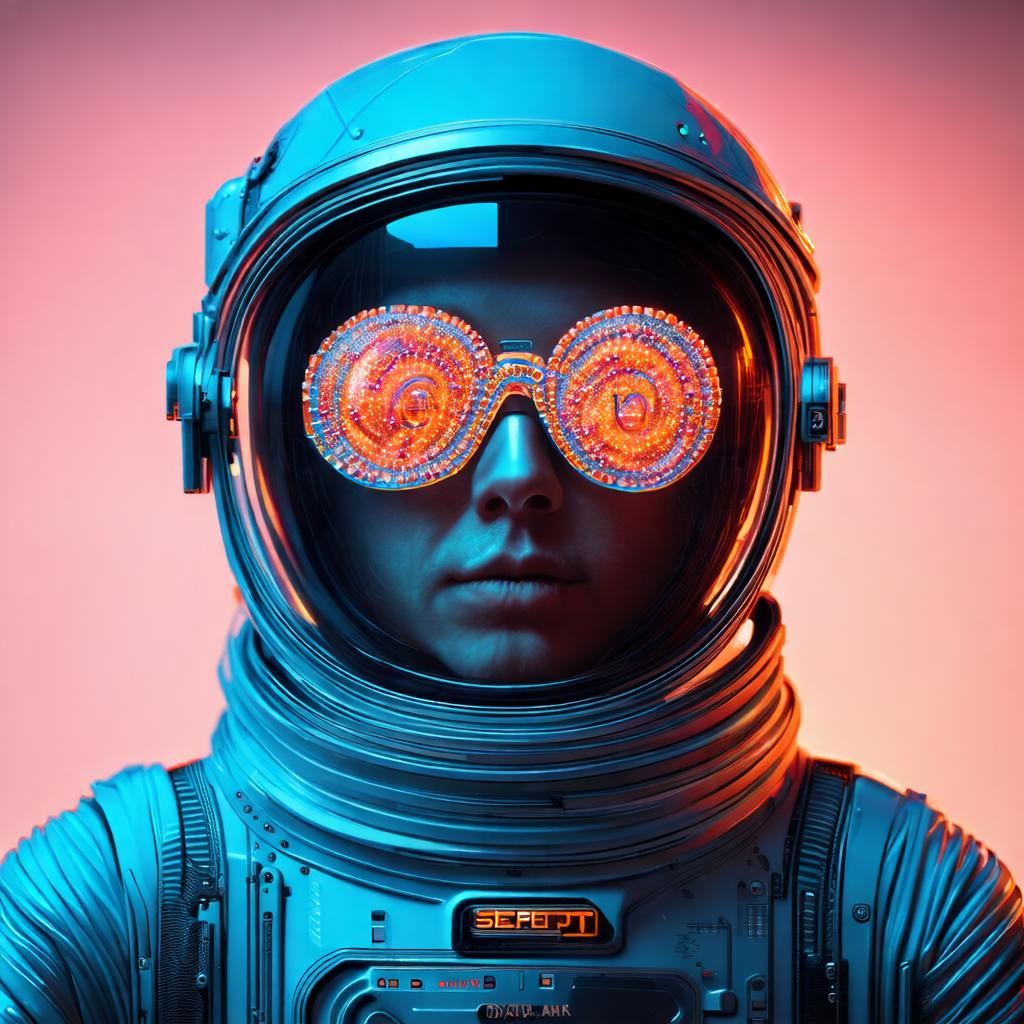}
    \end{minipage}
    \hfill
    \begin{minipage}{0.19\textwidth}
    \includegraphics[width=\textwidth]{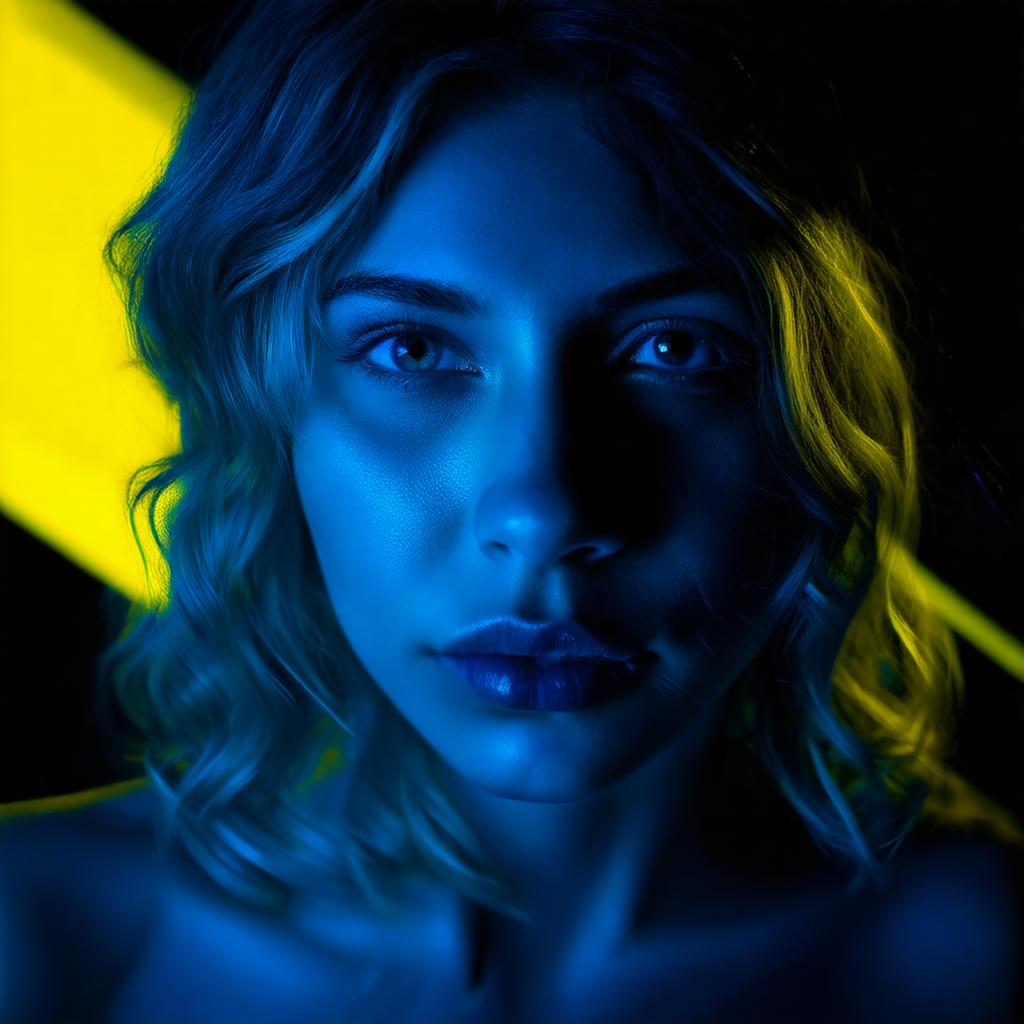}
    \end{minipage}
    \hfill
    \begin{minipage}{0.19\textwidth}
    \includegraphics[width=\textwidth]{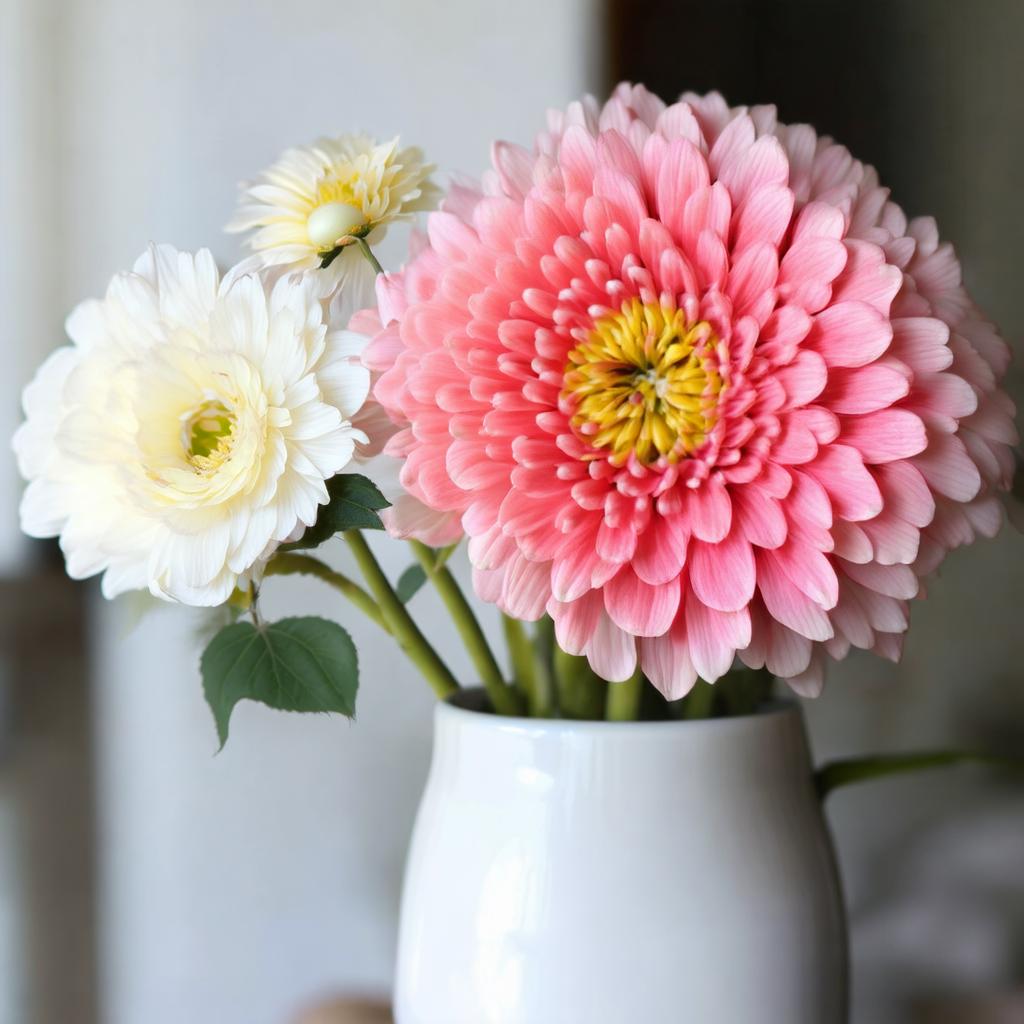}
    \end{minipage}

    \begin{minipage}{0.02\textwidth}\raggedright
       \rotatebox[origin=c]{90}{SANA-MMDiT}
    \end{minipage}
    \begin{minipage}{0.19\textwidth}
    \includegraphics[width=\textwidth]{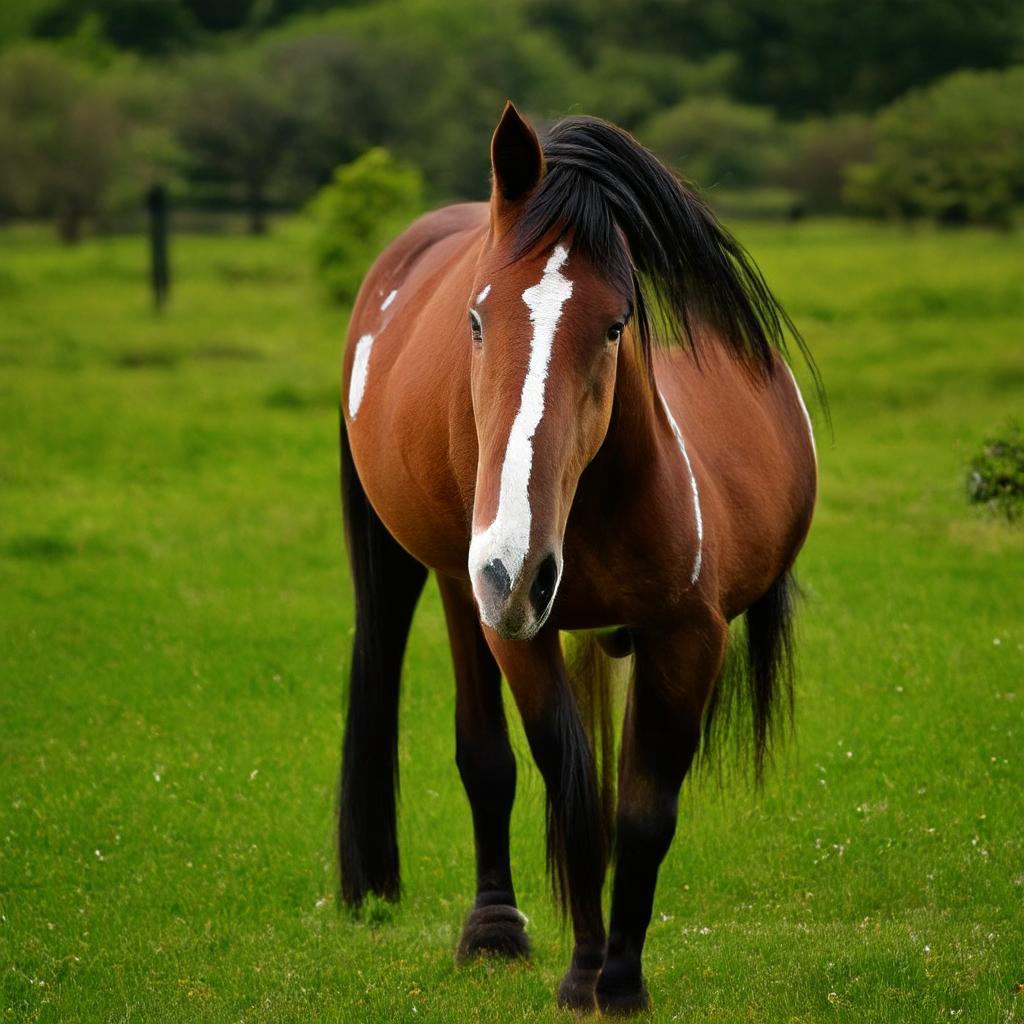}
    \end{minipage}
    \hfill
    \begin{minipage}{0.19\textwidth}
    \includegraphics[width=\textwidth]{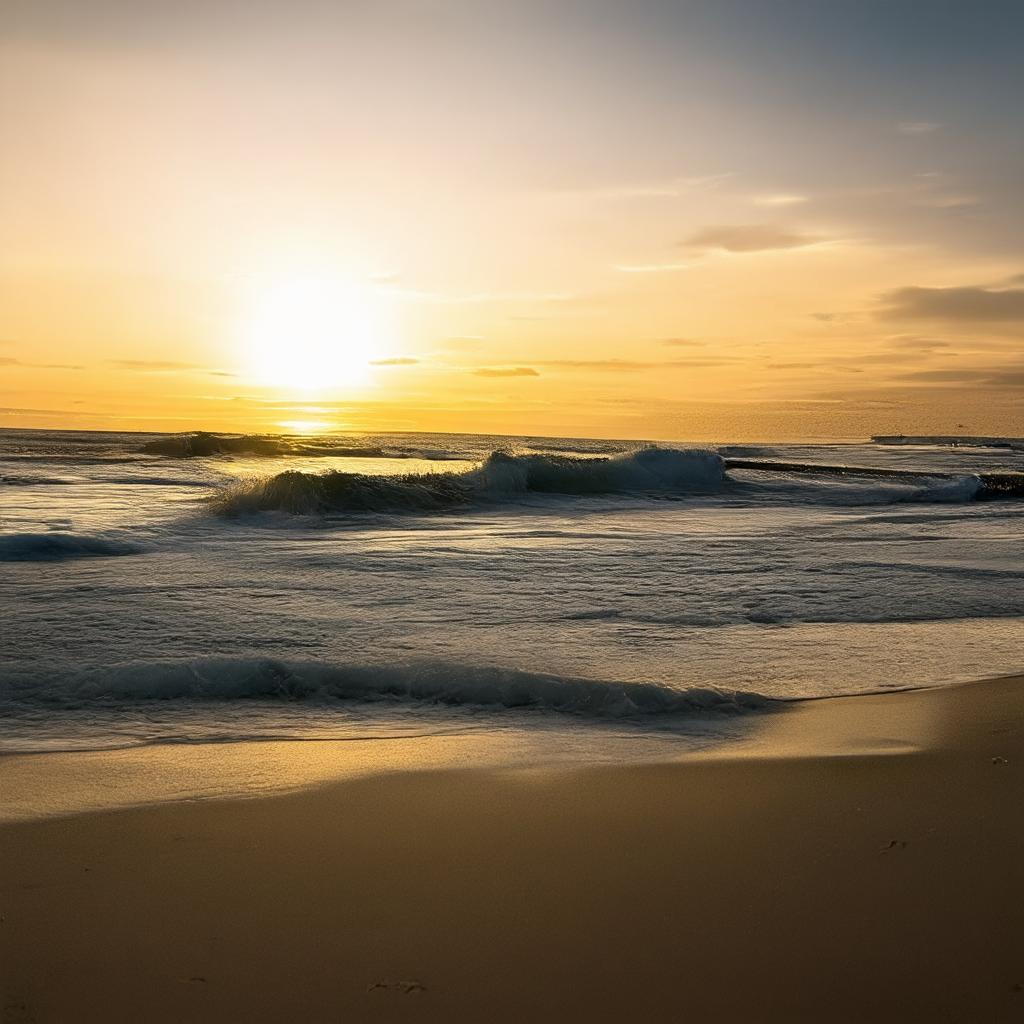}
    \end{minipage}
    \hfill
    \begin{minipage}{0.19\textwidth}
    \includegraphics[width=\textwidth]{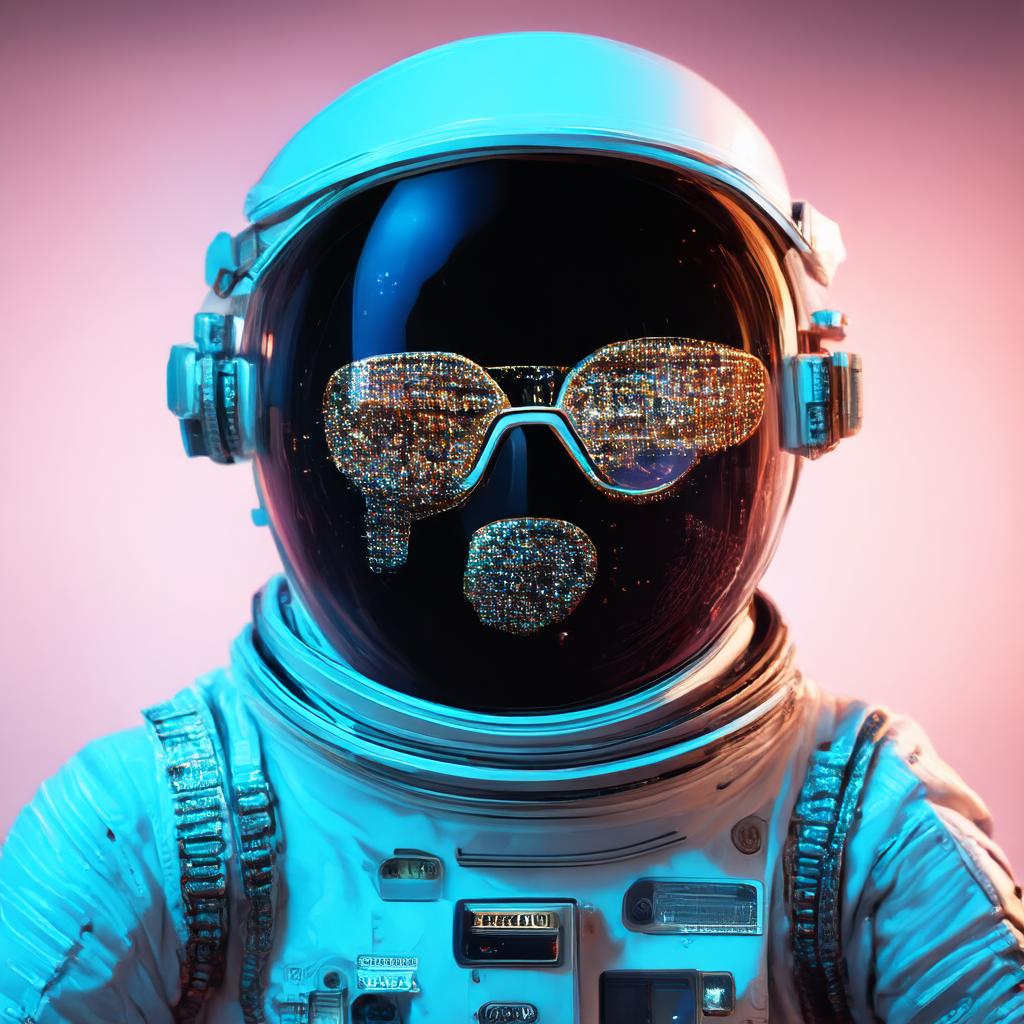}
    \end{minipage}
    \hfill
    \begin{minipage}{0.19\textwidth}
    \includegraphics[width=\textwidth]{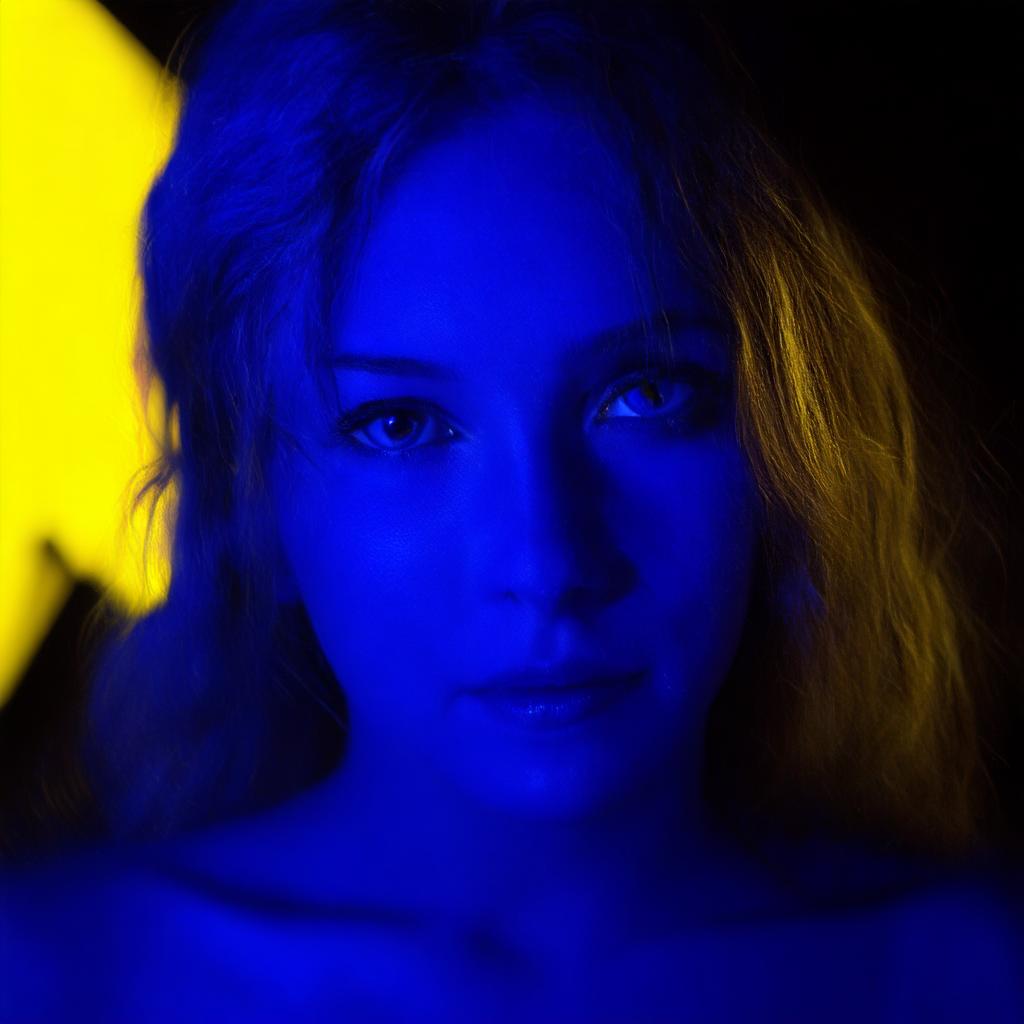}
    \end{minipage}
    \hfill
    \begin{minipage}{0.19\textwidth}
    \includegraphics[width=\textwidth]{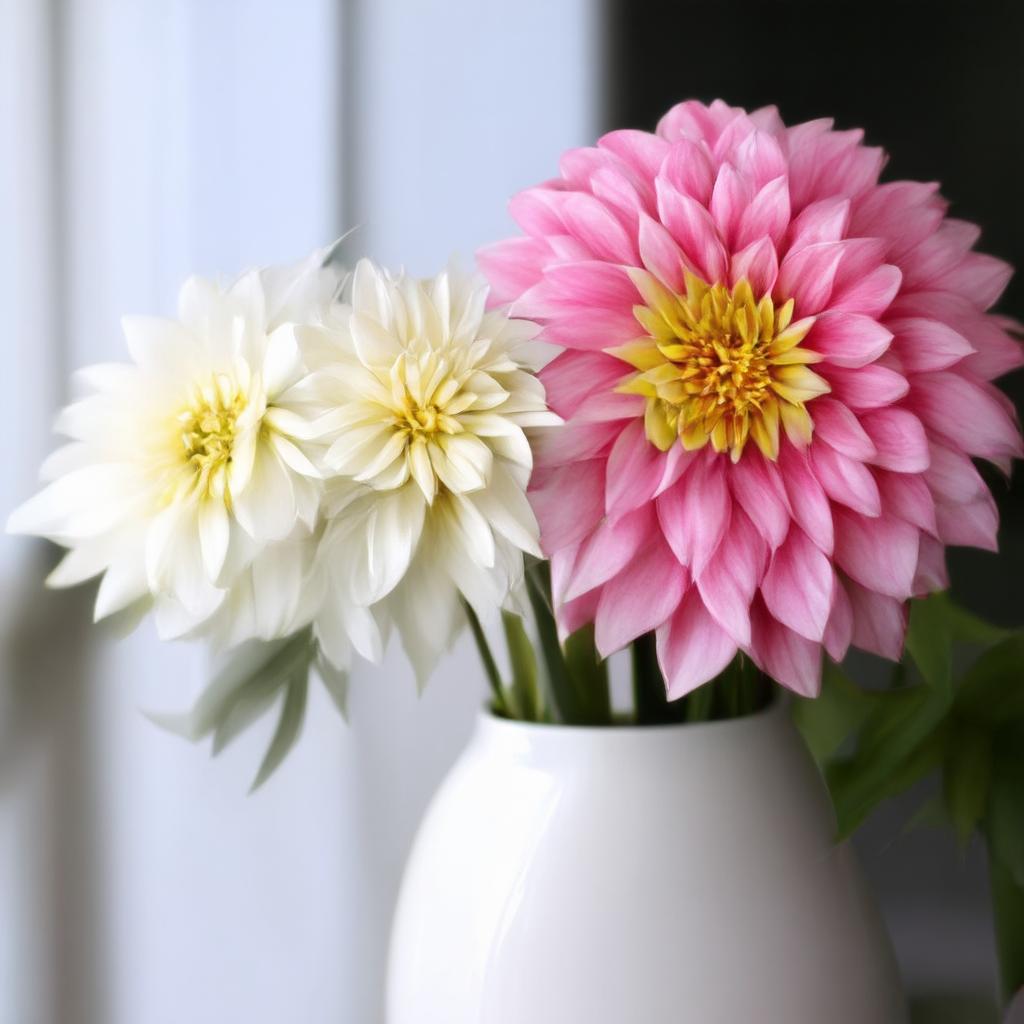}
    \end{minipage}
    
    \caption{Qualitative results for EDiT and MM-EDiT models at a resolution of $1024 \times 1024$ compared to base models Pixart-$\Sigma$ and SD-v3.5M, and SANA-based SANA-DiT and SANA-MMDiT respectively. The prompts used to generate these images and results for all other baselines and ablations can be found in Appendix \ref{app:sec:cap}.}
    \label{fig:edit_vis}
\end{figure*}

%% file: sec/5_related_work.tex
\section{Related Work}

Beyond, SANA~\cite{xie2024sana} and LinFusion~\cite{liu2024linfusion}, which we previously discussed in Section~\ref{sec:background}, several other works use linear attention in DiTs or enhance their efficiency by other means.
SANA 1.5 \cite{xie2025sana15} added RMS normalization to Queries and Keys for efficient scaling of linear attention from 1.6B to 4.8B while maintaining efficiency and training stability.
ZIGMA~\cite{hu2024zigma} leverages a specialized form of causal linear attention known as Mamba~\cite{dao2024transformers} to develop efficient diffusion models for text and video generation. However, the causal nature of Mamba introduces complexity in the DiT framework, necessitating the rearrangement of tokens after each block.
DIMBA~\cite{fei2024dimba} also incorporates Mamba into a DiT but does so in conjunction with existing self-attention blocks. While this approach yields slight performance improvements, it also increases computational demands, resulting in a quadratic complexity.
Notably, none of these works discussed so far have addressed MM-DiTs.

\noindent
To the best of our knowledge 
CLEAR~\cite{liu2024clear}  is the only existing linear-time approach to MM-DiTs. 
However, they use sparse neighborhood attention~\cite{hassani2023neighborhood}, which requires intricate memory access patterns and often only results in practical efficiency gains for very high-resolution images ( $>  2048 \times 2048$ pixels). Further, it relies on specialized implementations, limiting its applicability to different hardware.
More broadly, linear DiTs can be combined with masking approaches to increase training efficiency as done for vanilla DiTs in PatchMixer \cite{Sehwag2025patchmixer}, which is a promising research direction.

%% file: sec/6_conclusion.tex
\section{Conclusion}

In this paper, we addressed the computational bottleneck of Diffusion Transformers (DiTs) by introducing the Efficient Diffusion Transformer (EDiT) with a linear compressed attention mechanism that integrates local information in queries and aggregates token information for keys and values. We demonstrated that distilling PixArt-$\Sigma$ with EDiT achieves substantial speedups without compromising image quality. Additionally, we extended this approach to Multimodal DiTs by proposing a hybrid attention mechanism, resulting in the Multimodal Efficient Diffusion Transformer (MM-EDiT), which combines EDiT's linear attention for image interactions with standard attention for prompts. Distilling Stable Diffusion 3.5-Medium with MM-EDiT maintained performance while achieving linear time complexity.

%% file: appendix/limitations.tex
\section{Limitations}
Our evaluations were primarily conducted on PixArt-$\Sigma$ and Stable Diffusion 3.5-Medium. Future work should investigate the generalizability of our approach to larger models such as Flux~\cite{flux2024}. Additionally, while quantitative metrics were used, the qualitative assessment of generated images remains subjective, and human evaluation could provide further insights into image quality.

%% file: appendix/runtime.tex
\section{Additional Runtime Results}
\label{sec:add_rt}
\begin{table*}[t]
\begin{minipage}[t]{0.5\textwidth}
\centering
{\small
\begin{tabular}{l|c}
    Model & Latency (s) \\
    \hline
    PixArt-$\Sigma$  & 5.44  \\ 
    \textbf{EDiT} (ours)  & 4.15  \\
    \hline
    SANA-DiT & 4.0 \\ 
    LinFusion-DiT  & 4.2 \\ 
    KV Comp. ($k=2$) & 5.0  \\ 
    \hline
    \multirow{4}{*}{\rotatebox[origin=c
    ]{90}{Ablations} \begin{tabular}{ccc}
        Q & K & V  \\
        \hline
        CF & CF & - \\
        - & SC & SC \\
        CF & - & -
    \end{tabular}}
     & \\ 
    & 4.55  \\ 
    & 3.8 \\ 
    & 4.3  \\ 
\end{tabular}   
}
\end{minipage}
\begin{minipage}[t]{0.5\textwidth}%
\centering
{\small
\begin{tabular}{l|c}
    Model & Latency (s) \\
    \hline
    SD-v3.5M & 10.9 \\ 
    \textbf{MM-EDiT} (Ours) & 8.0  \\
    \hline
    MM-EDiT with $\eta^{Lin}$  & 8.6 \\
    MM-EDiT no $\phi_{CF}$ and $\phi_{SC}$  &  7.6 \\
    \hline
    SANA-MM-DiT & 7.6 \\
    Linear MM-DiT-$\alpha$ & 7.7 \\
    Linear MM-DiT-$\beta$ & 7.75 \\
\end{tabular}   
}
\end{minipage}
\caption{Latency of generating a single $1024 \times 1024$ pixel image on a consumer-grade Nvidia 3090 RTX.
PixArt-$\Sigma$, EDiT, and all the respective ablations use 20 diffusion steps.
SD-v3.5M, MM-EDiT, and all the respective ablations use 28 diffusion steps. 
These results further emphasize that, while producing images of similar quality to their respective teachers, EDiT and MM-EDiT improve efficiency at the practically relevant resolution of $1024 \times 1024$ and strike the best compromise of runtime and quality across all considered variants.
}
\label{sup:tab:rt}
\end{table*}

Table~\ref{sup:tab:rt} shows additional runtime results for our models, the teachers, and all baselines and ablations for the relevant use-case of generating $1024 \times 1024$ images on consumer-grade hardware. 
 

%% file: appendix/finetuning.tex
\section{Finetuning}
\label{sec:ft}
\begin{table*}[]
    \centering
    \small{
    \begin{tabular}{l||cc|cc}
    
     &  FID & FID &   FID & FID \\
     & (Inception-v3) & CLIP & (Inception-v3) & CLIP\\
    \hline
    &  \multicolumn{2}{c|}{$512 \times 512$} &  \multicolumn{2}{c}{$1024 \times 1024$}  \\
     \hline
     PixArt-$\Sigma$ & 7.49 & 2.52 & 8.55 & 3.17 \\
     EDiT (before finetuing) &  7.36 & 2.53 & 10.1 & 3.42 \\
     EDiT (after finetuing) &  5.17 & 1.84 & 4.72 & 1.39 \\
     \hline
     SD-v3.5 &  11.5 & 4.41 & 8.91 & 5.07\\
     MM-EDiT (before finetuing) &  8.43 & 2.81 & 11.6 & 5.28\\
     MM-EDiT (after finetuing) &  4.46 & 1.69 & 7.15 & 2.89\\
     \bottomrule
    \end{tabular}
    }
    \caption{Quantitative comparison of fine-tuning EDiT and MM-EDiT to fine-tuned pre-trained models. The results show that EDiT amd MM-EDiT outperform fine-tuned baselines.}
    \label{tab:finetuning}
\end{table*}

\noindent
To show that our models are amenable to standard fine-tuning, we generate a high-quality synthetic dataset by generating $\approx 960,000$ images with Flux[Schnell]~\cite{flux2024} re-using the YE-POP captions. 
From these, we use $\approx 24,000$ for validation and testing each. 
We then fine-tune our best distilled EDiT and MM-EDiT checkpoints using the remaining images while selecting the model using the validation set. 

\noindent
The results in Table~\ref{tab:finetuning} show that fine-tuning, as expected, improves FID on the test set, indicating that our models can not only be trained by disitillation but also using the standard diffusion losses of the teachers.  

%% file: appendix/captions.tex
\section{Prompts and Additional Qualitative Results}
\label{app:sec:cap}

\input{figures/pixart_images_1k/_pixart_quali_supp}
\input{figures/sd35_images_1k/_sd35_quali_sup}

The prompts used to generate the images for the $1024 \times 1024$ pixel images generated with EDiT, PixArt-$\Sigma$ and corresponding baselines and ablations. (from left to right): 
\begin{enumerate}
    \item  "Pastel landscape, nature, detail"
    \item "A polar bear walking through icy and snowy terrain" 
    \item "Painting of a wild mystical druid holding a shining golden etched dagger standing beside a crystal clear lake, woodland forest surrounded by yellow and purple coloured wild flowers, hanging moss and vines, moody lighting"
    \item "A soft black and white charcoal sketch of a beautiful young mother holding her baby, blue eyes with dark circles around the irises, old canvas, minimalist, blocky, gritty"
    \item "The image captures a group of people standing on a stone wall, overlooking a breathtaking mountainous landscape. The sky above them is a dramatic blend of blue and white, suggesting an overcast day. The mountains in the distance are blanketed in snow, adding a serene beauty to the scene. The perspective of the image is from behind the group, looking out towards the mountains, giving a sense of depth and scale to the landscape. The colors in the image are muted, with the blue of the sky and the white of the snow standing out against the darker tones of the mountains and the stone wall"    
    \end{enumerate}

\noindent
The prompts used to generate the images for the $2048 \times 2048$ pixel images with EDiT and PixArt-$\Sigma$
\begin{enumerate}
    \item  Realistic portrait pet portrait
    \item  older adult alien with feathers, no hair, in vertical striped pyjamas, white background, small ears, wearing a scarf
\end{enumerate}

\noindent
The prompts used to generate the images for the $1024 \times 1024$ pixel images generated with MM-EDiT, SD-v3.5 and corresponding baselines and ablations. (from left to right): 
\begin{enumerate}
    \item "In the heart of a verdant field, a majestic brown and white horse stands, its gaze meeting the camera with an air of quiet confidence. The horse's coat is a rich brown, adorned with white spots that add a touch of whimsy to its appearance. A distinctive white stripe runs down its nose, adding to its unique charm. Its mane and tail, both a deep black, contrast beautifully with its coat. The horse's eyes, full of life and curiosity, seem to be looking directly at the camera, creating a sense of connection between the viewer and the subject. The background is a lush green field, dotted with trees and bushes, providing a serene and natural backdrop to this captivating scene. The image exudes a sense of tranquility and harmony with nature." \item " A beach with the sun rising and the waves crashing on the sand, beauty light" 
    \item "Super cool astronaut NFT, facing camera, stylish, bling glasses, photochromatic, plain background, futuristic 8k. $->$ Super cool astronaut NFT, facing camera, stylish, bling glasses, photochromatic, plain background." 
    \item " The image captures a close-up of a woman's face, bathed in a mesmerizing blend of blue and yellow lights. The woman's eyes, wide open, gaze directly into the camera. Her hair, styled in loose waves, adds a touch of softness to the intense lighting. The background is shrouded in darkness, punctuated by streaks of light from the left side of the frame. The image does not contain any discernible text or countable objects, and there are no visible actions taking place. The overall composition is balanced, with the woman's face as the central focus." 
    \item "The image captures a serene indoor setting, dominated by a white vase that holds a bouquet of flowers. The vase, positioned on the left side of the frame, is filled with three distinct types of flowers, each contributing to a harmonious blend of colors and shapes. The most prominent flower is a pink dahlia, its petals radiating out from a yellow center. This dahlia is situated on the right side of the vase, its vibrant color standing out against the white of the vase. In addition to the dahlia, there are two white chrysanthemums. These flowers are located on the left side of the vase, their pristine white petals adding a touch of elegance to the arrangement. The background of the image is blurred, drawing focus to the flowers in the vase. The overall composition of the image creates a sense of tranquility and beauty." 
\end{enumerate}

%% file: figures/pixart_images_1k/_pixart_quali_supp.tex
\begin{figure*}[t]
    \centering
    
    \begin{minipage}{0.02\textwidth}\raggedright
    \rotatebox[origin=c]{90}{LinFusion-DiT}
    \end{minipage}
    \begin{minipage}{0.19\textwidth}
    \includegraphics[width=\textwidth]{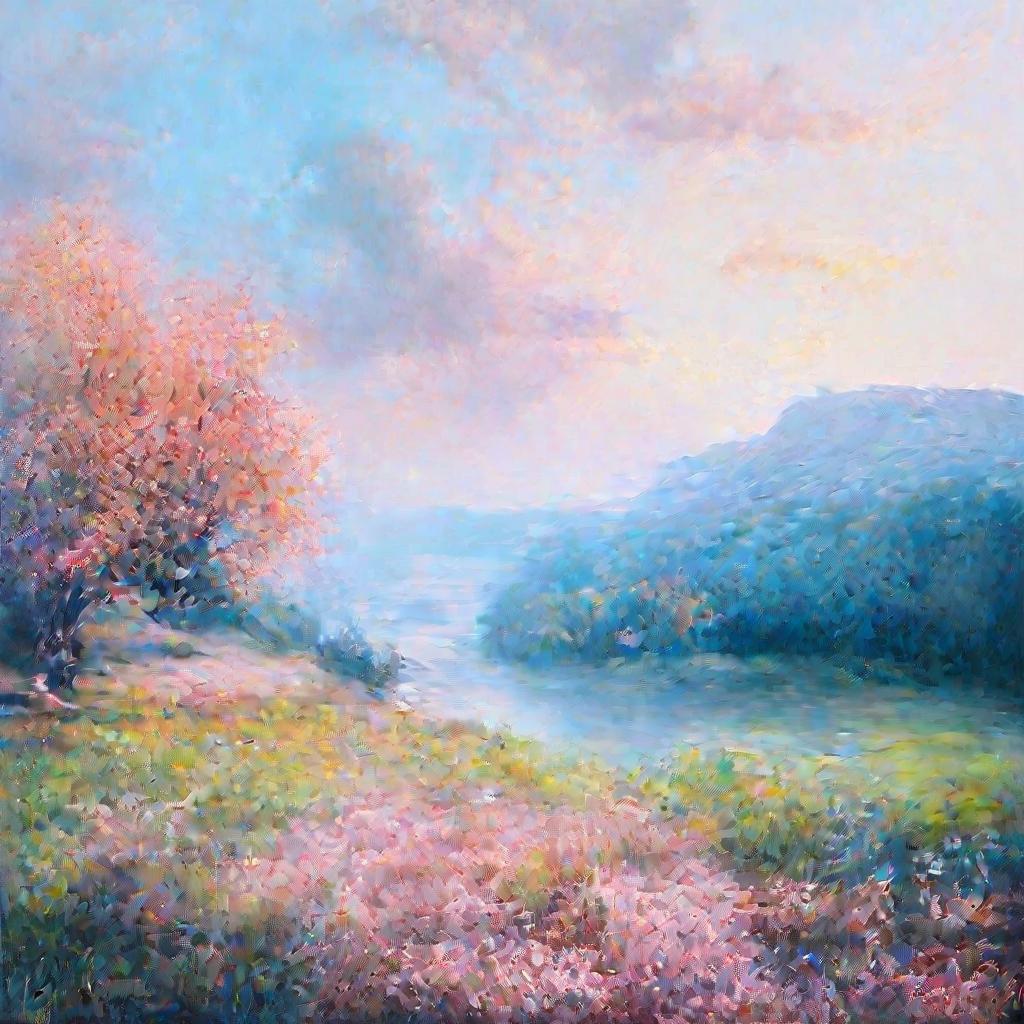}
    \end{minipage}
    \hfill
    \begin{minipage}{0.19\textwidth}
    \includegraphics[width=\textwidth]{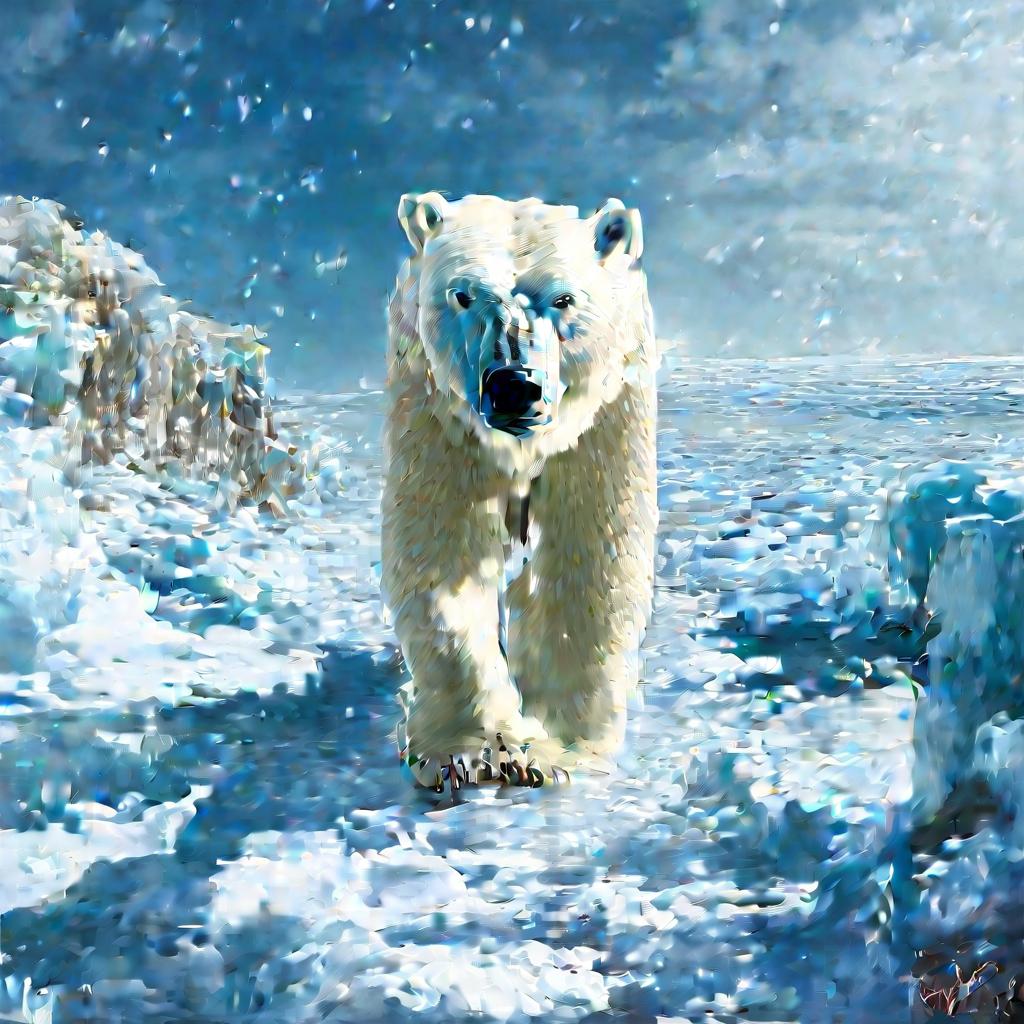}
    \end{minipage}
    \hfill
    \begin{minipage}{0.19\textwidth}
    \includegraphics[width=\textwidth]{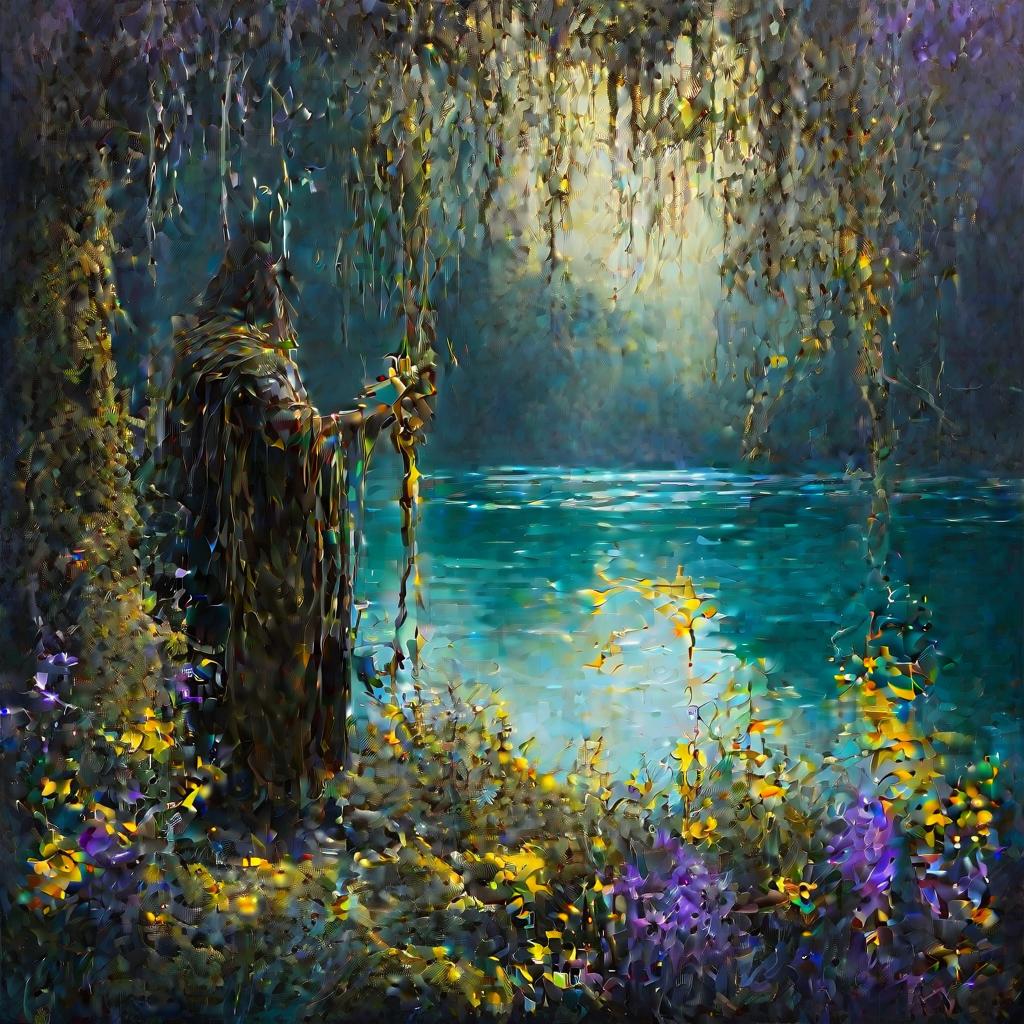}
    \end{minipage}
    \hfill
    \begin{minipage}{0.19\textwidth}
    \includegraphics[width=\textwidth]{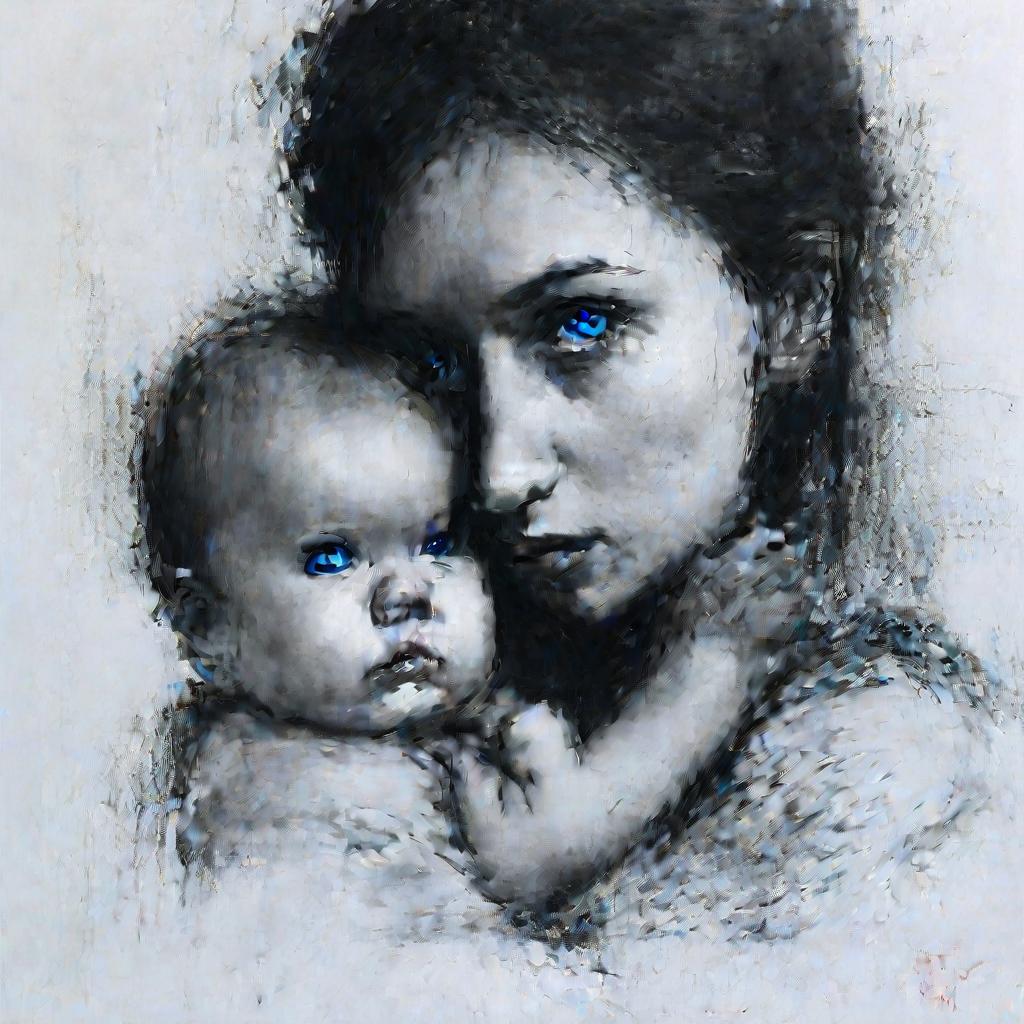}
    \end{minipage}
    \hfill
    \begin{minipage}{0.19\textwidth}
    \includegraphics[width=\textwidth]{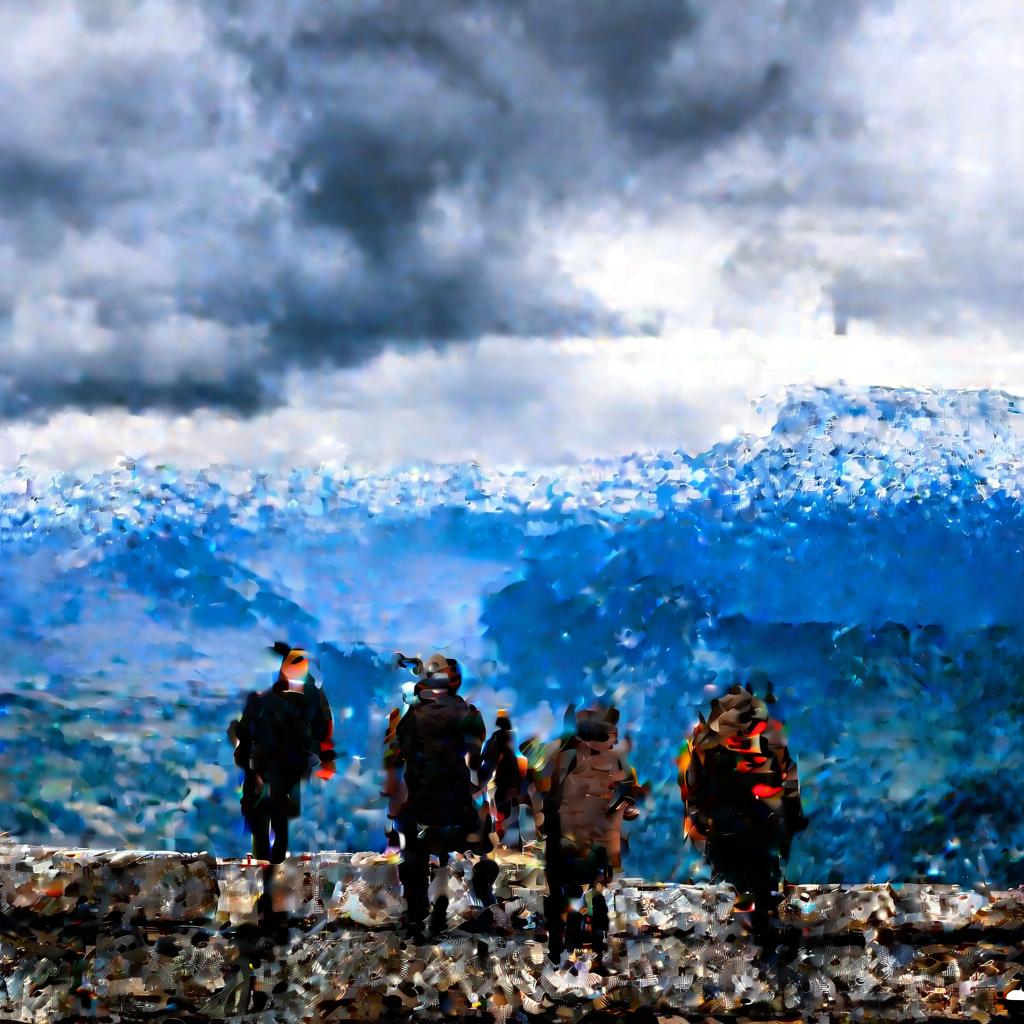}
    \end{minipage}

    \begin{minipage}{0.02\textwidth}\raggedright
    \rotatebox[origin=c]{90}{KV Comp. ($k=2$)}
    \end{minipage}
    \begin{minipage}{0.19\textwidth}
    \includegraphics[width=\textwidth]{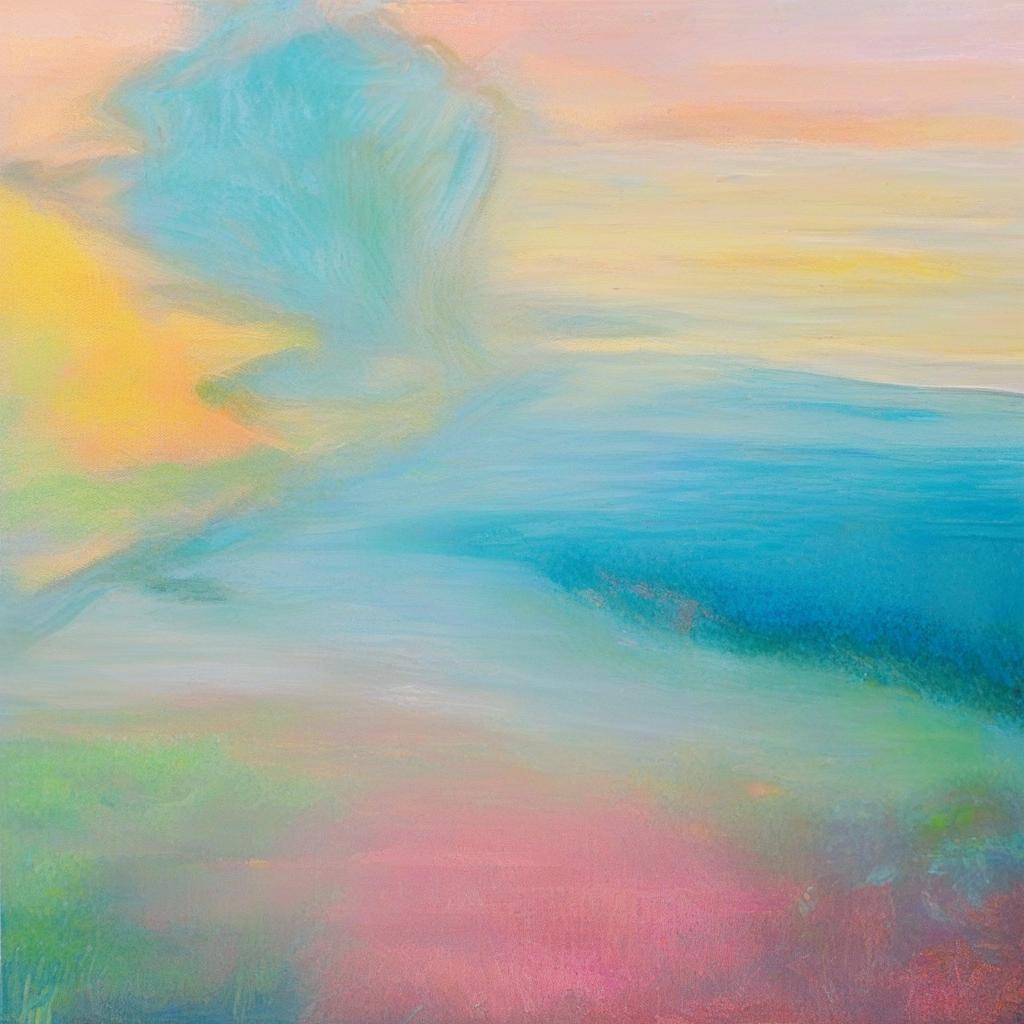}
    \end{minipage}
    \hfill
    \begin{minipage}{0.19\textwidth}
    \includegraphics[width=\textwidth]{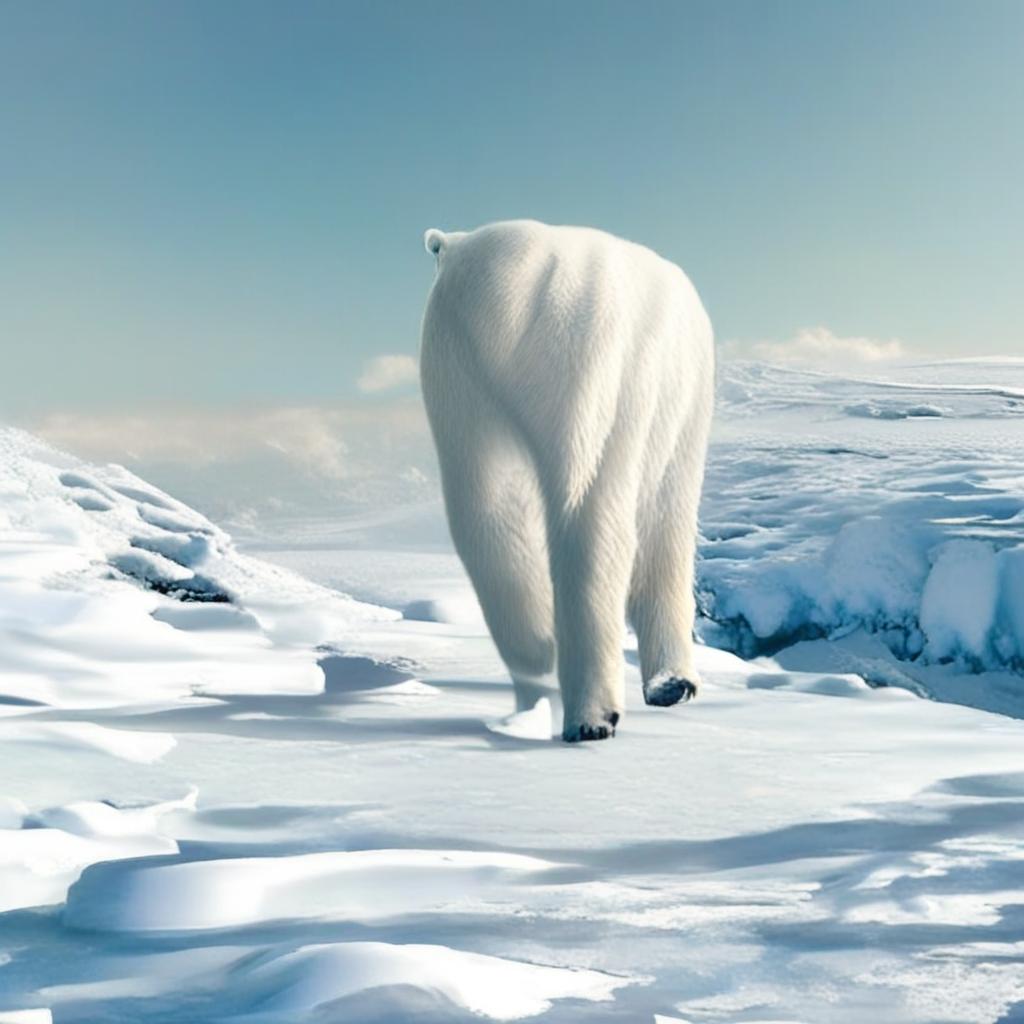}
    \end{minipage}
    \hfill
    \begin{minipage}{0.19\textwidth}
    \includegraphics[width=\textwidth]{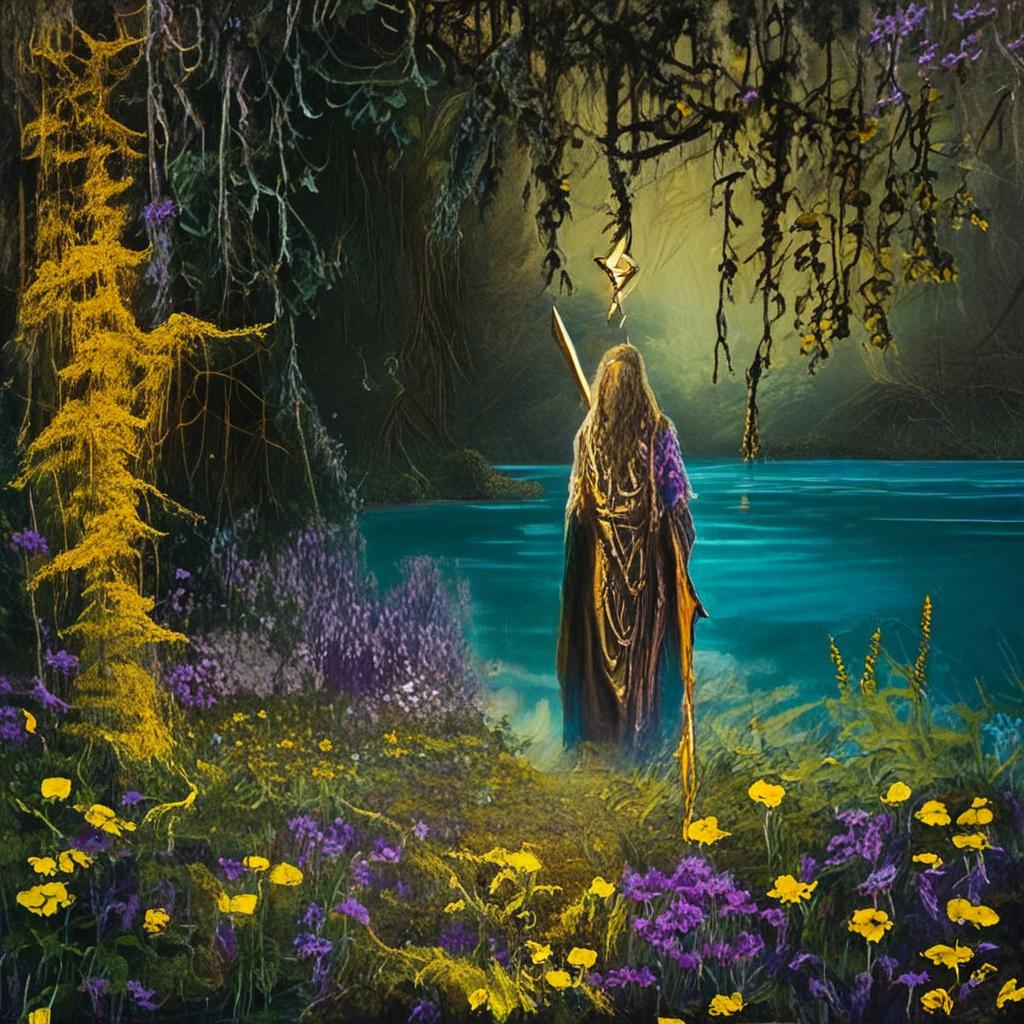}
    \end{minipage}
    \hfill
    \begin{minipage}{0.19\textwidth}
    \includegraphics[width=\textwidth]{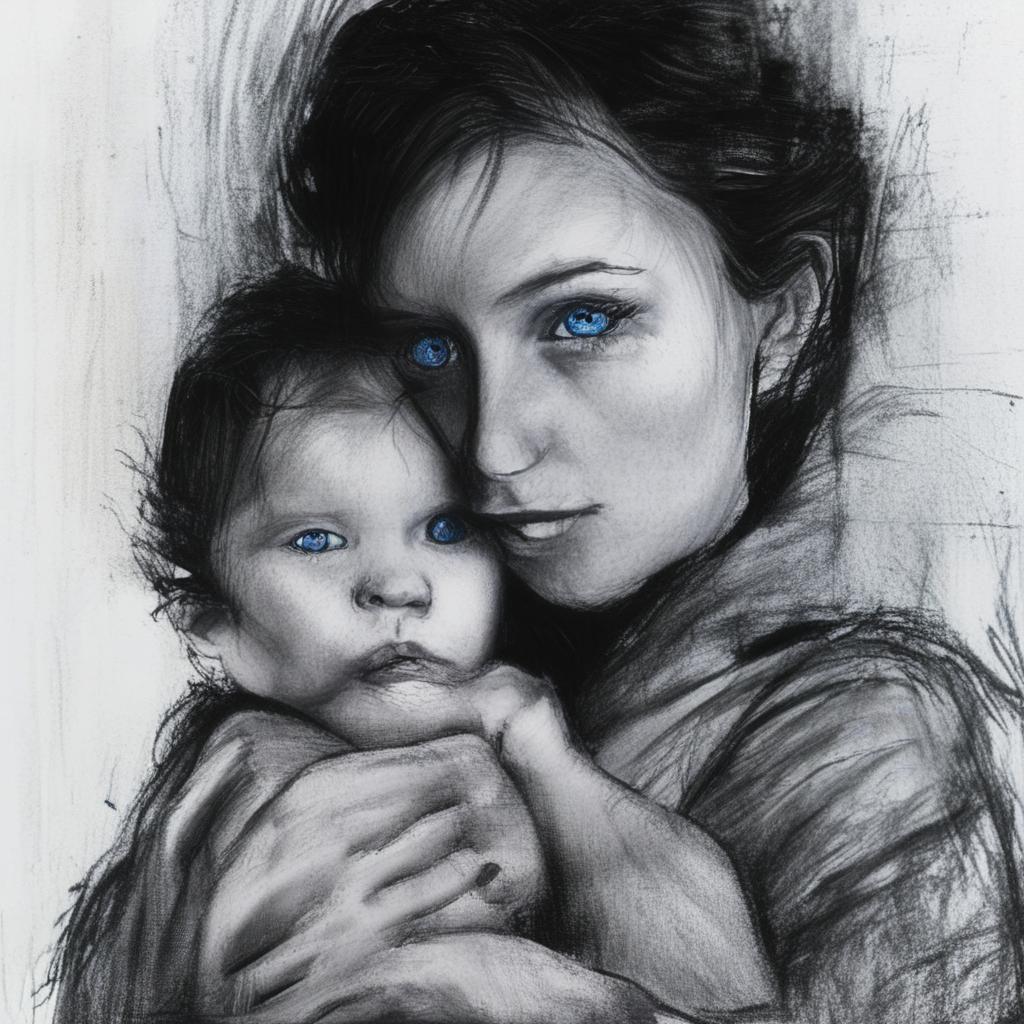}
    \end{minipage}
    \hfill
    \begin{minipage}{0.19\textwidth}
    \includegraphics[width=\textwidth]{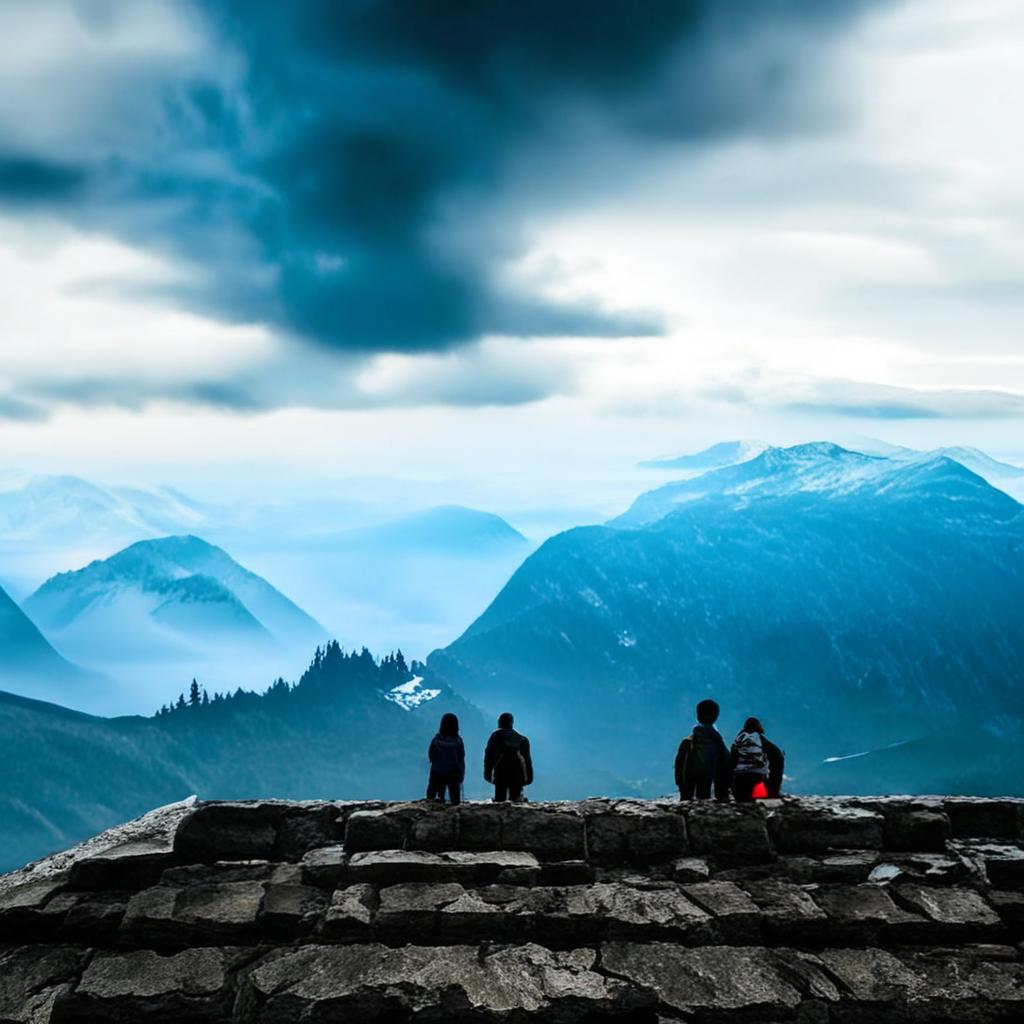}
    \end{minipage}

    \begin{minipage}{0.02\textwidth}\raggedright
    \rotatebox[origin=c]{90}{Q = CF, K = CF, V = -}
    \end{minipage}
    \begin{minipage}{0.19\textwidth}
    \includegraphics[width=\textwidth]{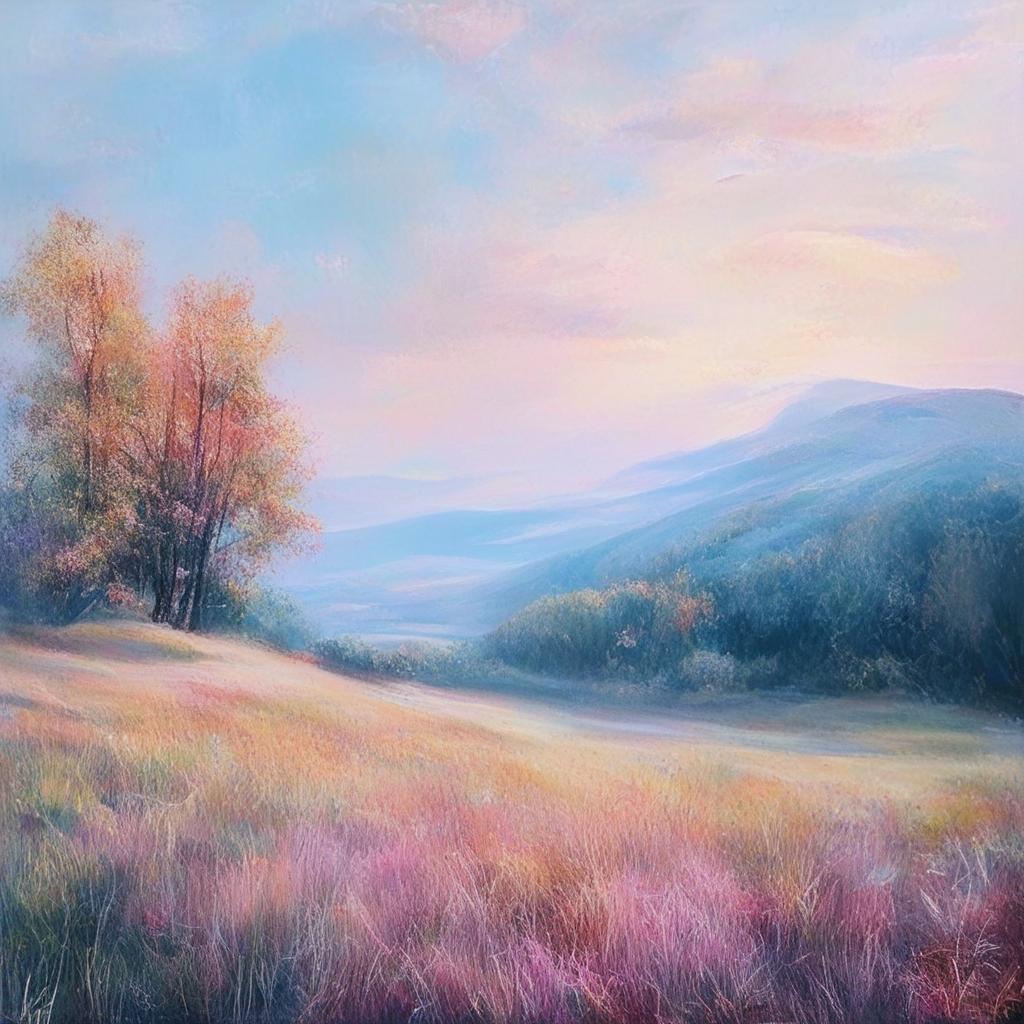}
    \end{minipage}
    \hfill
    \begin{minipage}{0.19\textwidth}
    \includegraphics[width=\textwidth]{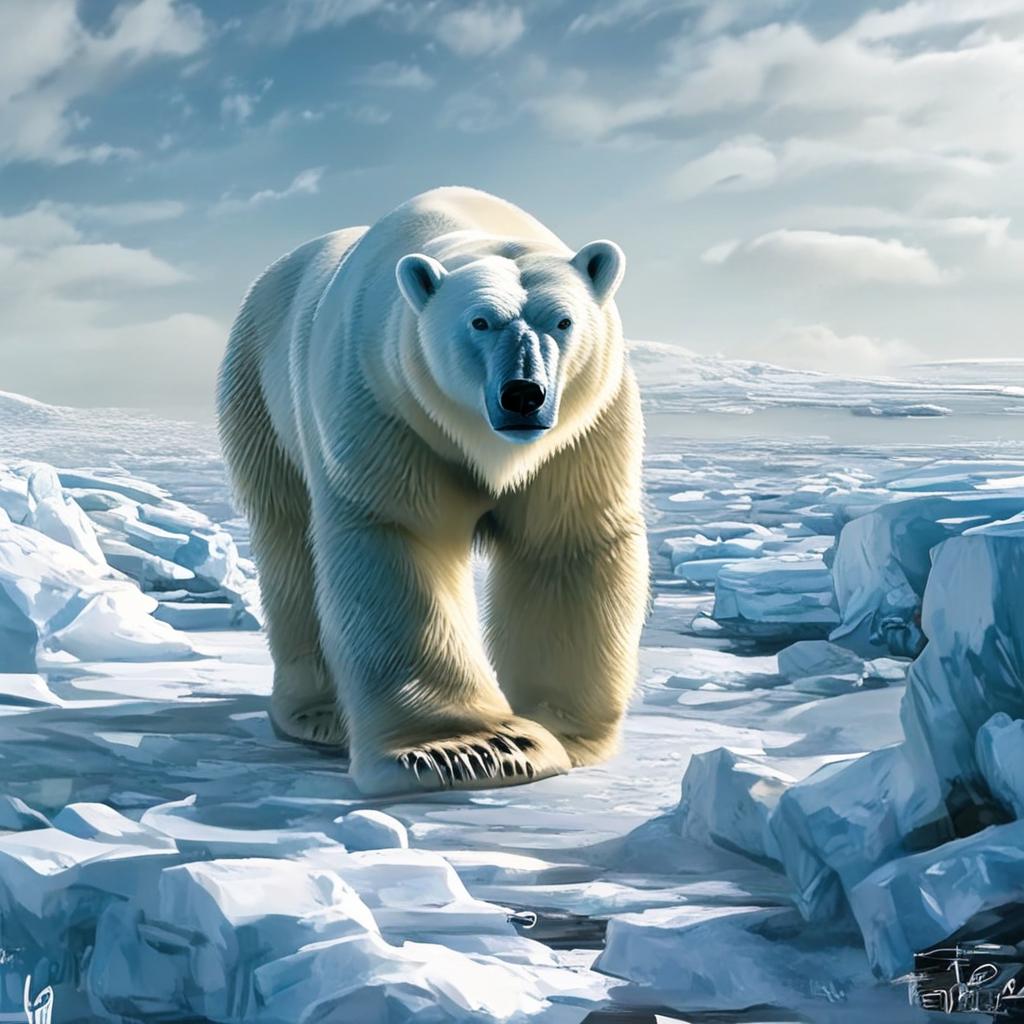}
    \end{minipage}
    \hfill
    \begin{minipage}{0.19\textwidth}
    \includegraphics[width=\textwidth]{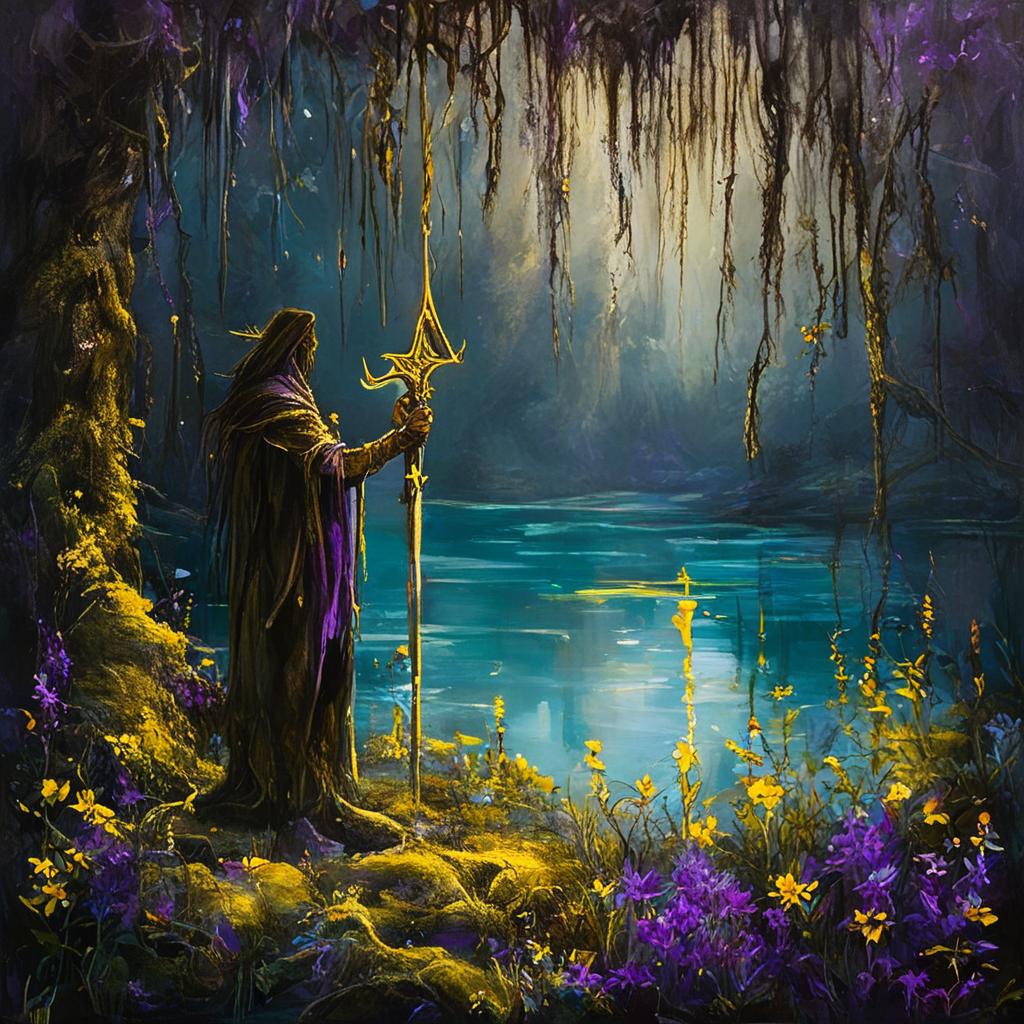}
    \end{minipage}
    \hfill
    \begin{minipage}{0.19\textwidth}
    \includegraphics[width=\textwidth]{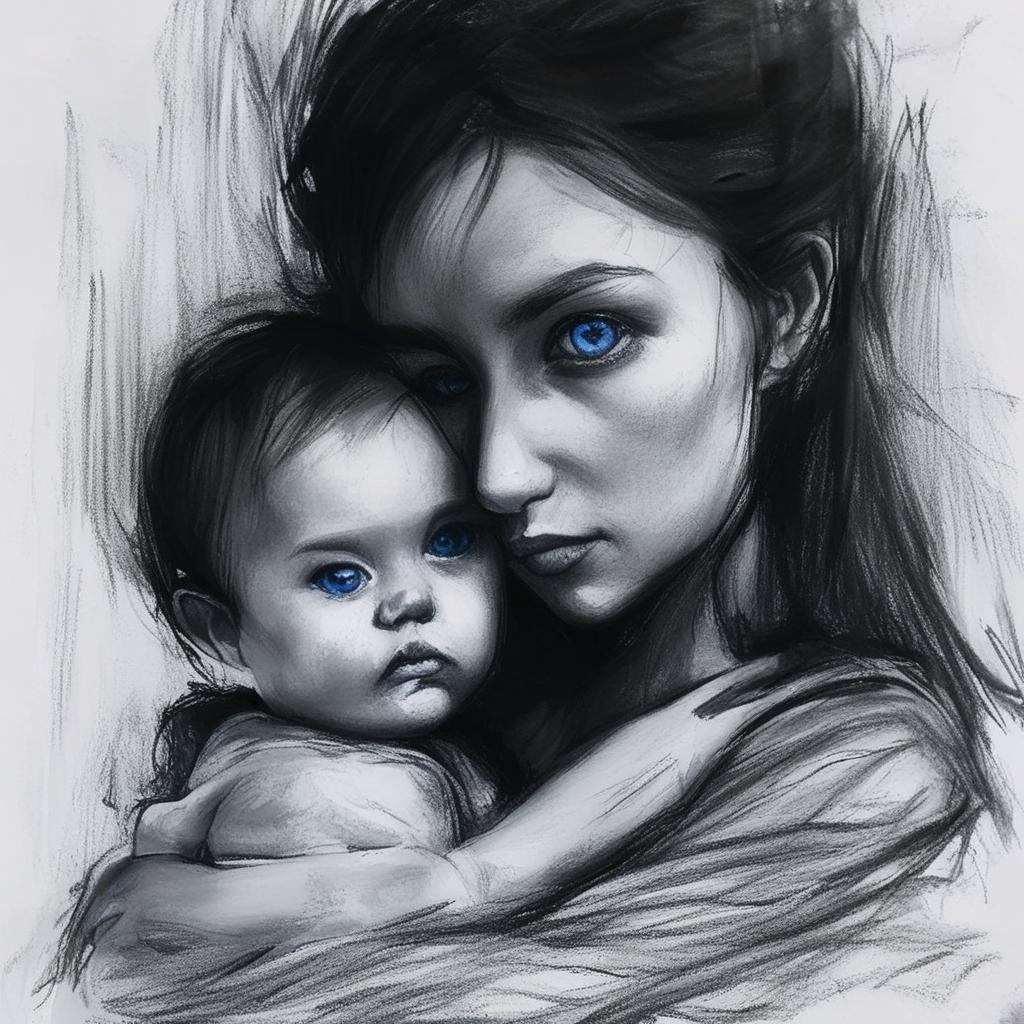}
    \end{minipage}
    \hfill
    \vspace{10pt}
    \begin{minipage}{0.19\textwidth}
    \includegraphics[width=\textwidth]{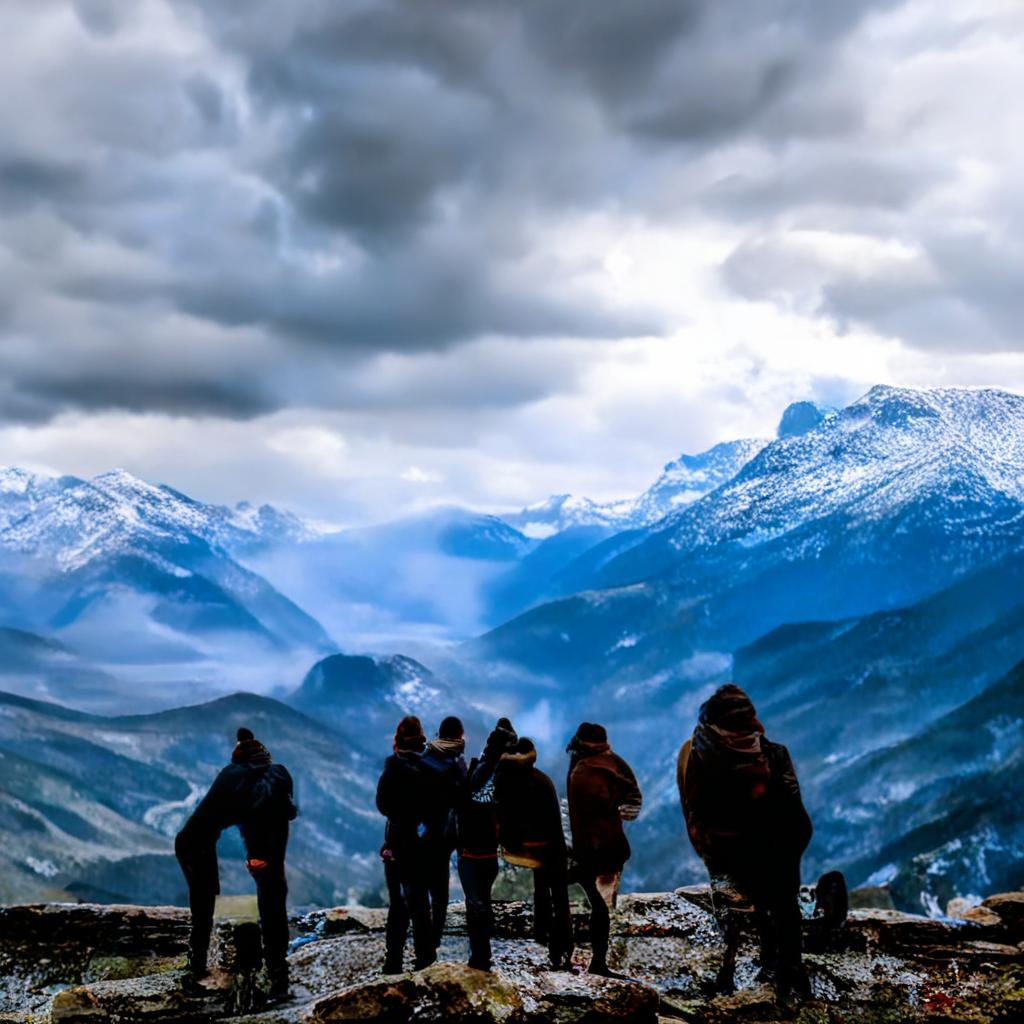}
    \end{minipage}

        \begin{minipage}{0.02\textwidth}\raggedright
    \rotatebox[origin=c]{90}{{Q = -, K = SC, V = SC}}
    \end{minipage}
    \begin{minipage}{0.19\textwidth}
    \includegraphics[width=\textwidth]{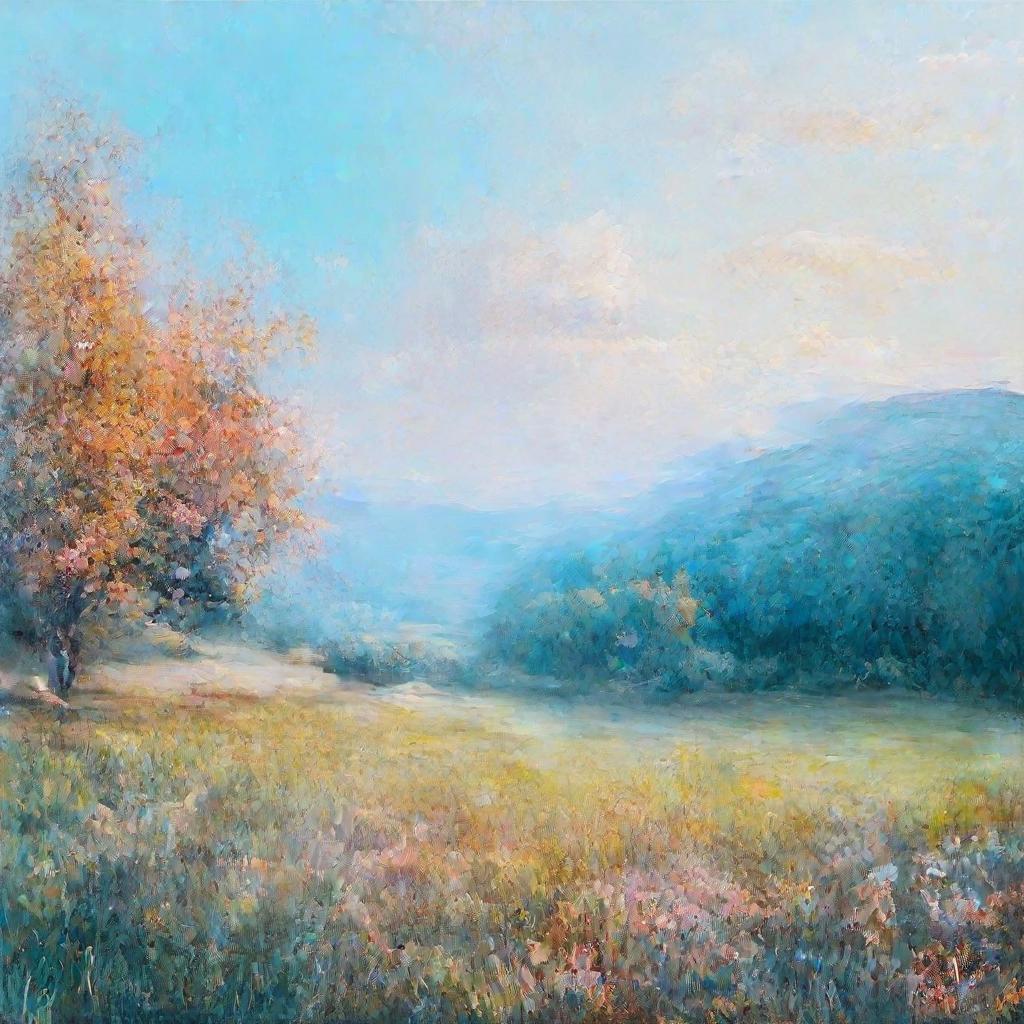}
    \end{minipage}
    \hfill
    \begin{minipage}{0.19\textwidth}
    \includegraphics[width=\textwidth]{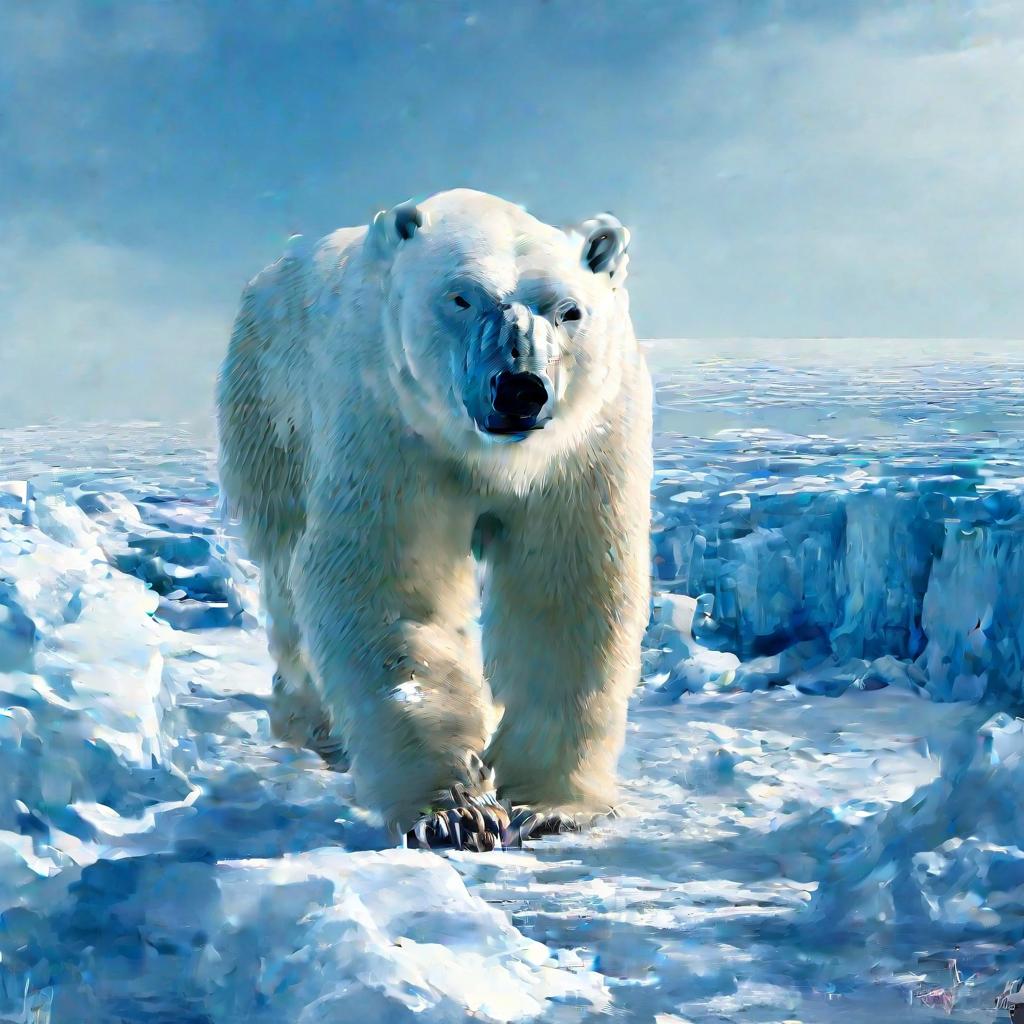}
    \end{minipage}
    \hfill
    \begin{minipage}{0.19\textwidth}
    \includegraphics[width=\textwidth]{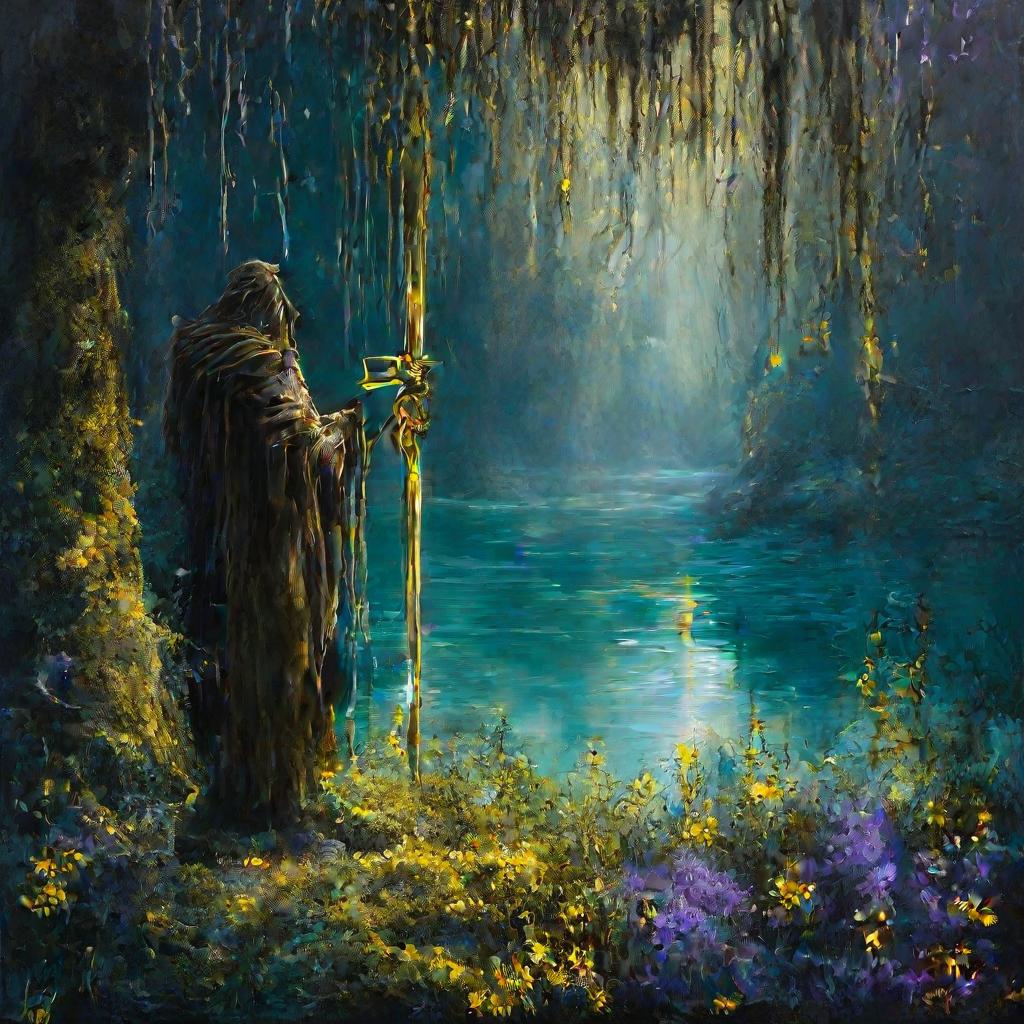}
    \end{minipage}
    \hfill
    \begin{minipage}{0.19\textwidth}
    \includegraphics[width=\textwidth]{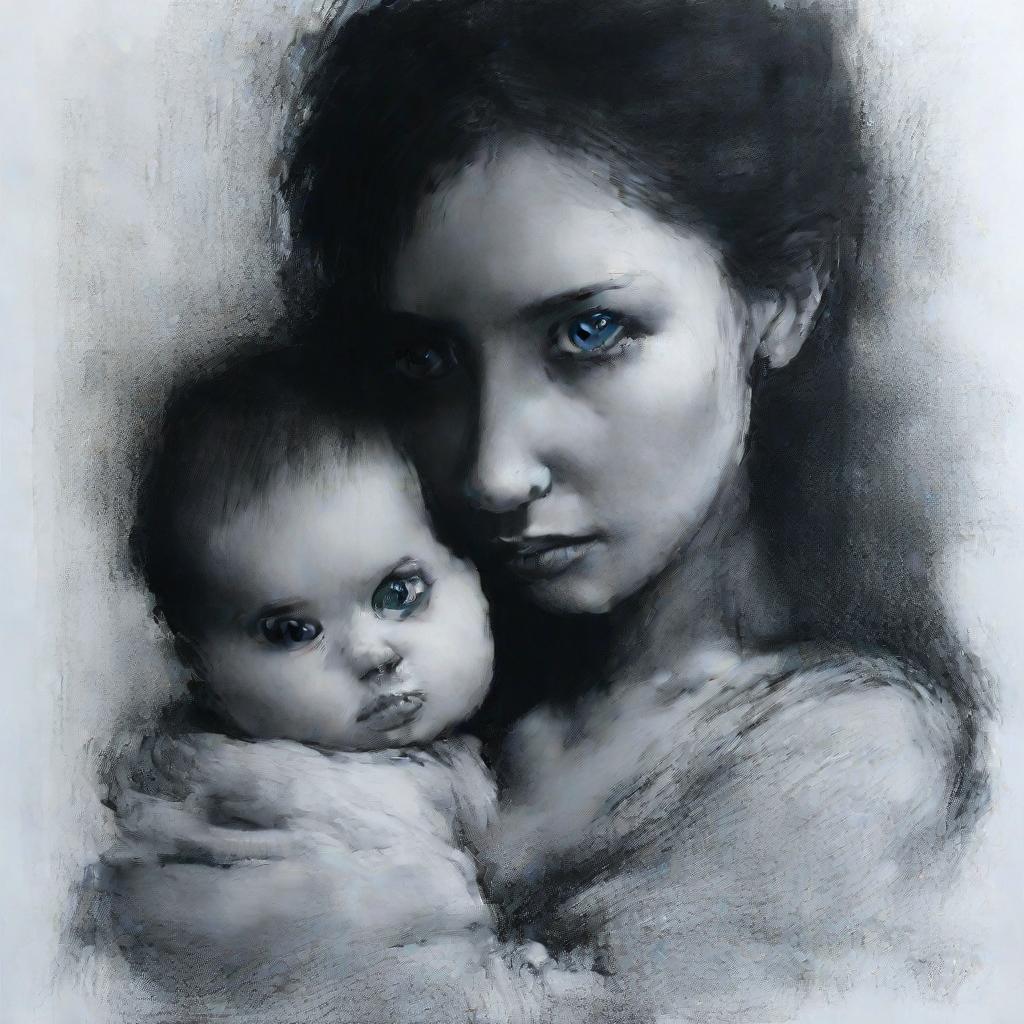}
    \end{minipage}
    \hfill
    \vspace{10pt}
    \begin{minipage}{0.19\textwidth}
    \includegraphics[width=\textwidth]{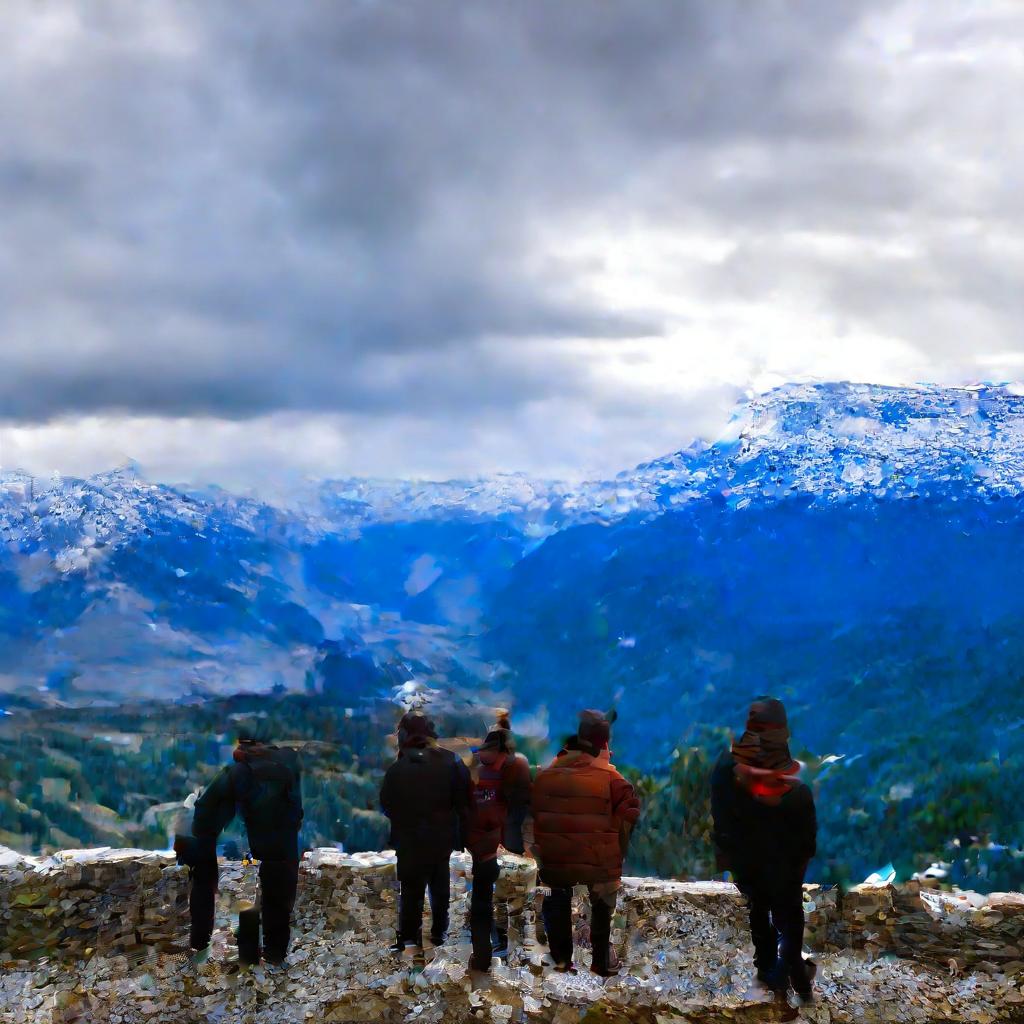}
    \end{minipage}

        \begin{minipage}{0.02\textwidth}\raggedright
    \rotatebox[origin=c]{90}{{Q = CF, K = -, V = -}}
    \end{minipage}
    \begin{minipage}{0.19\textwidth}
    \includegraphics[width=\textwidth]{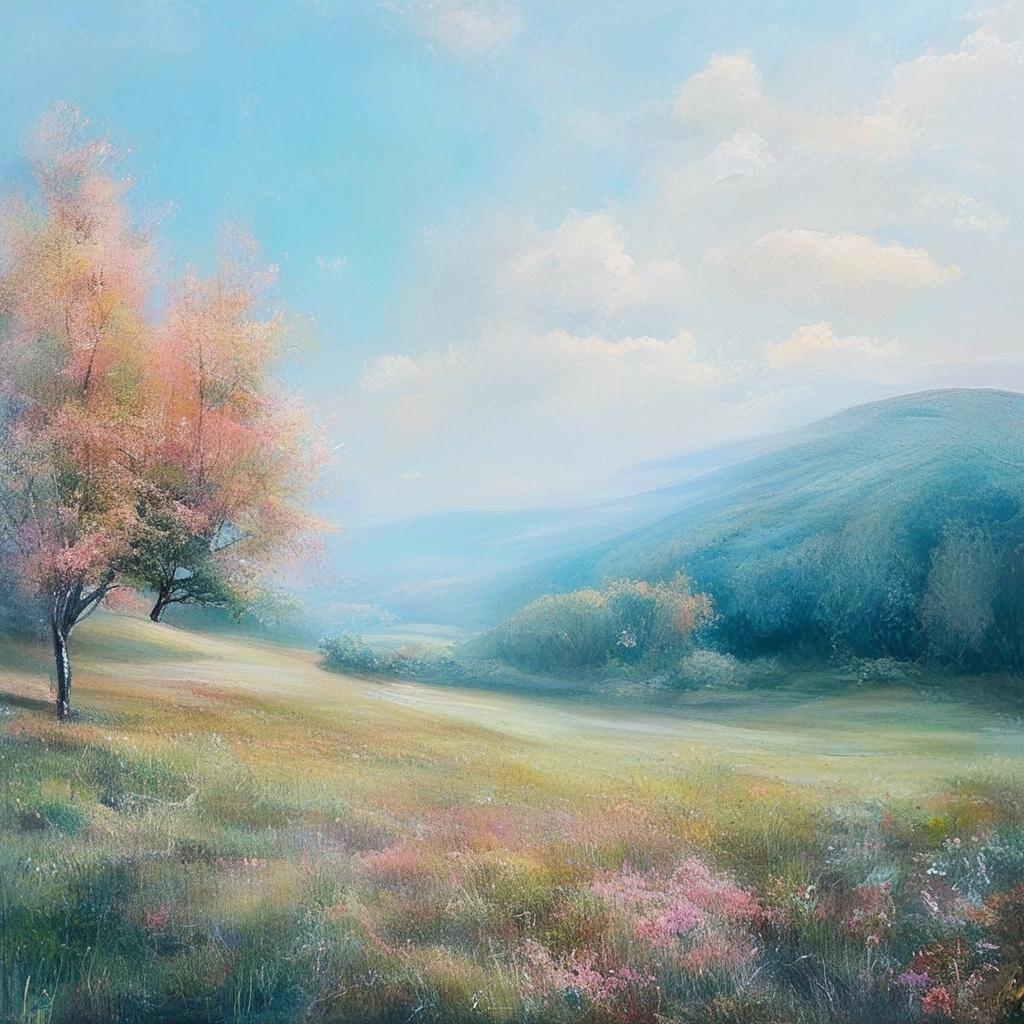}
    \end{minipage}
    \hfill
    \begin{minipage}{0.19\textwidth}
    \includegraphics[width=\textwidth]{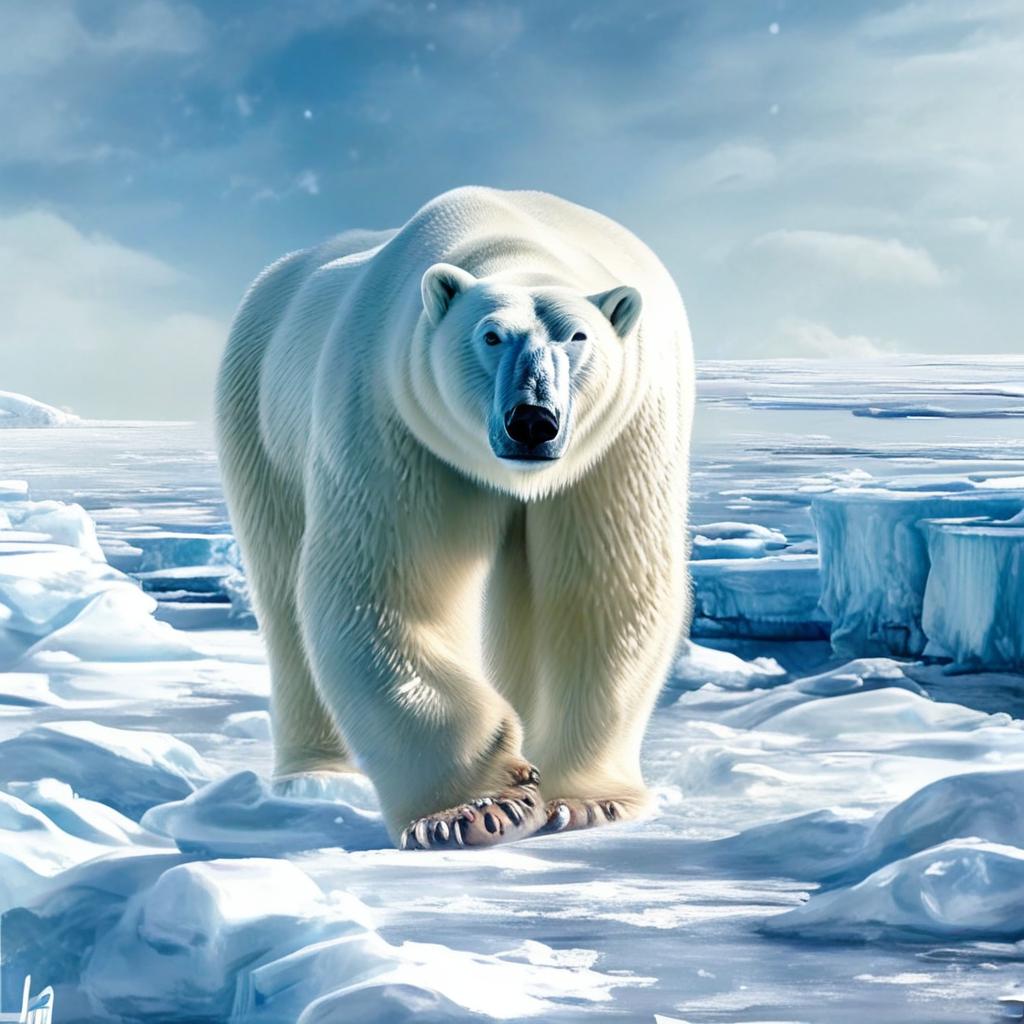}
    \end{minipage}
    \hfill
    \begin{minipage}{0.19\textwidth}
    \includegraphics[width=\textwidth]{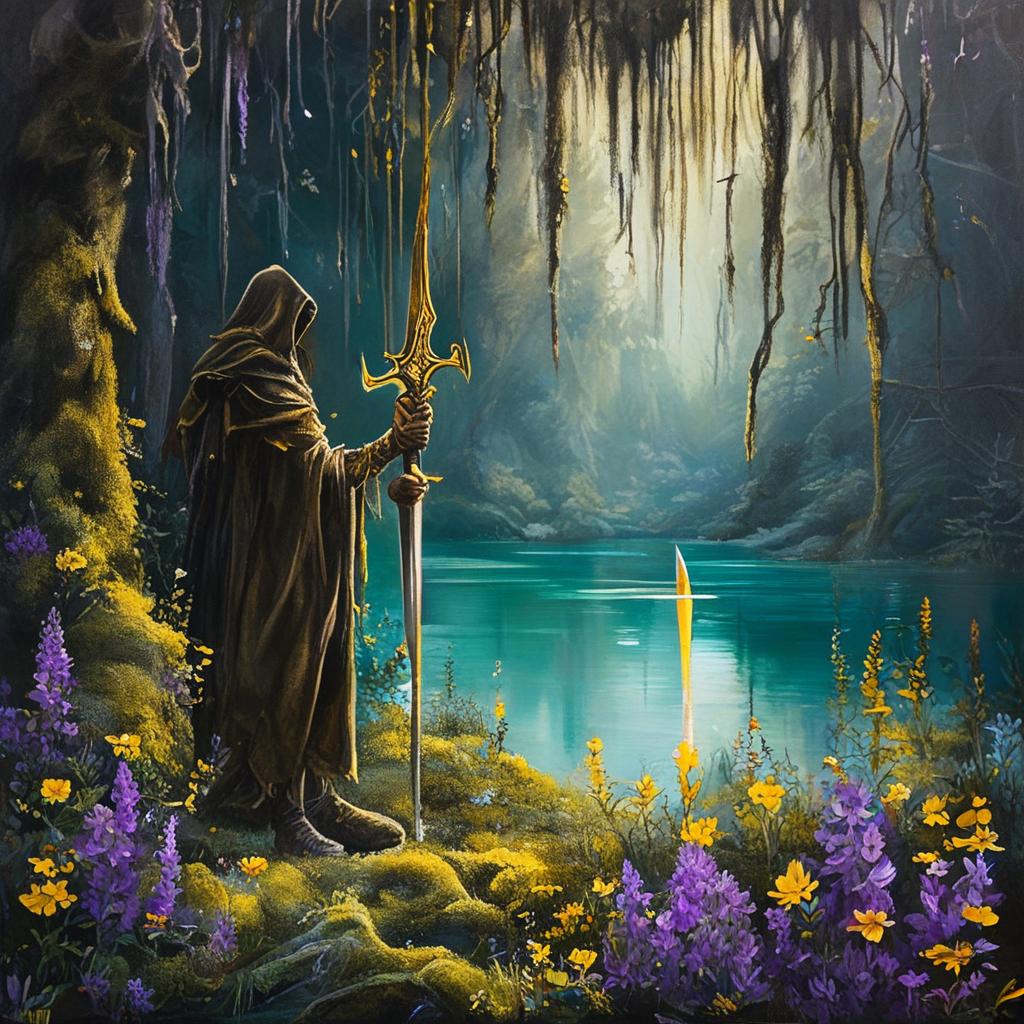}
    \end{minipage}
    \hfill
    \begin{minipage}{0.19\textwidth}
    \includegraphics[width=\textwidth]{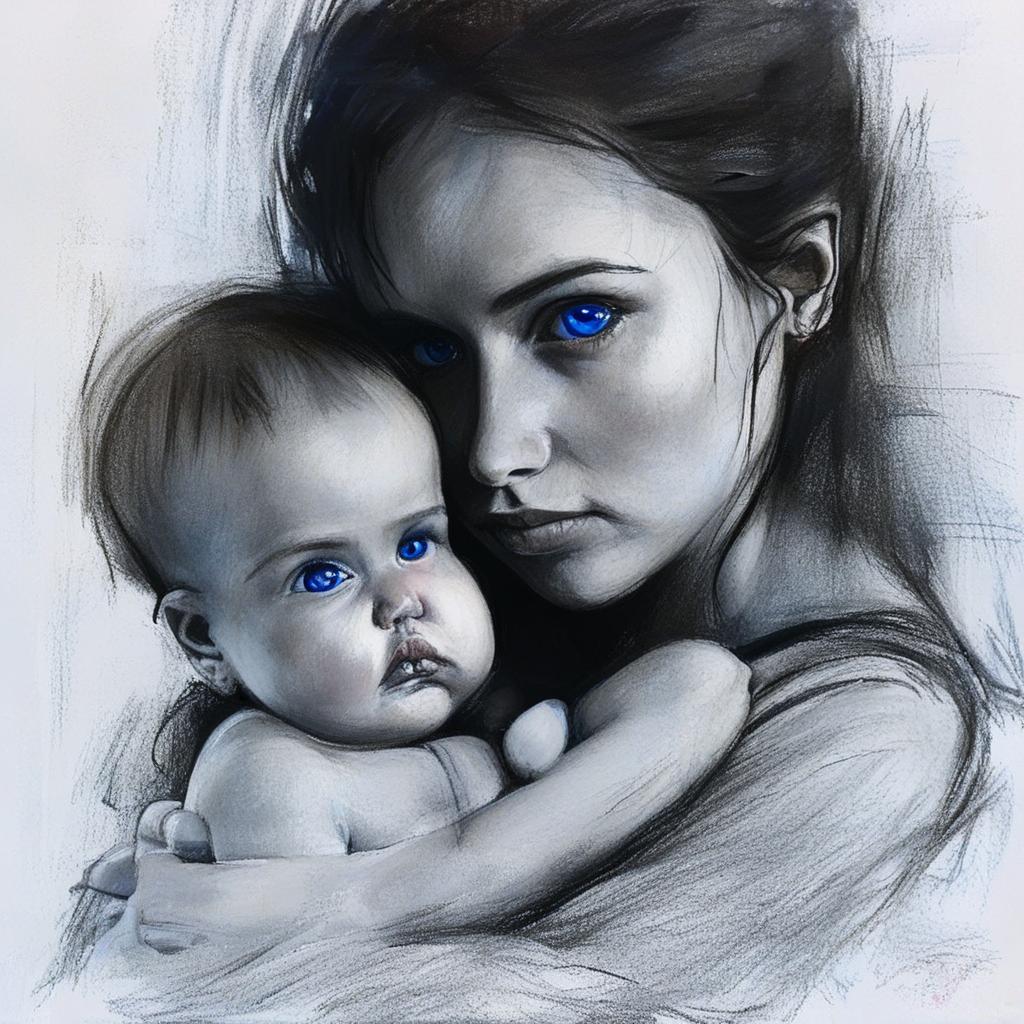}
    \end{minipage}
    \hfill
    \vspace{10pt}
    \begin{minipage}{0.19\textwidth}
    \includegraphics[width=\textwidth]{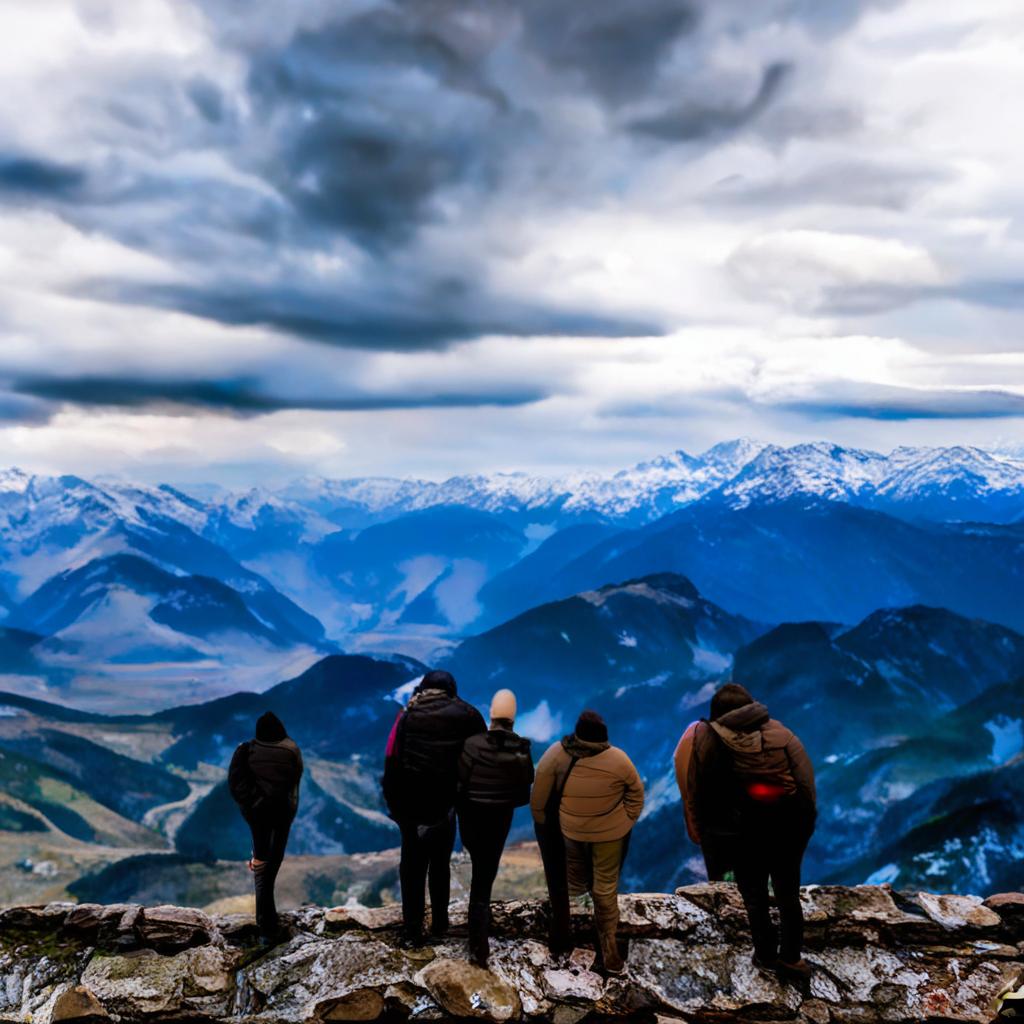}
    \end{minipage}
    \caption{$1024 \times 1024$ pixel images for the EDiT ablations and baselines not shown in the main paper.}
    \label{fig:supp_pixart}
\end{figure*}

%% file: figures/sd35_images_1k/_sd35_quali_sup.tex
\begin{figure*}
\begin{minipage}{0.02\textwidth}\raggedright
    \rotatebox[origin=c]{90}{MM-EDiT with $\eta^\textrm{Lin}$}
    \end{minipage}
    \begin{minipage}{0.19\textwidth}
    \includegraphics[width=\textwidth]{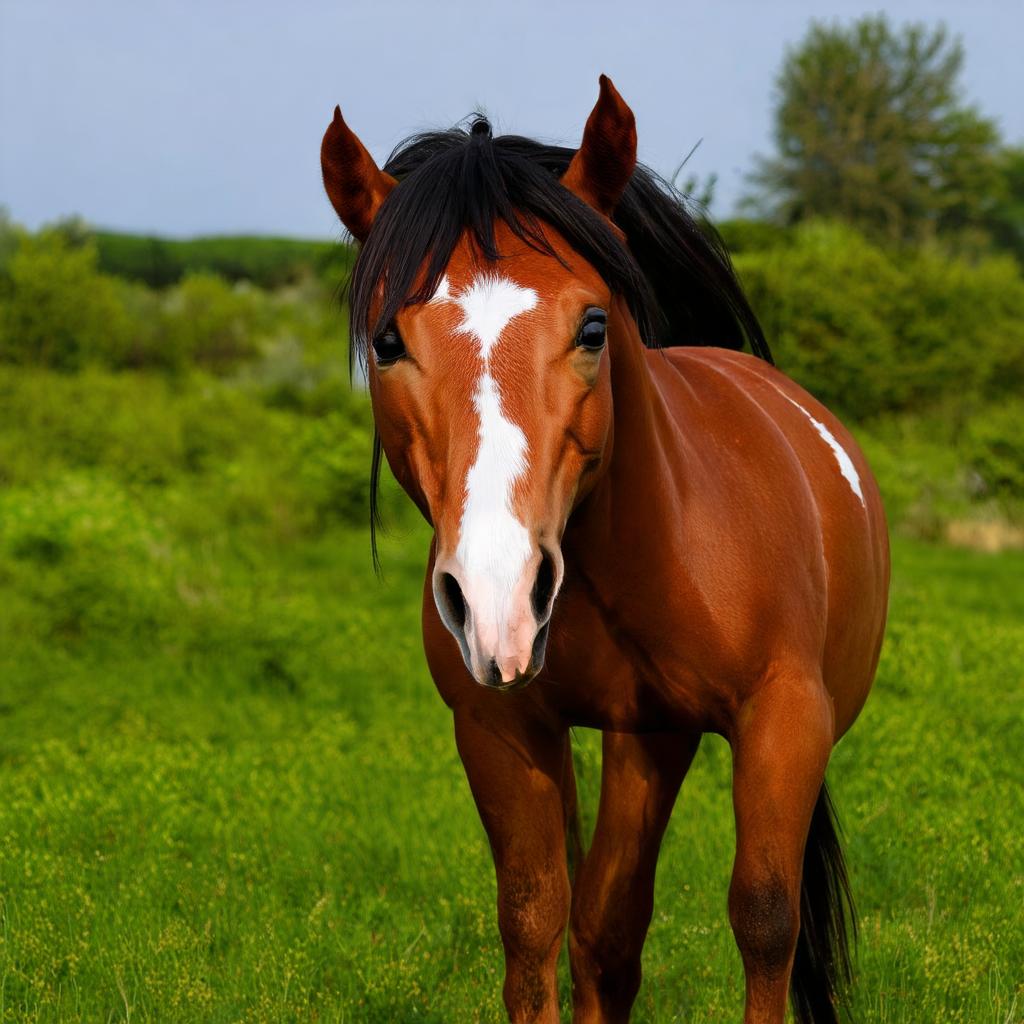}
    \end{minipage}
    \hfill
    \begin{minipage}{0.19\textwidth}
    \includegraphics[width=\textwidth]{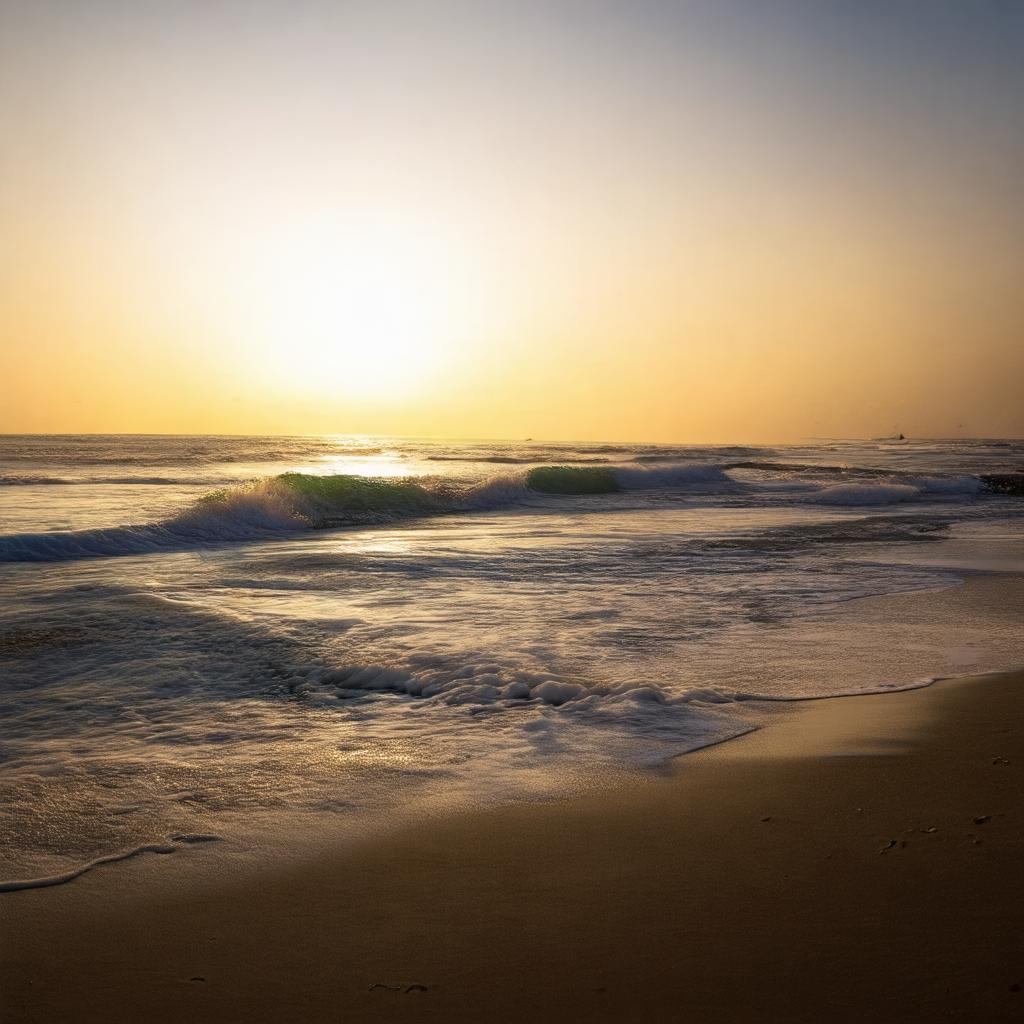}
    \end{minipage}
    \hfill
    \begin{minipage}{0.19\textwidth}
    \includegraphics[width=\textwidth]{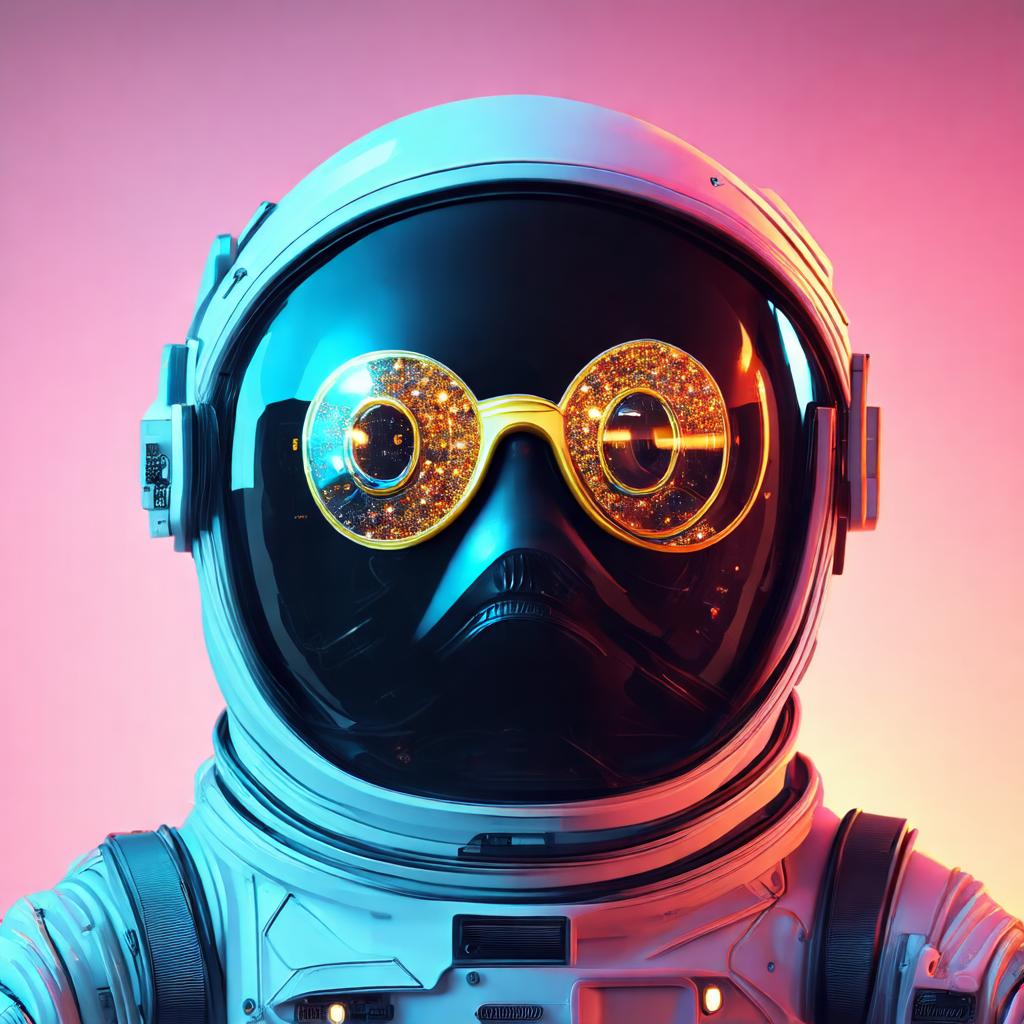}
    \end{minipage}
    \hfill
    \begin{minipage}{0.19\textwidth}
    \includegraphics[width=\textwidth]{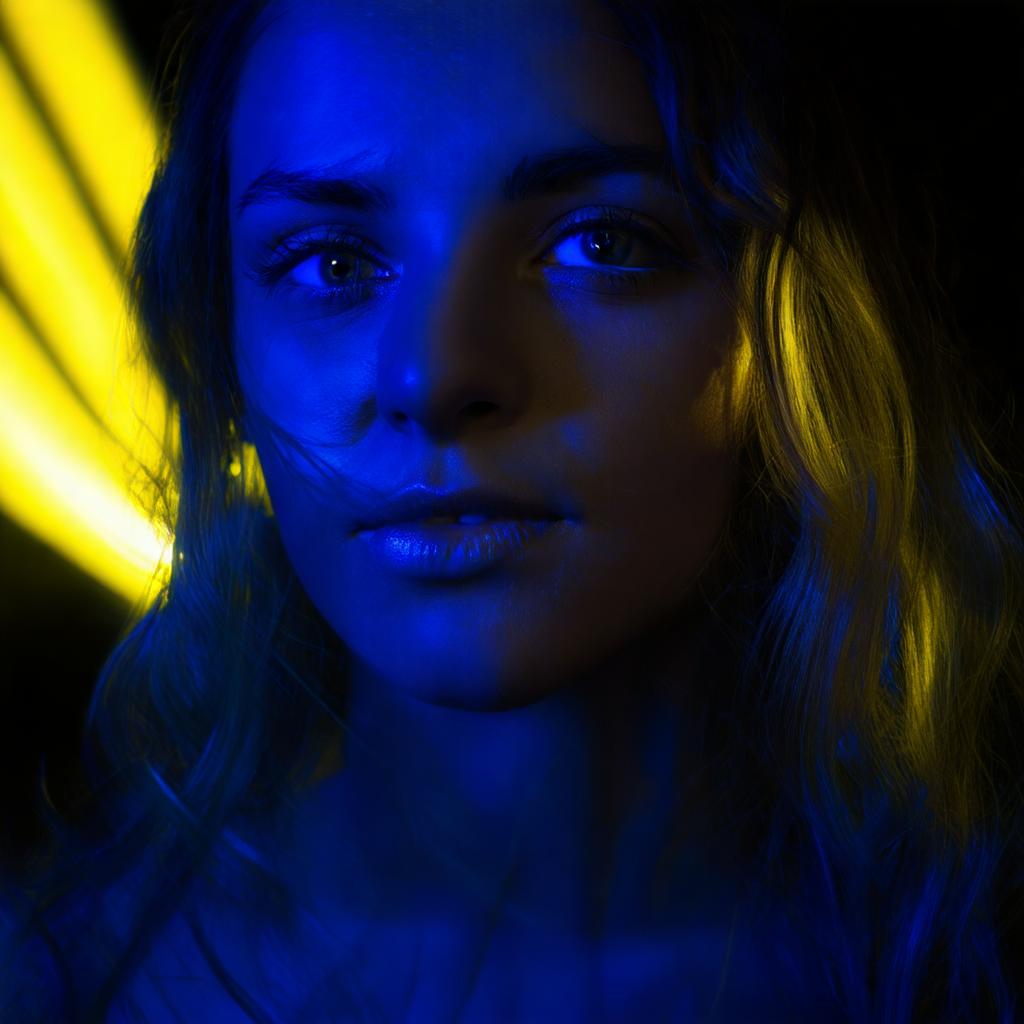}
    \end{minipage}
    \hfill
    \begin{minipage}{0.19\textwidth}
    \includegraphics[width=\textwidth]{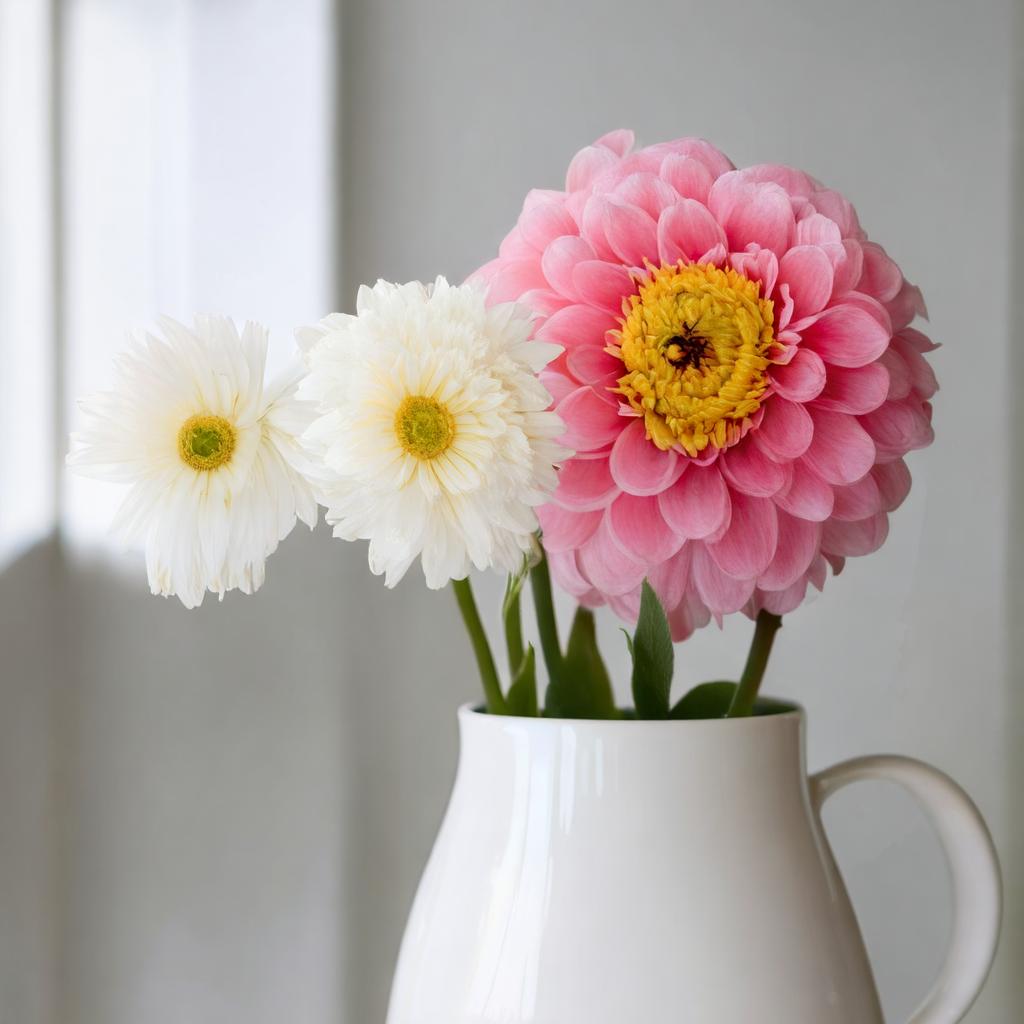}
    \end{minipage}

    \begin{minipage}{0.02\textwidth}\raggedright
     \rotatebox[origin=c]{90}{MM-EDiT no $\phi_\textrm{CF}$ and $\phi_\textrm{SC}$}
    \end{minipage}
    \begin{minipage}{0.19\textwidth}
    \includegraphics[width=\textwidth]{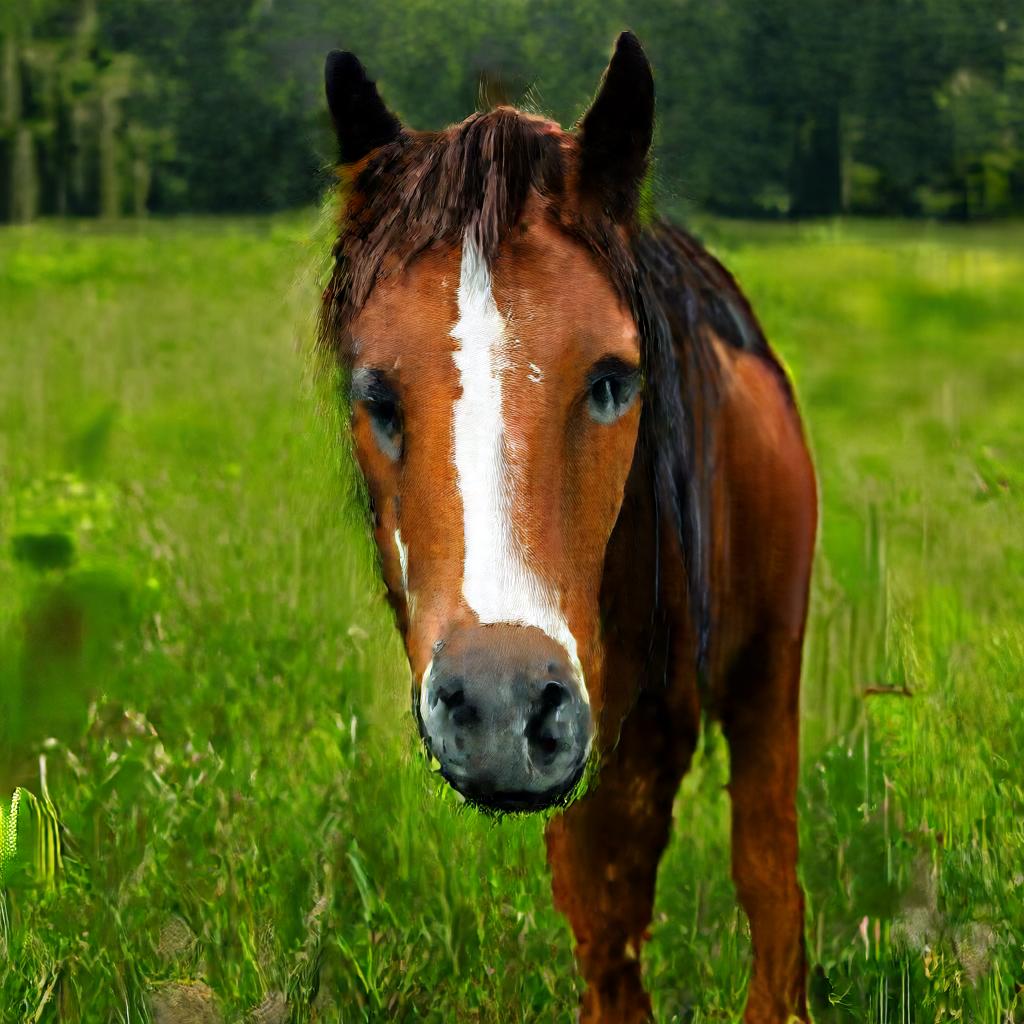}
    \end{minipage}
    \hfill
    \begin{minipage}{0.19\textwidth}
    \includegraphics[width=\textwidth]{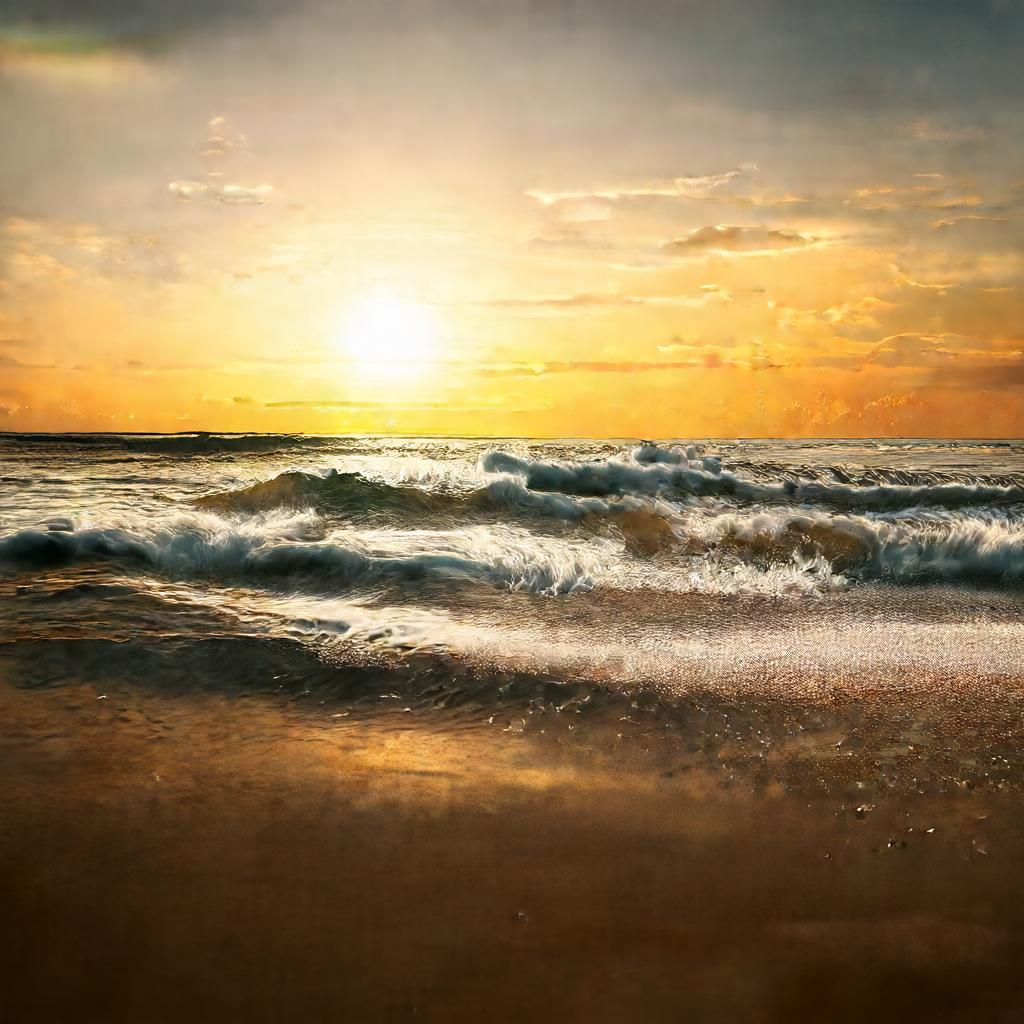}
    \end{minipage}
    \hfill
    \begin{minipage}{0.19\textwidth}
    \includegraphics[width=\textwidth]{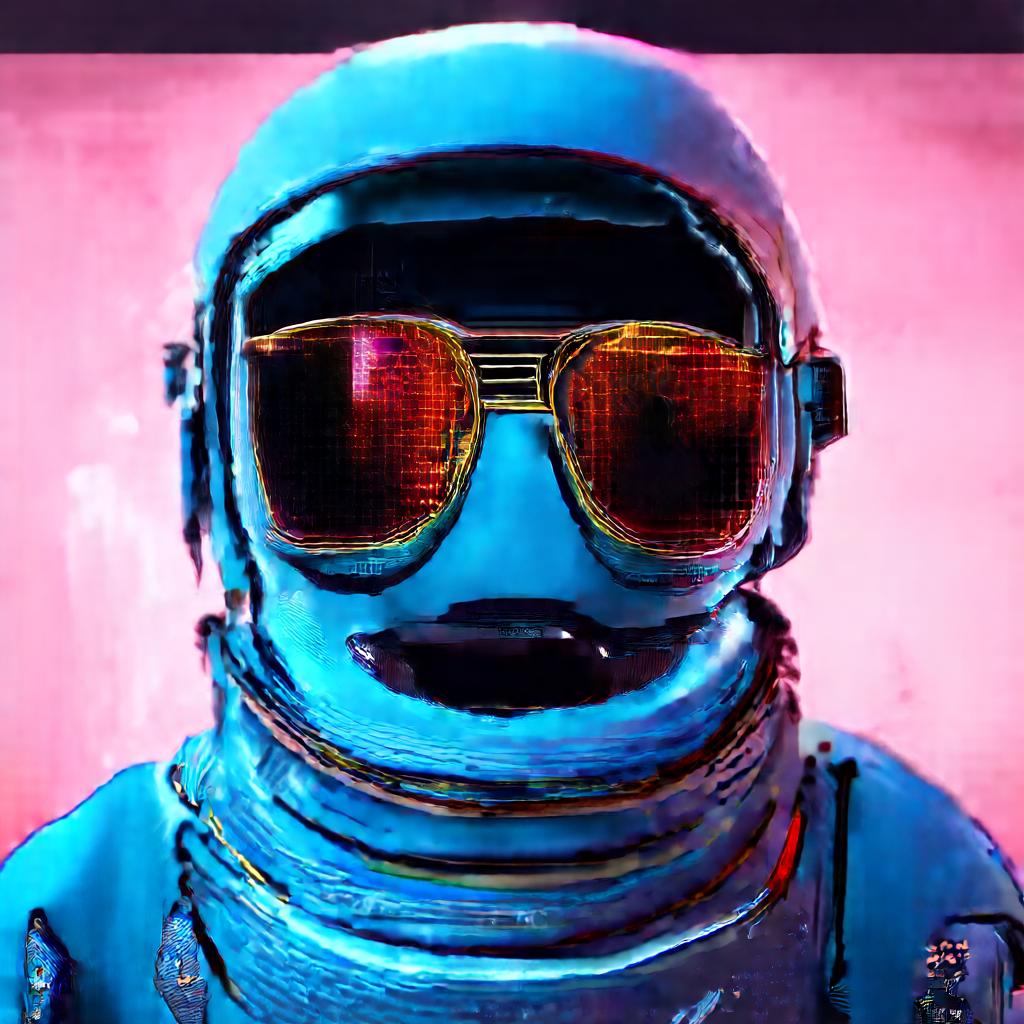}
    \end{minipage}
    \hfill
    \begin{minipage}{0.19\textwidth}
    \includegraphics[width=\textwidth]{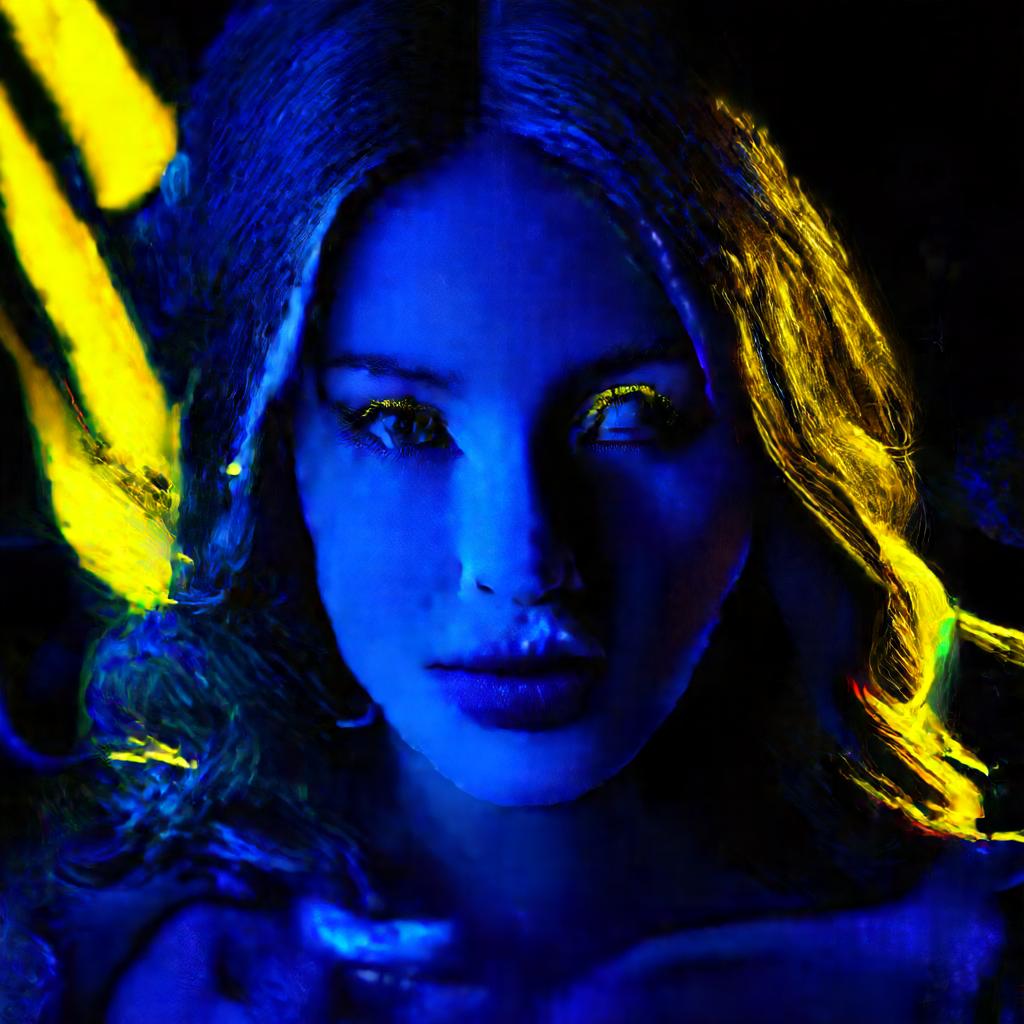}
    \end{minipage}
    \hfill
    \begin{minipage}{0.19\textwidth}
    \includegraphics[width=\textwidth]{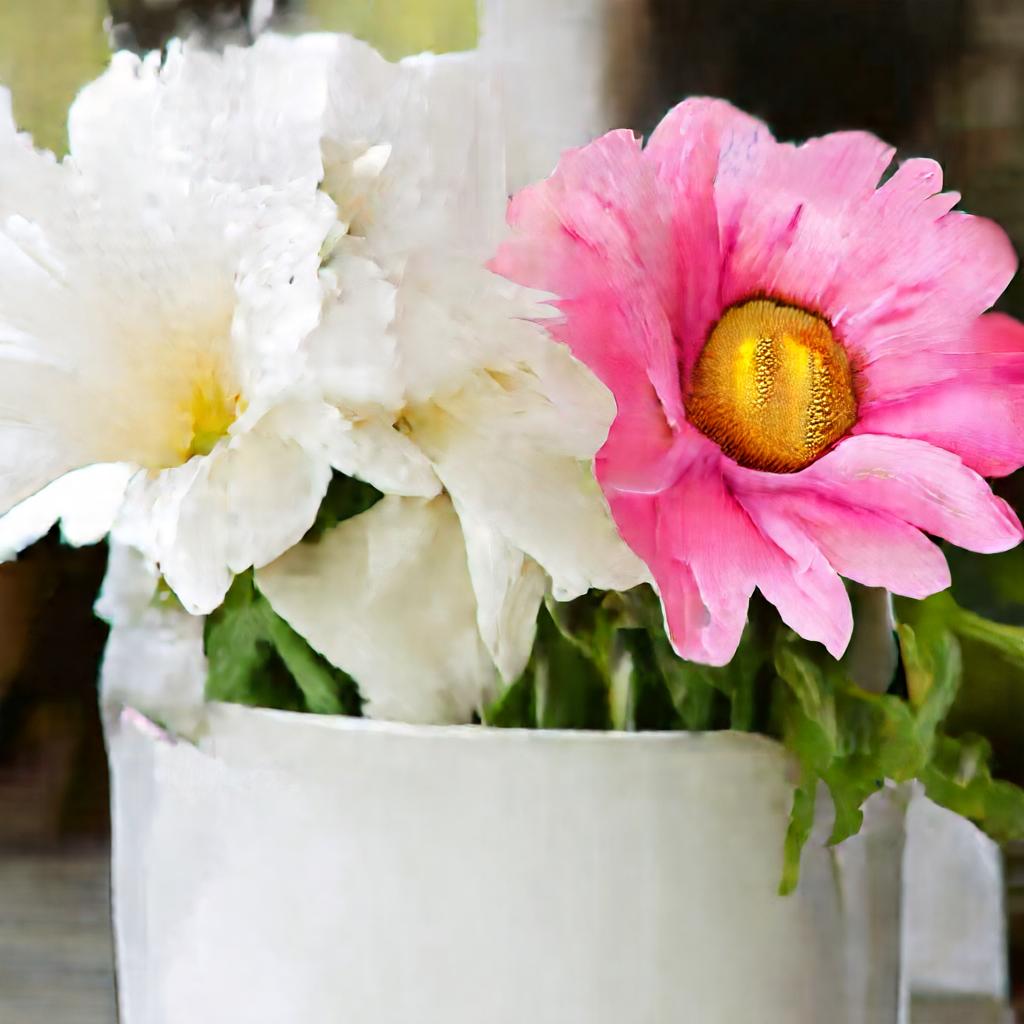}
    \end{minipage}

    \begin{minipage}{0.02\textwidth}\raggedright
       \rotatebox[origin=c]{90}{Linear MM-DiT-$\alpha$}
    \end{minipage}
    \begin{minipage}{0.19\textwidth}
    \includegraphics[width=\textwidth]{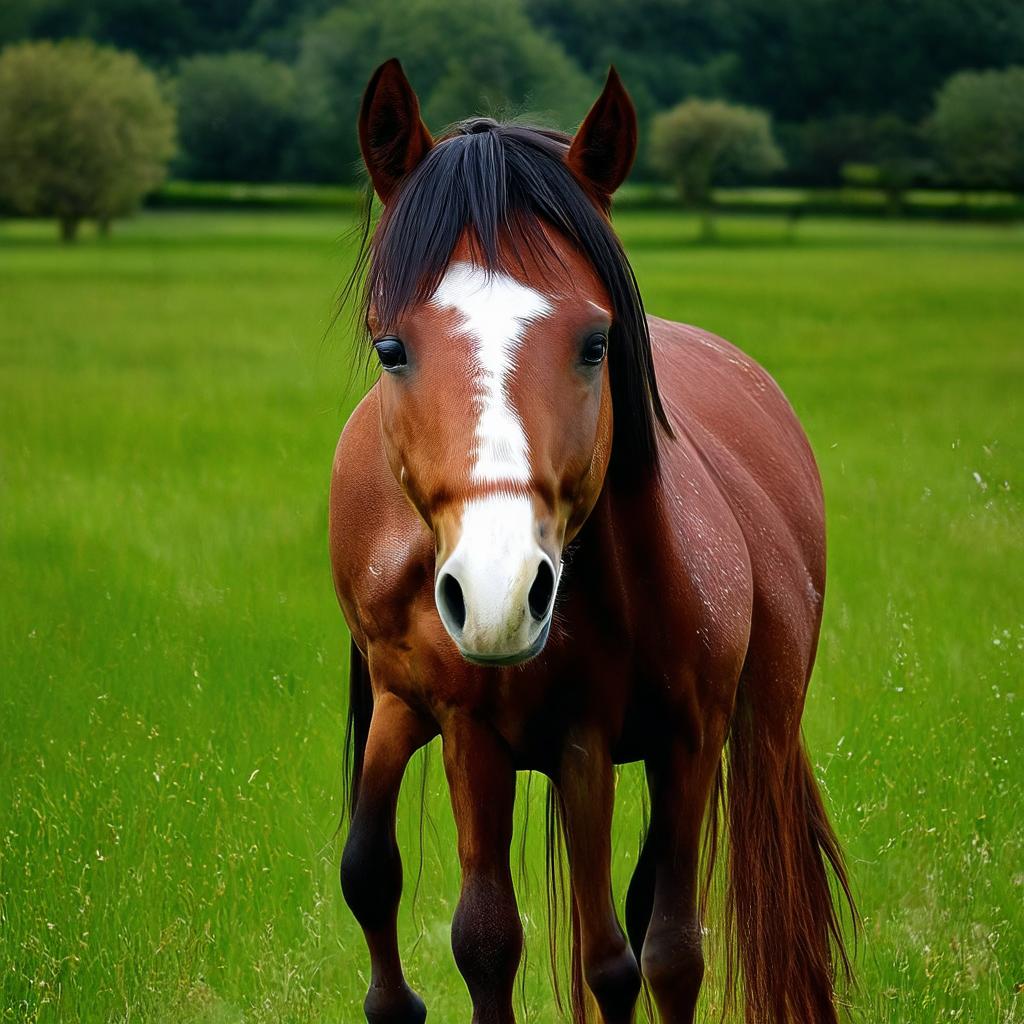}
    \end{minipage}
    \hfill
    \begin{minipage}{0.19\textwidth}
    \includegraphics[width=\textwidth]{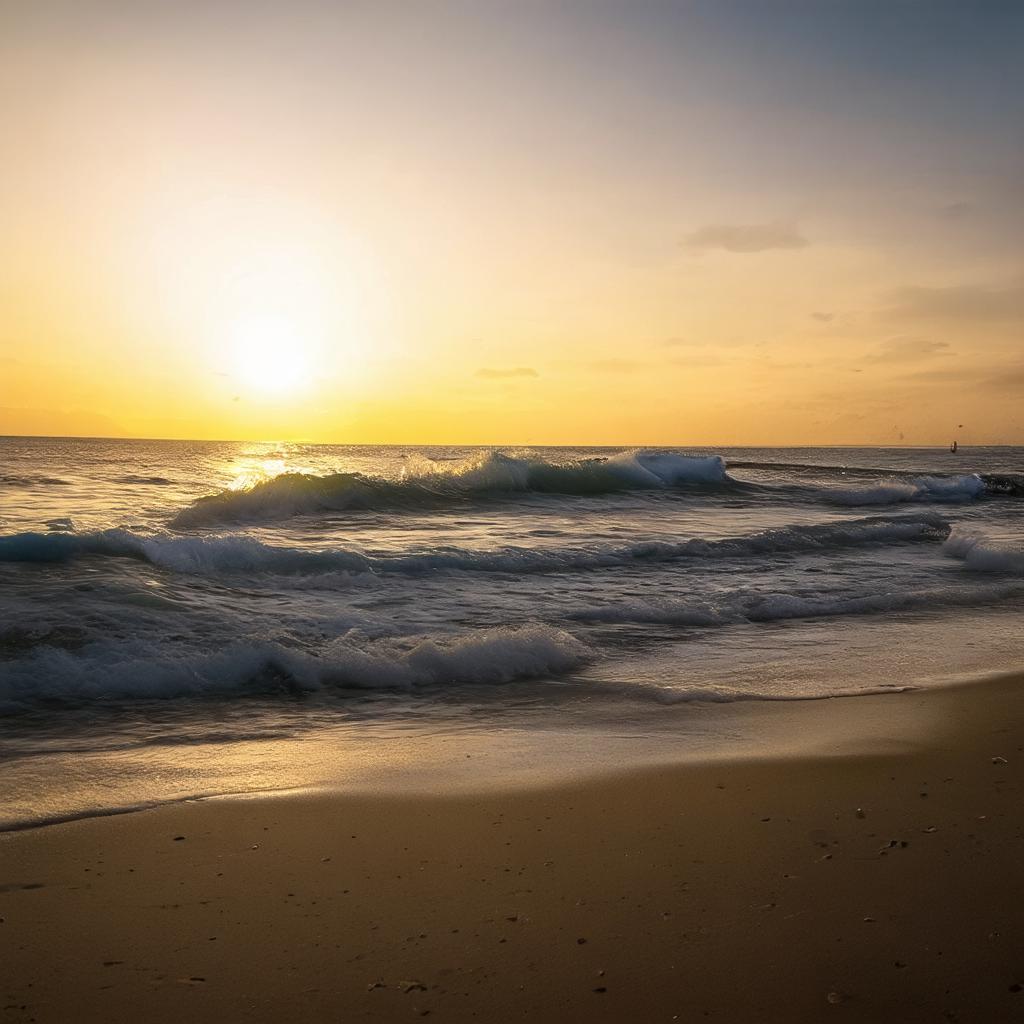}
    \end{minipage}
    \hfill
    \begin{minipage}{0.19\textwidth}
    \includegraphics[width=\textwidth]{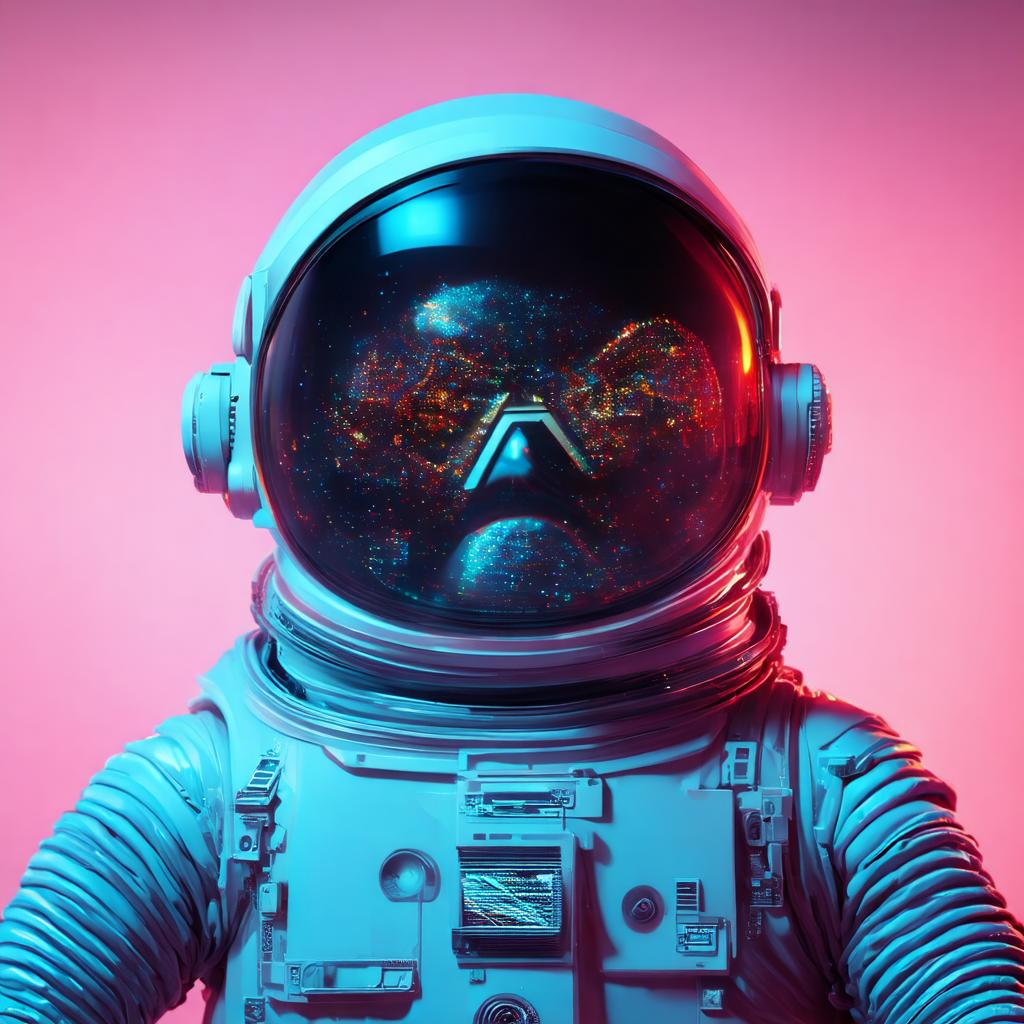}
    \end{minipage}
    \hfill
    \begin{minipage}{0.19\textwidth}
    \includegraphics[width=\textwidth]{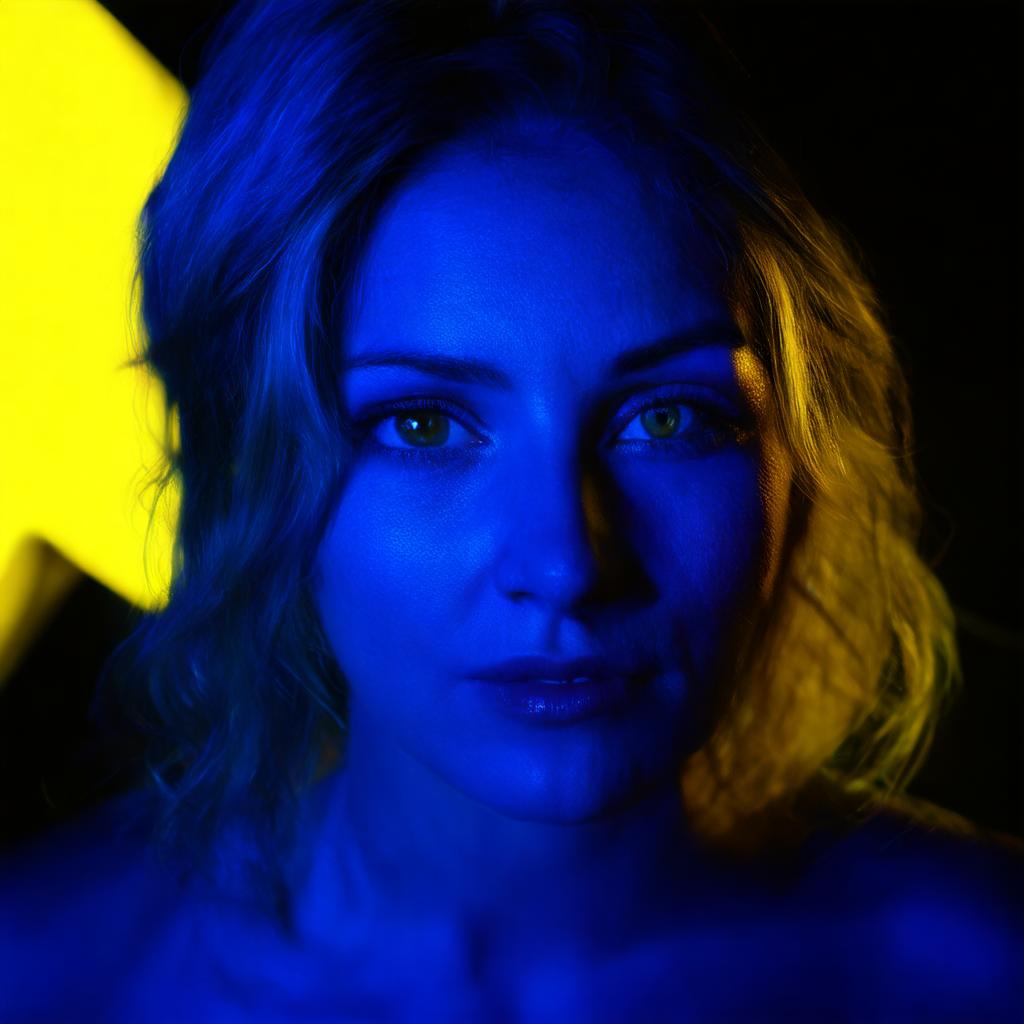}
    \end{minipage}
    \hfill
    \begin{minipage}{0.19\textwidth}
    \includegraphics[width=\textwidth]{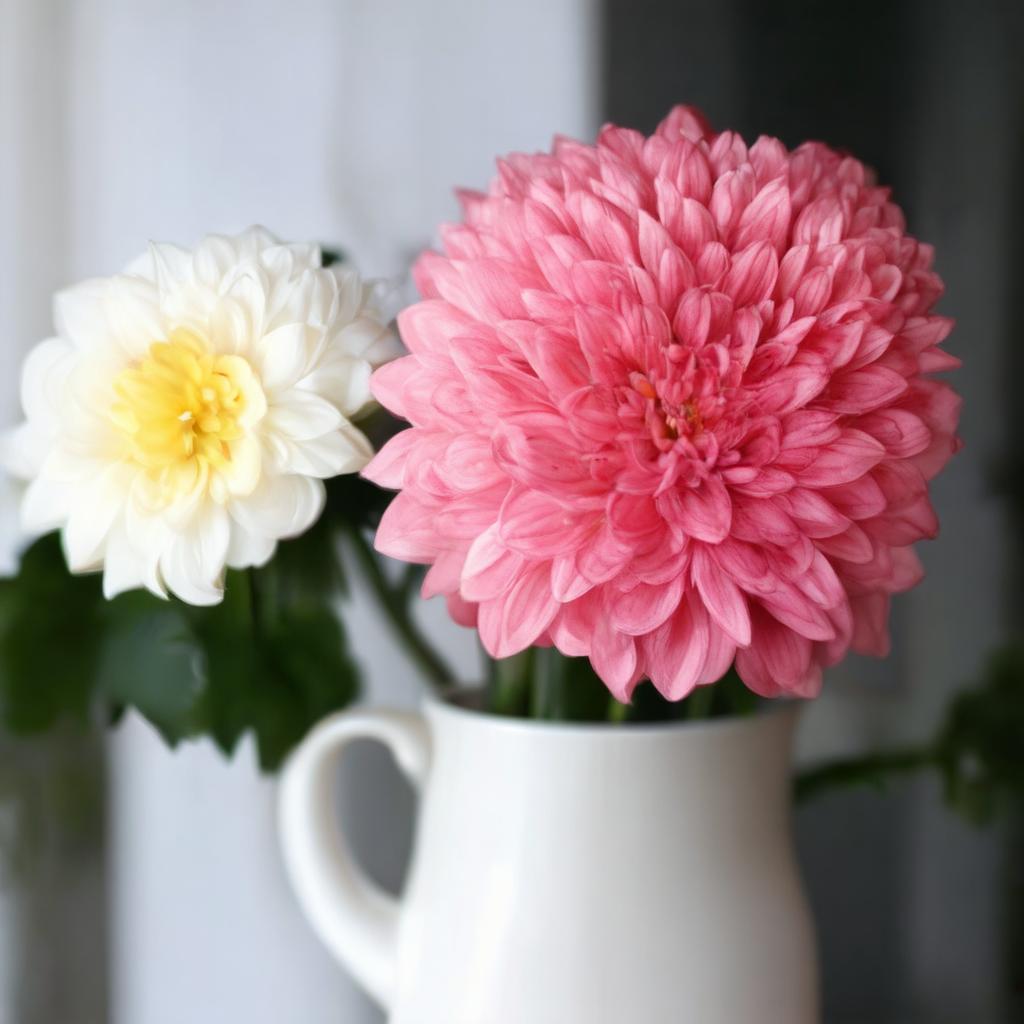}
    \end{minipage}
      \begin{minipage}{0.02\textwidth}\raggedright
       \rotatebox[origin=c]{90}{Linear MM-DiT-$\beta$}
    \end{minipage}
    \begin{minipage}{0.19\textwidth}
    \includegraphics[width=\textwidth]{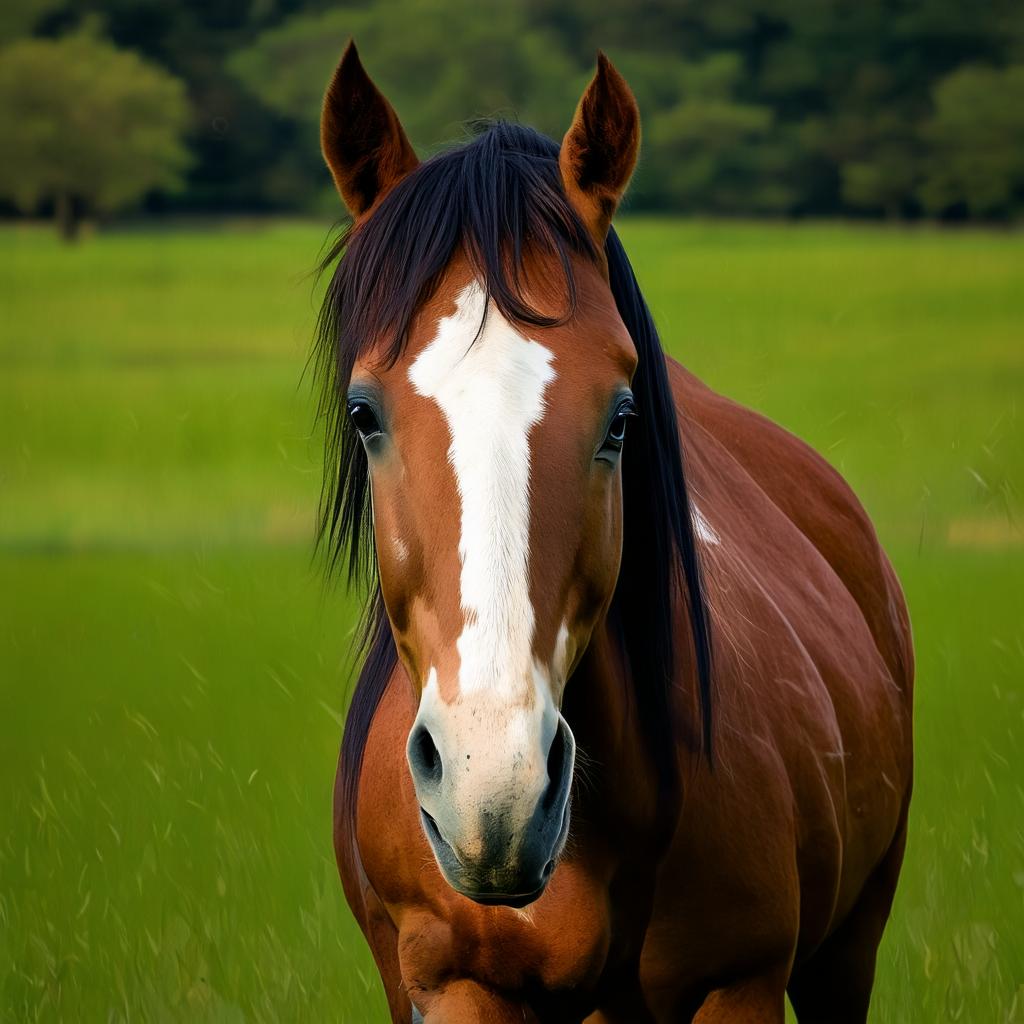}
    \end{minipage}
    \hfill
    \begin{minipage}{0.19\textwidth}
    \includegraphics[width=\textwidth]{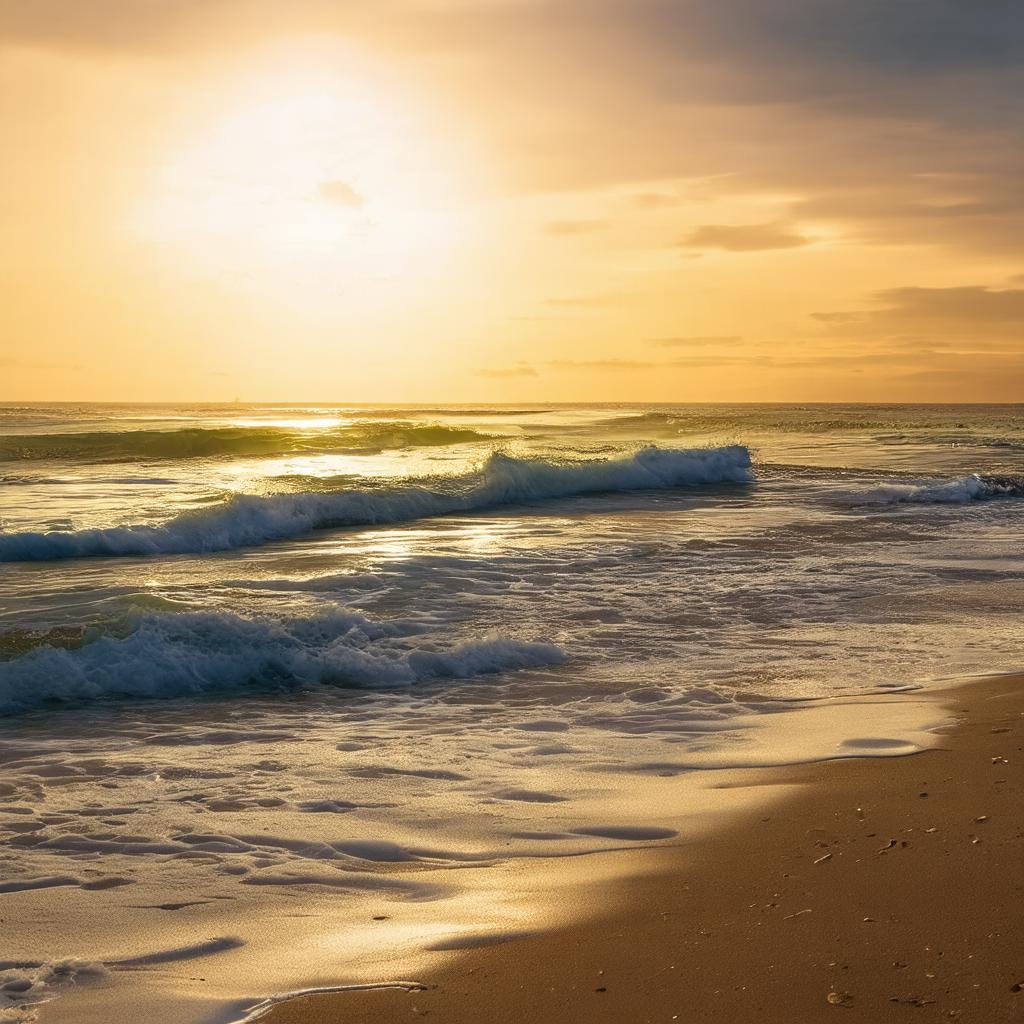}
    \end{minipage}
    \hfill
    \begin{minipage}{0.19\textwidth}
    \includegraphics[width=\textwidth]{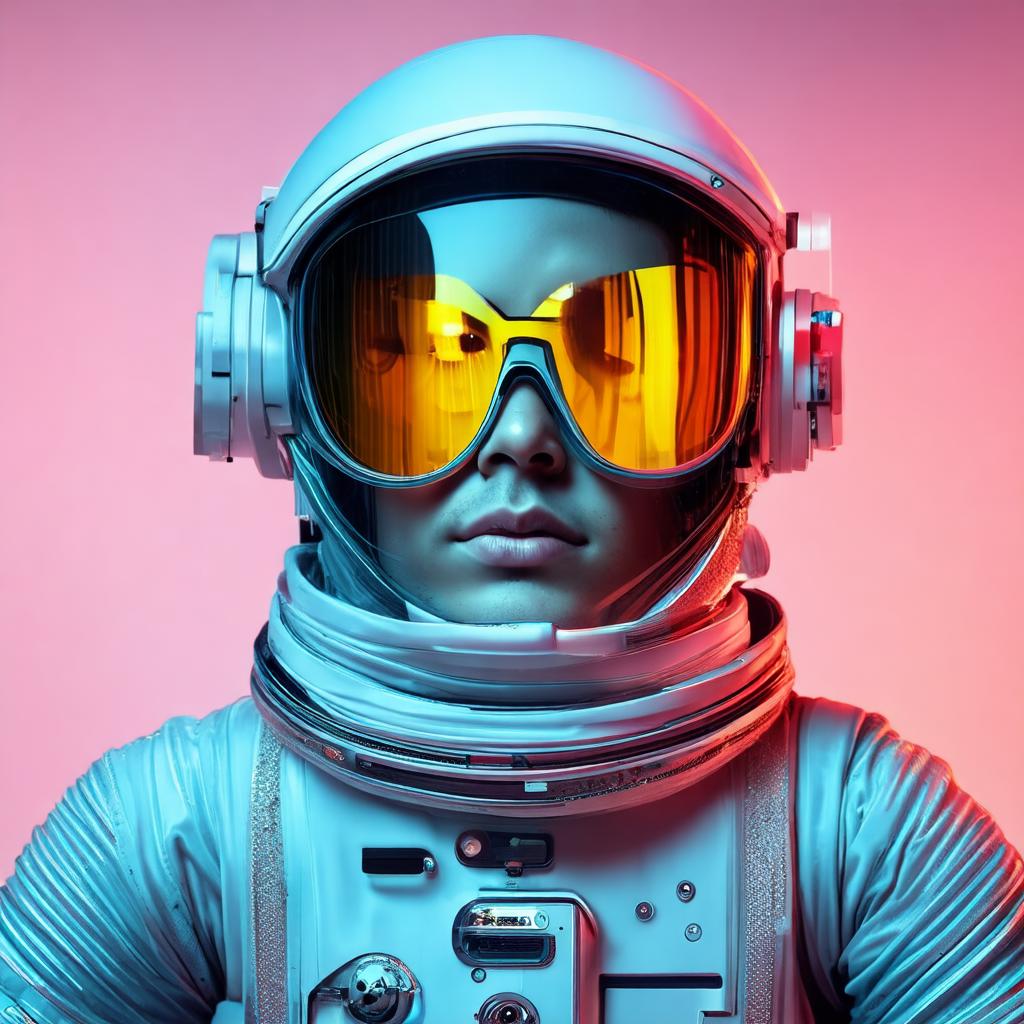}
    \end{minipage}
    \hfill
    \begin{minipage}{0.19\textwidth}
    \includegraphics[width=\textwidth]{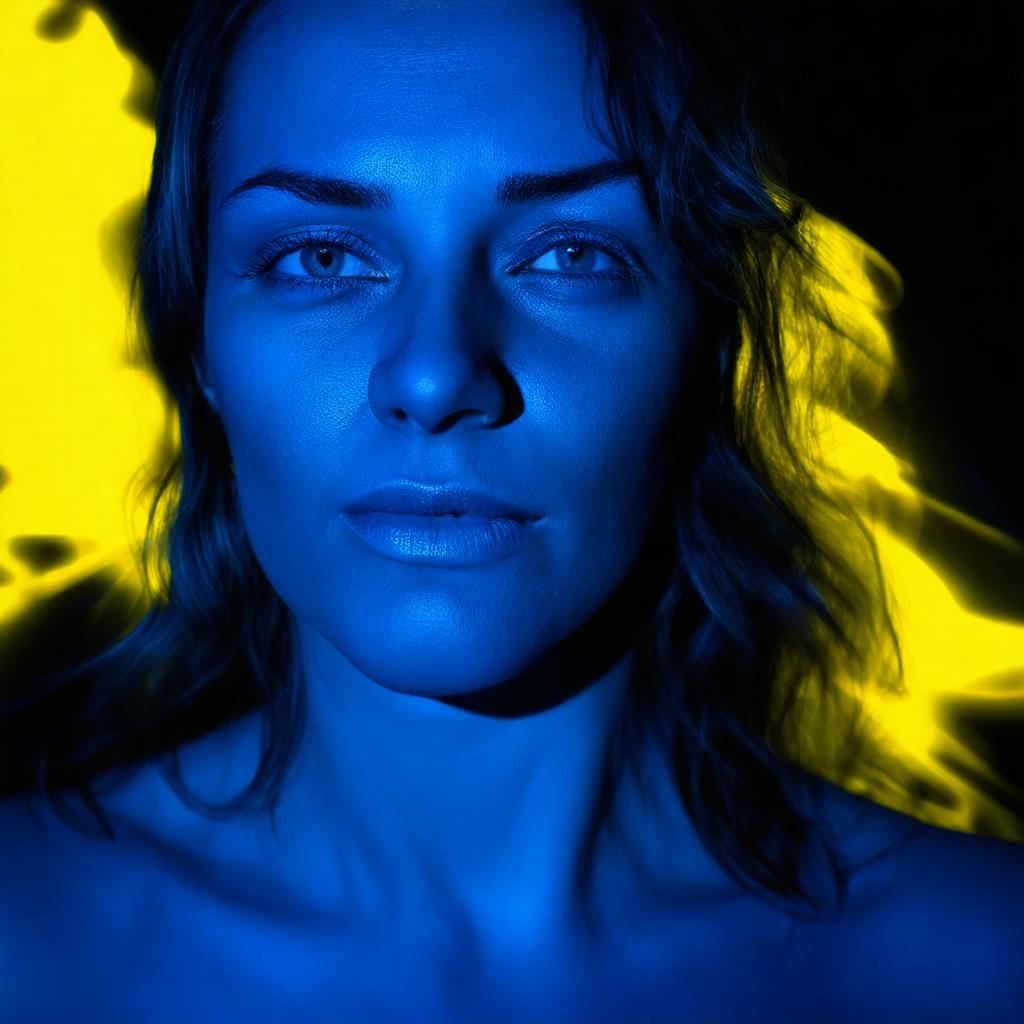}
    \end{minipage}
    \hfill
    \begin{minipage}{0.19\textwidth}
    \includegraphics[width=\textwidth]{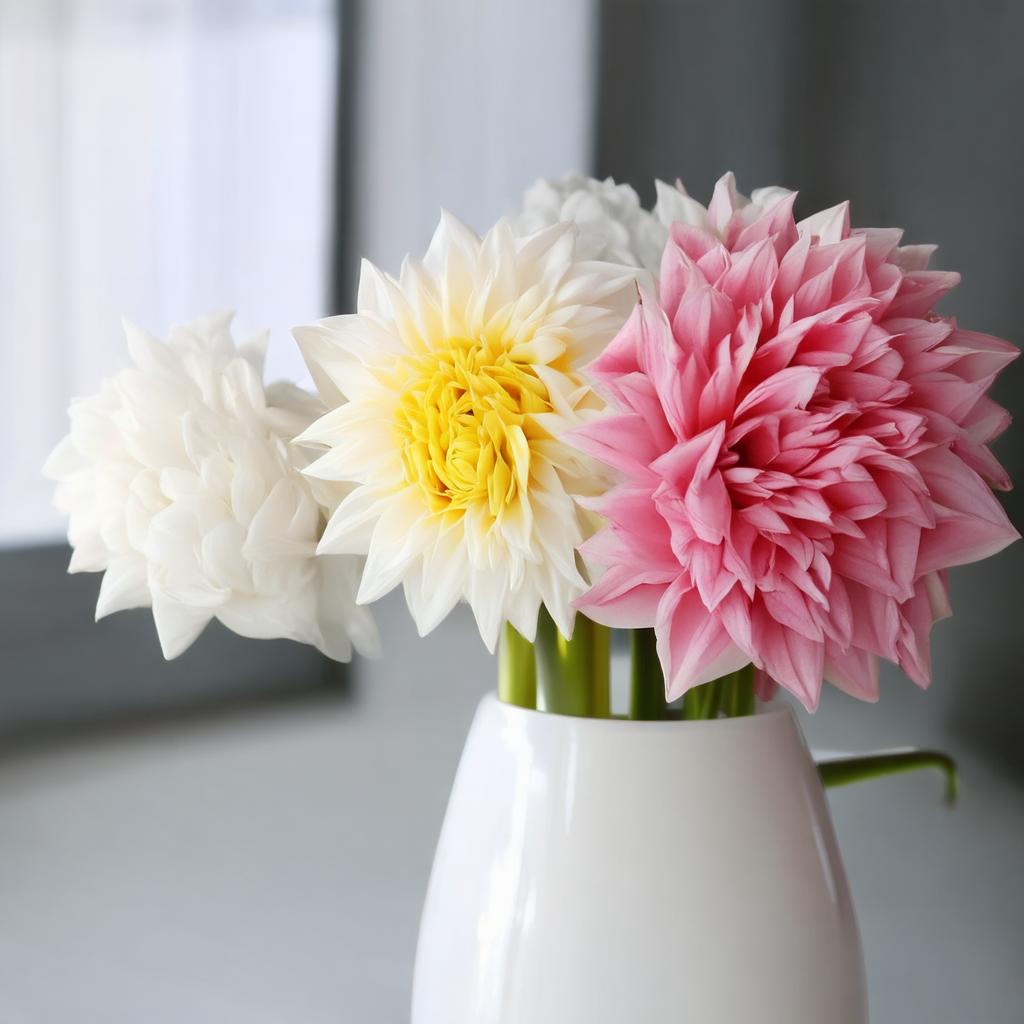}
    \end{minipage}
    \caption{$1024 \times 1024$ pixel images for the MM-EDiT ablations and baselines not shown in the main paper.}
    \label{fig:supp_sd35}
\end{figure*}

%% file: main.bbl
\begin{thebibliography}{24}
\providecommand{\natexlab}[1]{#1}
\providecommand{\url}[1]{\texttt{#1}}
\expandafter\ifx\csname urlstyle\endcsname\relax
  \providecommand{\doi}[1]{doi: #1}\else
  \providecommand{\doi}{doi: \begingroup \urlstyle{rm}\Url}\fi

\bibitem[Ba et~al.(2016)Ba, Kiros, and Hinton]{ba2016layer}
Jimmy~Lei Ba, Jamie~Ryan Kiros, and Geoffrey~E Hinton.
\newblock Layer normalization.
\newblock \emph{arXiv preprint arXiv:1607.06450}, 2016.

\bibitem[Chen et~al.(2024{\natexlab{a}})Chen, Ge, Xie, Wu, Yao, Ren, Wang, Luo, Lu, and Li]{chen2024pixartsigma}
Junsong Chen, Chongjian Ge, Enze Xie, Yue Wu, Lewei Yao, Xiaozhe Ren, Zhongdao Wang, Ping Luo, Huchuan Lu, and Zhenguo Li.
\newblock Pixart-$\sigma$: Weak-to-strong training of diffusion transformer for 4k text-to-image generation.
\newblock In \emph{European Conference on Computer Vision}, pages 74--91. Springer, 2024{\natexlab{a}}.

\bibitem[Chen et~al.(2024{\natexlab{b}})Chen, YU, GE, Yao, Xie, Wang, Kwok, Luo, Lu, and Li]{chen2024pixartalpha}
Junsong Chen, Jincheng YU, Chongjian GE, Lewei Yao, Enze Xie, Zhongdao Wang, James Kwok, Ping Luo, Huchuan Lu, and Zhenguo Li.
\newblock Pixart-\${\textbackslash}alpha\$: Fast training of diffusion transformer for photorealistic text-to-image synthesis.
\newblock In \emph{The Twelfth International Conference on Learning Representations}, 2024{\natexlab{b}}.

\bibitem[Dao and Gu(2024)]{dao2024transformers}
Tri Dao and Albert Gu.
\newblock Transformers are ssms: generalized models and efficient algorithms through structured state space duality.
\newblock In \emph{Proceedings of the 41st International Conference on Machine Learning}, pages 10041--10071, 2024.

\bibitem[Esser et~al.(2024)Esser, Kulal, Blattmann, Entezari, M{\"u}ller, Saini, Levi, Lorenz, Sauer, Boesel, et~al.]{mmdit}
Patrick Esser, Sumith Kulal, Andreas Blattmann, Rahim Entezari, Jonas M{\"u}ller, Harry Saini, Yam Levi, Dominik Lorenz, Axel Sauer, Frederic Boesel, et~al.
\newblock Scaling rectified flow transformers for high-resolution image synthesis.
\newblock In \emph{Forty-first international conference on machine learning}, 2024.

\bibitem[Fei et~al.(2024)Fei, Fan, Yu, Li, Zhang, and Huang]{fei2024dimba}
Zhengcong Fei, Mingyuan Fan, Changqian Yu, Debang Li, Youqiang Zhang, and Junshi Huang.
\newblock Dimba: Transformer-mamba diffusion models.
\newblock \emph{arXiv preprint arXiv:2406.01159}, 2024.

\bibitem[Gu and Dao(2023)]{gu2023mamba}
Albert Gu and Tri Dao.
\newblock Mamba: Linear-time sequence modeling with selective state spaces.
\newblock \emph{arXiv preprint arXiv:2312.00752}, 2023.

\bibitem[Hassani et~al.(2023)Hassani, Walton, Li, Li, and Shi]{hassani2023neighborhood}
Ali Hassani, Steven Walton, Jiachen Li, Shen Li, and Humphrey Shi.
\newblock Neighborhood attention transformer.
\newblock In \emph{Proceedings of the IEEE/CVF conference on computer vision and pattern recognition}, pages 6185--6194, 2023.

\bibitem[Hu et~al.(2024)Hu, Baumann, Gui, Grebenkova, Ma, Fischer, and Ommer]{hu2024zigma}
Vincent~Tao Hu, Stefan~Andreas Baumann, Ming Gui, Olga Grebenkova, Pingchuan Ma, Johannes Fischer, and Bj{\"o}rn Ommer.
\newblock Zigma: A dit-style zigzag mamba diffusion model.
\newblock In \emph{European Conference on Computer Vision}, pages 148--166. Springer, 2024.

\bibitem[Katharopoulos et~al.(2020)Katharopoulos, Vyas, Pappas, and Fleuret]{katharopoulos2020transformers}
Angelos Katharopoulos, Apoorv Vyas, Nikolaos Pappas, and Fran{\c{c}}ois Fleuret.
\newblock Transformers are rnns: Fast autoregressive transformers with linear attention.
\newblock In \emph{International conference on machine learning}, pages 5156--5165. PMLR, 2020.

\bibitem[Labs(2024)]{flux2024}
Black~Forest Labs.
\newblock Flux.
\newblock \url{https://github.com/black-forest-labs/flux}, 2024.

\bibitem[Lee et~al.(2025)Lee, Park, Cho, Lee, and Hwang]{lee2025koala}
Youngwan Lee, Kwanyong Park, Yoorhim Cho, Yong-Ju Lee, and Sung~Ju Hwang.
\newblock Koala: Empirical lessons toward memory-efficient and fast diffusion models for text-to-image synthesis.
\newblock \emph{Advances in Neural Information Processing Systems}, 37:\penalty0 51597--51633, 2025.

\bibitem[Liu et~al.(2024{\natexlab{a}})Liu, Tan, and Wang]{liu2024clear}
Songhua Liu, Zhenxiong Tan, and Xinchao Wang.
\newblock Clear: Conv-like linearization revs pre-trained diffusion transformers up, 2024{\natexlab{a}}.

\bibitem[Liu et~al.(2024{\natexlab{b}})Liu, Yu, Tan, and Wang]{liu2024linfusion}
Songhua Liu, Weihao Yu, Zhenxiong Tan, and Xinchao Wang.
\newblock Linfusion: 1 gpu, 1 minute, 16k image.
\newblock \emph{arXiv preprint arXiv:2409.02097}, 2024{\natexlab{b}}.

\bibitem[Liu et~al.(2023)Liu, Gong, and Liu]{liu2023flow}
Xingchao Liu, Chengyue Gong, and Qiang Liu.
\newblock Flow straight and fast: Learning to generate and transfer data with rectified flow.
\newblock In \emph{The Eleventh International Conference on Learning Representations}, 2023.

\bibitem[Peebles and Xie(2023)]{peebles2023scalable}
William Peebles and Saining Xie.
\newblock Scalable diffusion models with transformers.
\newblock In \emph{Proceedings of the IEEE/CVF international conference on computer vision}, pages 4195--4205, 2023.

\bibitem[Podell et~al.(2023)Podell, English, Lacey, Blattmann, Dockhorn, M{\"u}ller, Penna, and Rombach]{podell2023sdxl}
Dustin Podell, Zion English, Kyle Lacey, Andreas Blattmann, Tim Dockhorn, Jonas M{\"u}ller, Joe Penna, and Robin Rombach.
\newblock Sdxl: Improving latent diffusion models for high-resolution image synthesis.
\newblock \emph{arXiv preprint arXiv:2307.01952}, 2023.

\bibitem[Rombach et~al.(2022)Rombach, Blattmann, Lorenz, Esser, and Ommer]{rombach2022sd15}
Robin Rombach, Andreas Blattmann, Dominik Lorenz, Patrick Esser, and Bj\"orn Ommer.
\newblock High-resolution image synthesis with latent diffusion models.
\newblock In \emph{Proceedings of the IEEE/CVF Conference on Computer Vision and Pattern Recognition (CVPR)}, pages 10684--10695, 2022.

\bibitem[Sehwag et~al.(2024)Sehwag, Kong, Li, Spranger, and Lyu]{Sehwag2025patchmixer}
Vikash Sehwag, Xianghao Kong, Jingtao Li, Michael Spranger, and Lingjuan Lyu.
\newblock Stretching each dollar: Diffusion training from scratch on a micro-budget.
\newblock \emph{arXiv preprint arXiv:2407.15811}, 2024.

\bibitem[Vaswani et~al.(2017)Vaswani, Shazeer, Parmar, Uszkoreit, Jones, Gomez, Kaiser, and Polosukhin]{vaswani2017attention}
Ashish Vaswani, Noam Shazeer, Niki Parmar, Jakob Uszkoreit, Llion Jones, Aidan~N Gomez, {\L}ukasz Kaiser, and Illia Polosukhin.
\newblock Attention is all you need.
\newblock \emph{Advances in neural information processing systems}, 30, 2017.

\bibitem[Wu and He(2018)]{wu2018group}
Yuxin Wu and Kaiming He.
\newblock Group normalization.
\newblock In \emph{Proceedings of the European conference on computer vision (ECCV)}, pages 3--19, 2018.

\bibitem[Xie et~al.(2024)Xie, Chen, Chen, Cai, Tang, Lin, Zhang, Li, Zhu, Lu, et~al.]{xie2024sana}
Enze Xie, Junsong Chen, Junyu Chen, Han Cai, Haotian Tang, Yujun Lin, Zhekai Zhang, Muyang Li, Ligeng Zhu, Yao Lu, et~al.
\newblock Sana: Efficient high-resolution image synthesis with linear diffusion transformers.
\newblock \emph{arXiv preprint arXiv:2410.10629}, 2024.

\bibitem[Xie et~al.(2025)Xie, Chen, Zhao, Yu, Zhu, Lin, Zhang, Li, Chen, Cai, Liu, Zhou, and Hans]{xie2025sana15}
Enze Xie, Junsong Chen, Yuyang Zhao, Jincheng Yu, Ligeng Zhu, Yujun Lin, Zhekai Zhang, Muyang Li, Junyu Chen, Han Cai, Bingchen Liu, Daquan Zhou, and Song Hans.
\newblock Sana 1.5: Efficient scaling of training-time and inference-time compute in linear diffusion transformer.
\newblock \emph{arXiv preprint arXiv:2501.18427}, 2025.

\bibitem[Zhuo et~al.(2024)Zhuo, Du, Xiao, Li, Liu, Huang, Liu, Zhao, Wang, Ma, et~al.]{lumina}
Le Zhuo, Ruoyi Du, Han Xiao, Yangguang Li, Dongyang Liu, Rongjie Huang, Wenze Liu, Lirui Zhao, Fu-Yun Wang, Zhanyu Ma, et~al.
\newblock Lumina-next: Making lumina-t2x stronger and faster with next-dit.
\newblock \emph{arXiv preprint arXiv:2406.18583}, 2024.

\end{thebibliography}
